\documentclass[%
 aip,
 amsmath,amssymb,
 reprint,%
]{revtex4-1}

\usepackage{graphicx}
\usepackage{dcolumn}
\usepackage{bm}

\usepackage[utf8]{inputenc}
\usepackage[T1]{fontenc}
\usepackage{hyperref}

\usepackage{amssymb}
\usepackage{amsmath}
\usepackage{subfigure}

\usepackage{array,multirow}

\usepackage[dvipsnames]{xcolor}

\begin{document}

\preprint{AIP/123-QED}

\title[A Point-Cloud Deep Learning Framework for Prediction of Fluid Flow Fields on Irregular Geometries]{A Point-Cloud Deep Learning Framework for Prediction of Fluid Flow Fields on Irregular Geometries}

\author{Ali Kashefi}

\email{kashefi@stanford.edu(The author to whom correspondence may be addressed.)}
 \affiliation{Department of Civil and Environmental Engineering, 
Stanford University, Stanford, CA 94305, USA}

\author{Davis Rempe}%
 \email{drempe@stanford.edu}
\author{Leonidas J. Guibas}
\email{guibas@cs.stanford.edu}
\affiliation{ 
Department of Computer Science, Stanford 
University, Stanford, CA 94305, USA
}%



\begin{abstract}

We present a novel deep learning framework for flow field predictions in irregular domains when the solution is a function of the geometry of either the domain or objects inside the domain. Grid vertices in a computational fluid dynamics (CFD) domain are viewed as point clouds and used as inputs to a neural network based on the PointNet architecture, which learns an end-to-end mapping between spatial positions and CFD quantities. Using our approach, (i) the network inherits desirable features of unstructured meshes (e.g., fine and coarse point spacing near the object surface and in the far field, respectively), which minimizes network training cost; (ii) object geometry is accurately represented through vertices located on object boundaries, which maintains boundary smoothness and allows the network to detect small changes between geometries; and (iii) no data interpolation is utilized for creating training data; thus accuracy of the CFD data is preserved. None of these features are achievable by extant methods based on projecting scattered CFD data into Cartesian grids and then using regular convolutional neural networks. Incompressible laminar steady flow past a cylinder with various shapes for its cross section is considered. The mass and momentum of predicted fields are conserved. We test the generalizability of our network by predicting the flow around multiple objects as well as an airfoil, even though only single objects and no airfoils are observed during training. The network predicts the flow fields hundreds of times faster than our conventional CFD solver, while maintaining excellent to reasonable accuracy.

\end{abstract}


\maketitle

%

\section{\label{Introduction}Introduction and motivation}

One of the main contributions of machine learning techniques to Computational Fluid Dynamics (CFD) simulations is reducing the computational costs. Even with the presence of high performance computing tools (see e.g., Refs. \onlinecite{gropp2001high,gicquel2011high,jin2011high,moureau2011design,nagel2019high}) and efficient numerical schemes (see e.g., Refs. \onlinecite{gupta1997compact,zhang1998fast,filelis2014numerical,kashefi2018finite,kashefi2020coarse}) to accelerate CFD simulations, investigation of design parameters for device optimization remains computationally expensive mainly because a huge number of iterations for flow field analysis is required by users \cite{sekar2019fast}. Meanwhile, with an increase in the usage of CFD both in the industry and academia, a growth in the collection of CFD data has been observed \cite{meneveau2017turbulence,ansari2018data,yu2019integrated}. Consequently, deep learning schemes as a data-driven strategy have been recently utilized in the CFD community as a solution to overcome the computational burden. To take advantage of this approach, a few or all the components of a CFD solver have been replaced by a neural network (see e.g., Refs. \onlinecite{sekar2019fast,guo2016convolutional,tompson2017accelerating,thuerey2019deep,bhatnagar2019prediction}). For instance, \citet{xiao2018novel} utilized a Convolutional Neural Network (CNN), but only as a replacement of the pressure Poisson equation, which is the most time consuming component of pressure correction methods \cite{chorin1968numerical, temam1969approximation,guermond2006overview,kashefi2020coarseb,kashefi2020coarseC}. They achieved a speedup factor of 90 for prediction of the inviscid flow field in comparison with the multigrid method introduced by \citet{mcadams2010parallel}. This is while \citet{thuerey2019deep} used U-Net \cite{ronneberger2015u} to completely skip a CFD solver and predicted the velocity and pressure fields around an airfoil.

To use neural networks as a replacement of CFD solvers, it is important to efficiently feed CFD data into a network, and thus, an effective data representation is crucial. The connection of neural networks with Cartesian grids is straightforward. For this scenario, using two and three-dimensional CNNs is a popular methodology among the CFD community (see e.g., Refs. \onlinecite{fukami2019super,lapeyre2019training,kim2020prediction}). In this method, each vertex of a Cartesian grid corresponds to a pixel of an image processed by a CNN. However, in real-world applications with complex geometries, employing unstructured grids is unavoidable. In contrast with Cartesian grids, the connection of unstructured grids, and consequently scatter CFD data, with neural networks becomes challenging, specifically because the flow field is function of the geometry of the CFD domain. The approach to connect scatter CFD data (arisen from unstructured grids) to a two or three-dimensional CNN is to use pixelation (voxelisation). In pixelation, scatter CFD data is projected into a two or three-dimensional Cartesian grid such that they become amenable to traditional CNNs (see e.g., Refs. \onlinecite{sekar2019fast,guo2016convolutional,tompson2017accelerating,thuerey2019deep,bhatnagar2019prediction,jin2018prediction,zhang2018application,han2019novel,hui2020fast,hasegawa2020cnn,duru2020cnnfoil,miyanawala2017aned,miyanawala2018bluff,mao2018cnn,bukka2020deep,miyanawala2019structure,bukka2020assessment,reddy2019reduced,peng2020unsteady,tangsali2020generalizability}).

\begin{figure}[hbt!]
\centering
\includegraphics[width=0.6\linewidth]{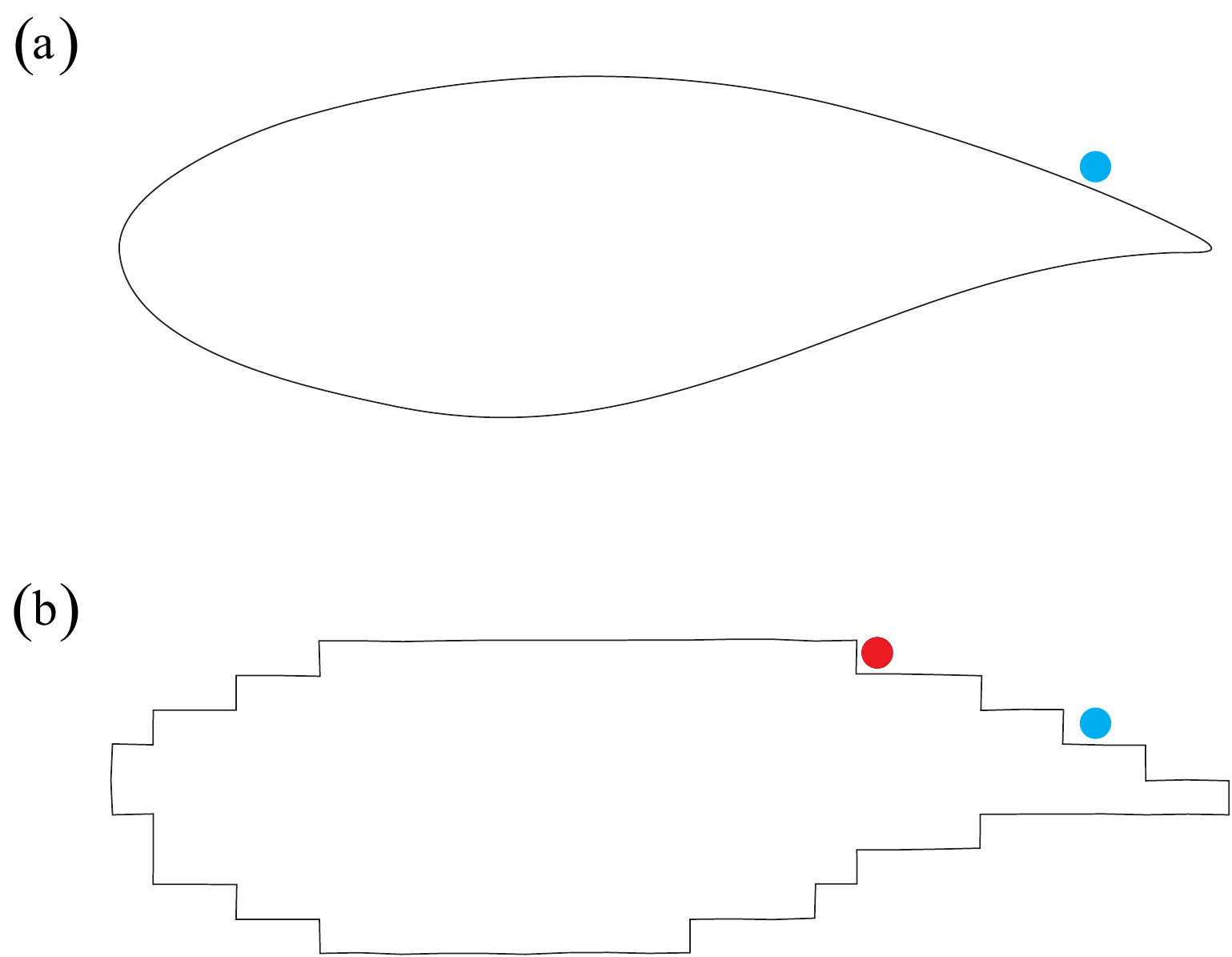}
\caption{Geometrical representation of the airfoil of RAE2822 \textbf{(a)} by vertices of an unstructured grid, and \textbf{(b)} by a pixelation procedure after a projection into a $32\times32$ Cartesian grid; The blue dot in \textbf{(a)} shows a virtual separation point in an imaginary Reynolds number, however, the corresponding point in \textbf{(b)} cannot characterize the separation as we expect that the flow separates at most at the red dot due to the artificial steps introduced to the surface by the pixelation.}
\label{Fig0}
\end{figure}

The pixelation strategy has five main shortcomings. First, the pixelation technique leads to decreasing the order of accuracy of CFD data due to the data interpolation or extrapolation. Nowadays, a considerable amount of effort is undertaken to design high-order methods for CFD solvers (e.g., see Ref. \onlinecite{bastian2019matrix}). However, all this effort is ignored by the pixelation approach. For instance, Fig. 8 of Ref. \onlinecite{bhatnagar2019prediction} shows the error introduced to the velocity field of a S814 airfoil as a consequence of the interpolation. Second, while the flow field around an object is highly sensitive to small changes (e.g., rotation or length increment of the object), the pixelation method cannot capture these changes unless a CNN with super resolved input is used, which by itself imposes high computational cost to the system. To the best of our knowledge, the maximum resolution of a CNN used in the CFD area (with application to prediction of flow around an object) is 201 $\times$ 301 pixels in two dimensions \cite{miyanawala2017aned} and 32 $\times$ 32 $\times$ 32 in three dimensions \cite{guo2016convolutional}. Thus, the study of geometrical parameters for an optimization goal using deep learning techniques remains unsolved in practice. Third, it is common to mask the interior points of objects (e.g., an airfoil) in the pixelation approach (see e.g., Fig. 1 of Ref. \onlinecite{thuerey2019deep} and Fig. 3 of Ref. \onlinecite{bhatnagar2019prediction}). However, by ignoring network-predicted pixels that lie inside the object, some computational capacity of the CNN is wasted. Fourth, the importance of information in a CFD domain is not equal. For instance, the velocity and pressure fields near the surface of an airfoil and in its wake region are more important than other areas. Nonetheless, when using pixelation, and consequently Cartesian grids, the distribution of CNN pixels is uniform everywhere in the domain. Thus, refining the resolution in the area of interest requires refining everywhere else too, which is computationally inefficient. Fifth, the pixelation procedure is usually along with coarsening previously-smooth boundaries of a shape and introduces artificial roughness to its surface (see e.g., Fig. 5 of Ref. \onlinecite{sekar2019fast}). This error can dramatically change the flow features such as the location of the detachment point on the surface of an airfoil (see e.g., Fig. \ref{Fig0}).

To obviate the aforementioned problems, we present a simple and elegant deep learning framework. This framework is fundamentally based on the PointNet architecture proposed by \citet{qi2017pointnet} in 2017. PointNet \cite{qi2017pointnet} was introduced for classification and segmentation of three-dimensional objects represented as point clouds. For our targets, grid vertices of a CFD domain are imagined as a point cloud and an end-to-end mapping between the spatial position of a vertex and its velocity and pressure values is defined. Accordingly, we take the segmentation track of PointNet and use mean squared error as a loss function to solve a gradient descent optimization problem. Using this framework, the point cloud simultaneously represents both the geometry of the shape and the space of flow field. There are several main advantages with this strategy. First, because no data interpolation or extrapolation is employed in this framework, the order of accuracy of CFD data is preserved for neural network training purposes. Second, object geometries are represented with the same accuracy as in the grid representation, since vertices that make up the point cloud are located on object boundaries in the CFD domain. Hence, the network is able to capture even minor variations from one geometry to another. Third, no artificial effects are introduced to the object geometry (see Fig. \ref{Fig0}). Fourth, since the point cloud is identical to the generated grid vertices, it inherits the mesh features in terms of spatial grid distribution (i.e., fine grids near the object and coarse grids in far fields). Thus, the computational expenses of the neural network training are optimized. Fifth, the spatial size of the physical domain can vary from one data to another. Our deep learning methodology is provided in detail in Sect. \ref{22}.

We assess the efficacy of the introduced deep learning framework through a representative benchmark problem: flow past a cylinder with different cross sections. An external flow past a cylinder is a model for canonical CFD problems and has been widely used for the study of incompressible Navier-Stokes equations in two dimensions (see e.g., Ref. \onlinecite{karniadakis1988numerical}). To generate our data, we select various cross sections with standard shapes already used in the literature: circle \cite{behr1995incompressible}, square \cite{sen2011flow}, triangle \cite{kumar2006numerical}, rectangle \cite{zhong2019flow}, ellipse \cite{mittal1996direct}, pentagon \cite{abedin2017simulation}, and hexagon \cite{abedin2017simulation}. For each shape, we consider different aspect ratios of its lengths and different poses. Length scales of the shapes and physical properties of the fluid are set such that the flow remains steady for all the data generated. Further details of data generation are provided in Sect. \ref{23}. We analyze the performance of our deep learning framework and accuracy of its prediction in several ways. First, we carry out a pointwise error measurement of the velocity and pressure fields predicted around shapes not seen during training. Second, to evaluate the ability of our deep learning framework for conservation of mass and momentum, we compute the residual of continuity and Navier-Stokes equations for the predicted flow fields. Third, we investigate the ability of our neural network for the prediction of the velocity and pressure fields around multiple objects placed within a certain distance from each other, though the network has only seen one object per data during the training process. Fourth, we examine the capability of the deep learning methodology for prediction of the velocity and pressure fields around a standard airfoil, though the network has never seen an airfoil during the training process.

The rest of this article is structured as follows. Section \ref{21} provides the governing equations of incompressible viscous flows as well as our strategy for the CFD aspects. We explain the procedure of data generation in Sect. \ref{23}. The structure of the neural network is illustrated in Sect. \ref{22}. Network training is described in Sect. \ref{24}. A typical performance analysis of the network along with graphical representations are discussed in Sect. \ref{31}. Conservation of mass and momentum for predicted fields are investigated in Sect. \ref{32}. We explain how different components of the network represent the problem physics in Sect. \ref{33prime}. We discuss the generalizability of our network in Sect. \ref{33}. Results of flow prediction around multiple objects are allocated in Sect. \ref{331}. Prediction of the velocity and pressure spaces around an airfoil is presented in Sect. \ref{332}. Potentials of the neural network for the prediction of turbulent flows are discussed in Sect. \ref{34}. Conclusions and directions for our future researches are given in Sect. \ref{Conclusions}.

\begin{figure*}
\centering
\includegraphics[width=1\linewidth]{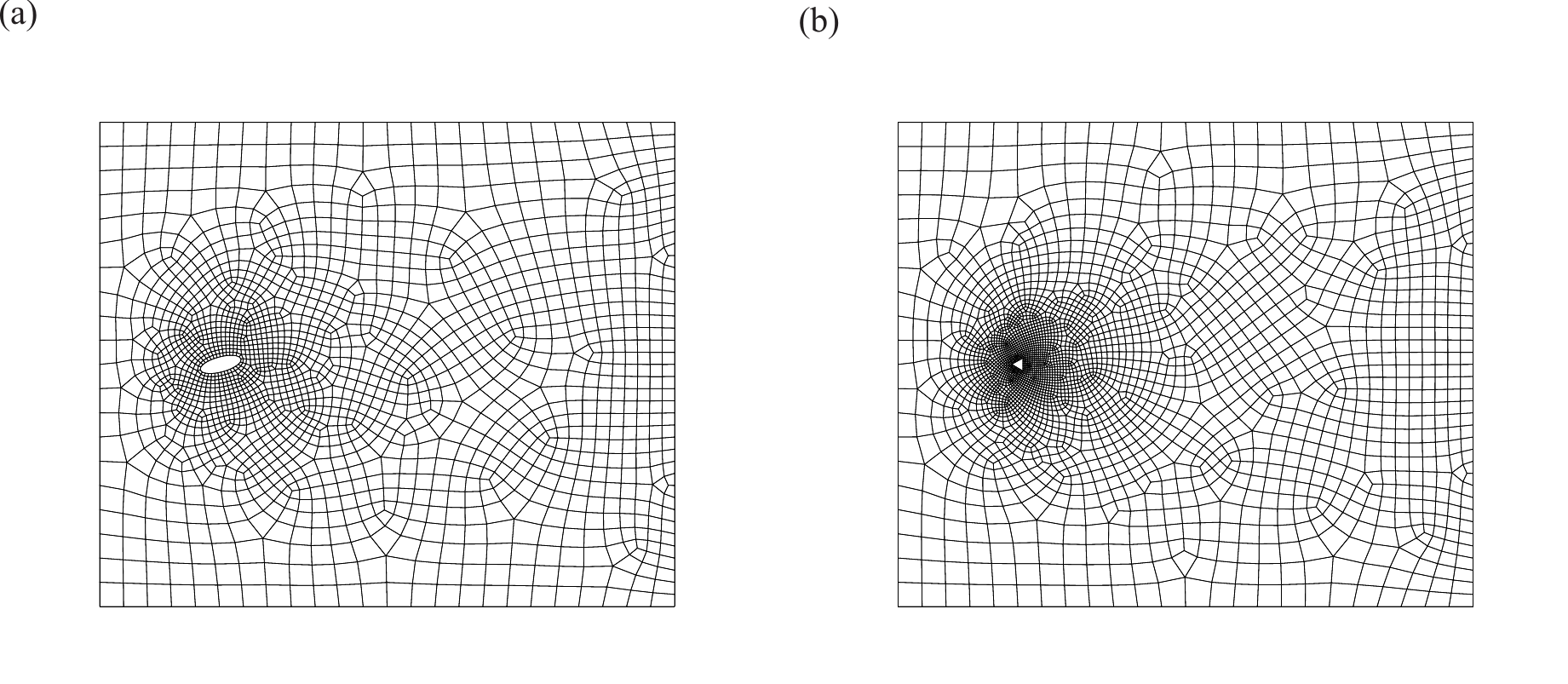}
\caption{Representation of the finite volume meshes used for solving the continuity and Navier-Stokes equations in the simulation of flow over a cylinder with \textbf{a} an elliptical cross section, 2672 vertices; and with \textbf{b} a triangular cross section, 2775 vertices}
\label{Fig1}
\end{figure*}

\begin{figure*}
\centering
\includegraphics[width=1.0\linewidth]{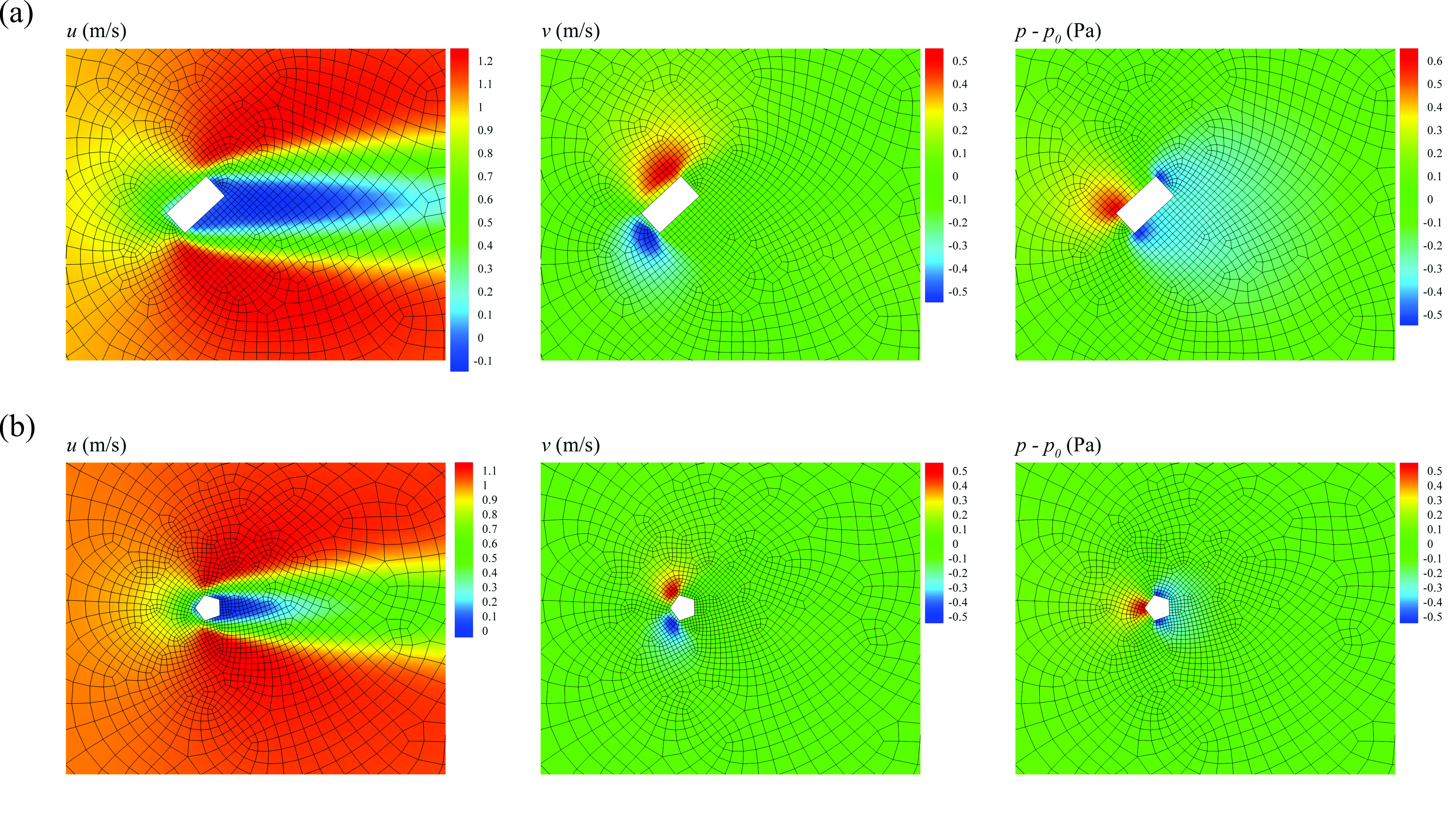}
\caption{Velocity and pressure fields for the steady-state flow over a cylinder with \textbf{a} rectangular cross section and \textbf{b} pentagonal cross section; $p_0$ is the atmospheric pressure.}
\label{Fig2}
\end{figure*}

\section{\label{ProblemFormulation}Problem formulation and methodology}

\subsection{\label{21}Governing equations of fluid dynamics}

The Navier-Stokes and continuity equations respectively describe the conservation of momentum and mass of an incompressible viscous Newtonian fluid flow. The governing equations along with boundary conditions are given by
\begin{eqnarray}
\label{Eq1}
\rho \bigg[\frac{\partial \textbf{\textit{u}}}{\partial t}+(\textbf{\textit{u}}\cdot \nabla)\textbf{\textit{u}}\bigg] -\mu \Delta \textbf{\textit{u}} + \nabla p=\textbf{\textit{f}} \textrm{ in } V,
\end{eqnarray}
\begin{eqnarray}
\label{Eq2}
\nabla \cdot \textbf{\textit{u}}=0 \textrm{ in } V,
\end{eqnarray}
\begin{eqnarray}
\label{Eq3}
\textbf{\textit{u}}=\textbf{\textit{u}}_{\Gamma_D} \textrm{ on } \Gamma_D,
\end{eqnarray}
\begin{eqnarray}
\label{Eq4}
-p\textbf{\textit{n}}+\mu \nabla \textbf{\textit{u}} \cdot \textbf{\textit{n}}=\textbf{\textit{t}}_{\Gamma_N} \textrm{ on } \Gamma_N,
\end{eqnarray}
where $\textbf{\textit{u}}$ indicates the velocity vector and $p$ denotes the absolute pressure of fluid in the space of $V$. $\rho$ and $\mu$ stand for the fluid density and the dynamic viscosity, respectively. The external body force vector is represented by $\textbf{\textit{f}}$. $\Gamma_D$ and $\Gamma_N$ are respectively the Dirichlet and Neumann boundaries, while there is no overlapping between them. $\textbf{\textit{t}}_{\Gamma_N}$ is the stress vector acting on $\Gamma_N$, while $\textbf{\textit{n}}$ is the outward unit vector normal to $\Gamma_N$. Furthermore, we indicate $x$ and $y$ components of the velocity vector respectively by $u$ and $v$.

Our CFD domain (which differs from the neural network domain) is a rectangular field $V:=$[0, 38 m]$\times$[0, 32 m]. An arbitrary two dimensional shape with center of mass lying at the point (8 m, 16 m) and a length scale of $L$ represents the cross section of an infinite cylinder. The object boundaries are assumed to be rigid, while we impose no-slip conditions on its surfaces. Free stream velocity with the magnitude of $u_{\infty}$ and parallel to the $x$-axis is enforced at the inflow, bottom, and top of the field. The open boundary condition 
\begin{equation}
\label{Eq5}
-p\textbf{\textit{n}}+\mu \nabla \textbf{\textit{u}} \cdot \textbf{\textit{n}}=0
\end{equation}
is specified in the outflow velocity. Note that the geometry and boundary conditions described here are mainly accorded to the literature \cite{behr1995incompressible,ding2004simulation,kashefi2018finite,kashefi2020coarse} to satisfy far-field assumptions. The fluid density and free stream velocity are set to 1.00, and the fluid viscosity is set to 0.05 in the International Unit System. The Reynolds number is calculated as
\begin{equation}
\label{Eq6}
Re=\frac{\rho L u_\infty}{\mu}.
\end{equation}
Given the fixed fluid density and viscosity as well as the free stream velocity, the only free parameter to vary the Reynolds number is the length scale, which depends on the object geometry.

For any given object with an arbitrary shape, the geometry of space $V$ is specified. We use the Gmsh \cite{geuzaine2009gmsh} application for the discretization of space $V$ and generating unstructured finite volume meshes with fine grid spacing near the object. Additionally, more grid points are placed in the wake region to capture the secondary flows accurately. As two examples, Figure \ref{Fig1} depicts the unstructured finite volume meshes for elliptical and triangular cross sections. In the next stage, we employ the OpenFOAM software \cite{weller1998tensorial} in order to obtain the numerical solution to the governing equations. Specifically, Semi-Implicit Method for Pressure Linked Equations (SIMPLE) \cite{caretto1973two} is used. The numerical computations are executed until the $L^2$ norm of residuals for both the continuity and momentum equations reaches $10^{-3}$. Accordingly, we consider a converged solution as steady-state flow if this criterion is satisfied. Figure \ref{Fig2}, for instance, demonstrates the velocity and pressure fields obtained by the CFD software for cylinders with rectangular and pentagonal cross sections.

As a final note in this subsection, flow past a cylinder is an example of external flows when the solution is a function of the geometry of inner boundaries. However, our deep learning framework introduced in this study is also usable for flow predictions in cases when variations in the geometry of outer boundaries matter (see e.g., Ref. \onlinecite{xiao2020flows}).

\subsection{\label{23}Data generation}

As discussed in Sect. \ref{21}, we investigate the prediction of viscous flow past a cylinder to evaluate the capability of the neural network presented in this study. The solution of the continuity and Navier-Stokes equations is a function of the shape of the cylinder cross section. With this in mind, variations in the shape of the cross section generate the data set used to train our model. Seven various shapes are considered: circle \cite{behr1995incompressible}, square \cite{sen2011flow}, triangle \cite{kumar2006numerical}, rectangle \cite{zhong2019flow}, ellipse \cite{mittal1996direct}, pentagon \cite{abedin2017simulation}, and hexagon \cite{abedin2017simulation}. In practice, we extend our data set by rotating the cylinder and altering the cross section size. A summary of the cross section shapes is tabulated in Table \ref{Tab1}. Considering the information provided in Table \ref{Tab1} along with the geometrical definition of $a$ and $b$ for each cross section, the length scale ($L$) for the cross sections with the shape of circle, equilateral hexagon, equilateral pentagon, square, and equilateral triangle is $a$; while the length scale ($L$) is $b$ for the rectangle, ellipse, and triangle. Consequently, the range of the Reynolds number ($Re$) is 20.0 to 84.0 in the current study.

\begin{table*}
\caption{\label{Tab1}Description of the generated data}
\begin{ruledtabular}
\begin{tabular}{lllll}
 Shape & 
\vtop{\hbox{\strut Schematic}\hbox{\strut figure}}
 & \vtop{\hbox{\strut Variation in}\hbox{\strut orientation}} & \vtop{\hbox{\strut Variation in}\hbox{\strut length scale}} & \vtop{\hbox{\strut Number}\hbox{\strut of data}}\\
\hline
Circle & \includegraphics[width=0.08\linewidth]{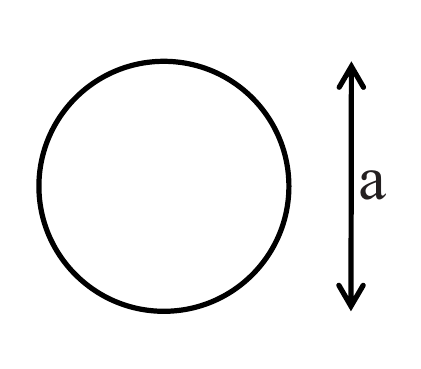} & - & $a=1$ m & 1 \\
Equilateral hexagon & \includegraphics[width=0.08\linewidth]{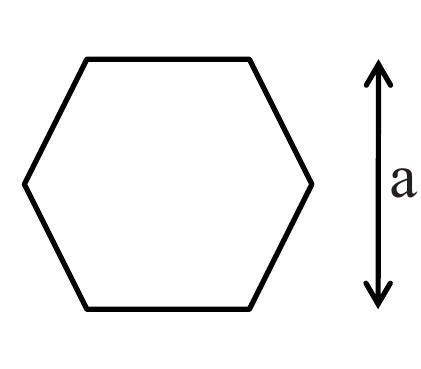} & $3^\circ$, $6^\circ$, \ldots, $60^\circ$ & $a=1$ m & 20 \\
Equilateral pentagon & \includegraphics[width=0.08\linewidth]{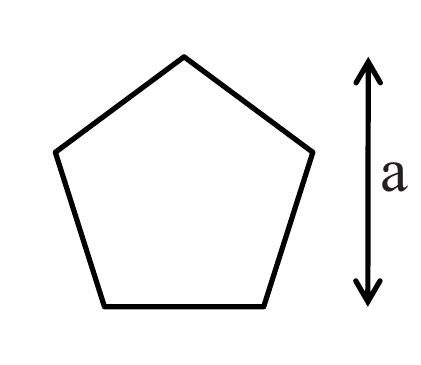} & $3^\circ$, $6^\circ$, \ldots, $72^\circ$ & $a=1$ m & 24 \\
Square & \includegraphics[width=0.08\linewidth]{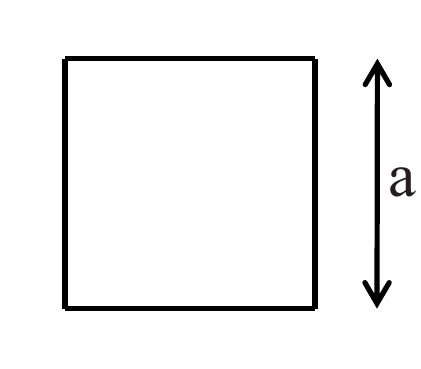} &  $3^\circ$, $6^\circ$, \ldots, $90^\circ$ & $a=1$ m & 30 \\
Equilateral triangle & \includegraphics[width=0.08\linewidth]{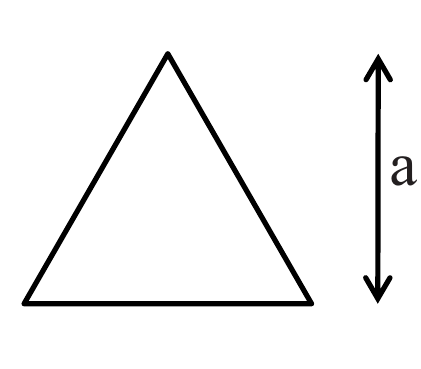} & $3^\circ$, $6^\circ$, \ldots, $180^\circ$ & $a=1$ m & 60\\
Rectangle & \includegraphics[width=0.08\linewidth]{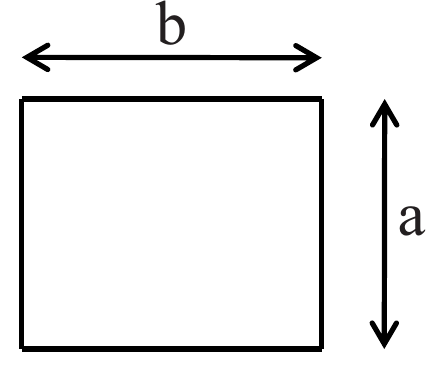} &  $3^\circ$, $6^\circ$, \ldots, $180^\circ$ & $a=1$ m; $b/a=$ 1.2, 1.4, \ldots, 3.6 & 780 \\
Ellipse & \includegraphics[width=0.08\linewidth]{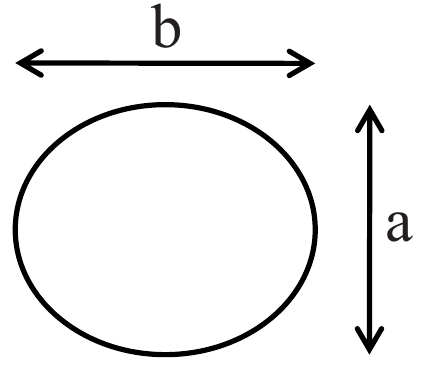} &  $3^\circ$, $6^\circ$, \ldots, $180^\circ$ & $a=1$ m; $b/a=$ 1.2, 1.4, \ldots, 4.2 & 960\\
Triangle & \includegraphics[width=0.08\linewidth]{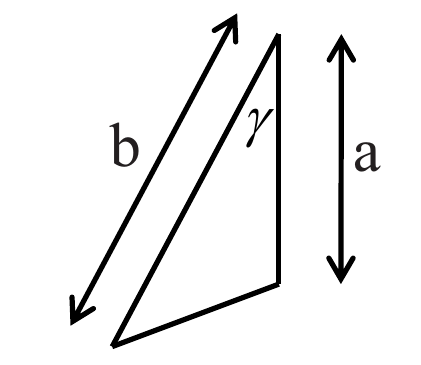} &  $3^\circ$, $6^\circ$, \ldots, $360^\circ$ & \vtop{\hbox{\strut $a=1$ m; $b/a=$ 1.5, 1.75}\hbox{\strut $\gamma=$ 40$^\circ$, 60$^\circ$, 80$^\circ$}} & 720 \\
\end{tabular}
\end{ruledtabular}
\end{table*}

We already described the procedure of data generation for a single shape in Sect. \ref{21}. Given a shape for the cross section, the geometry of the domain needs to be defined mathematically. Afterwards, an unstructured grid needs to be generated. Eventually, the CFD solver needs to numerically obtain the velocity and pressure spaces. Obviously, this procedure cannot be executed manually for 2595 different geometries and automation is unavoidable. This mission is accomplished using a set of \texttt{C++} codes, journal files, and batch files to automatically run the Gmsh application for mesh generation, and sequentially run the CFD software for solving Eqs. \ref{Eq1}--\ref{Eq4}, and finally write the numerical solutions on a hard disk. Note that for each orientation of bodies inside the domain (see the third column of Table \ref{Tab1}), we regenerate the mesh rather than rotating the previously-discretized domain, which is different from usual data augmentation techniques.

We described our CFD domain in Sect. \ref{21}. However, our neural network domain ($V_{NN}$) does not have to be the same as our CFD domain ($V$). Generally, a user of our machine learning framework can take any subdomain into account such that $V_{NN} \subset V$, and train the neural network for that specific area of interest. For the present problem, the flow field around the cylinder and in the wake region is critical, whereas the far-field data are unimportant. The input of our neural network is a set of the spatial coordinates of a point cloud. The point cloud takes the first $N$ closest grid points (vertices) to the center of mass of the cylinder. In other words, $V_{NN}$ contains $N$ points. In this study, we set $N=1024$. The output of our neural network is the corresponding velocity and pressure quantities at the given input points. It is important to note that our network covers the whole domain of interest using only $N=1024$ input points. In contrast, using regular CNNs needs a considerably higher number of input points (e.g., 128$\times$128 used in Ref. \onlinecite{thuerey2019deep}, and 150$\times$150 used in Ref. \onlinecite{bhatnagar2019prediction}).

\begin{figure*}
\centering
\includegraphics[width=1.0\linewidth]{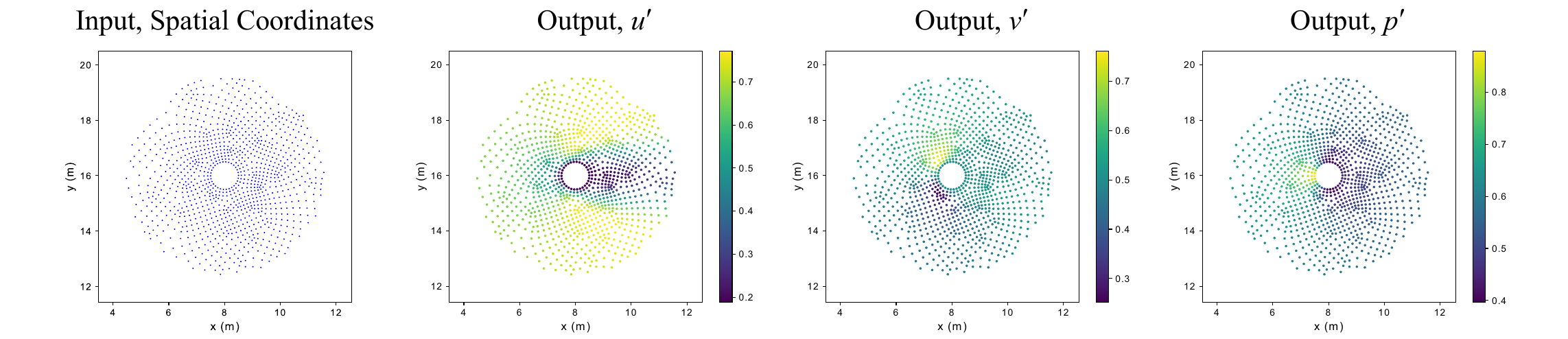}
\includegraphics[width=1.0\linewidth]{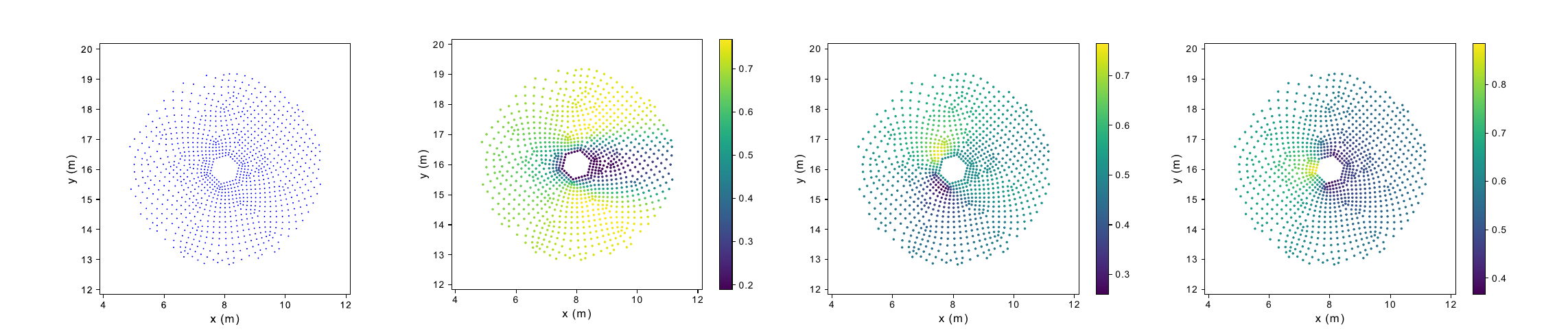}
\includegraphics[width=1.0\linewidth]{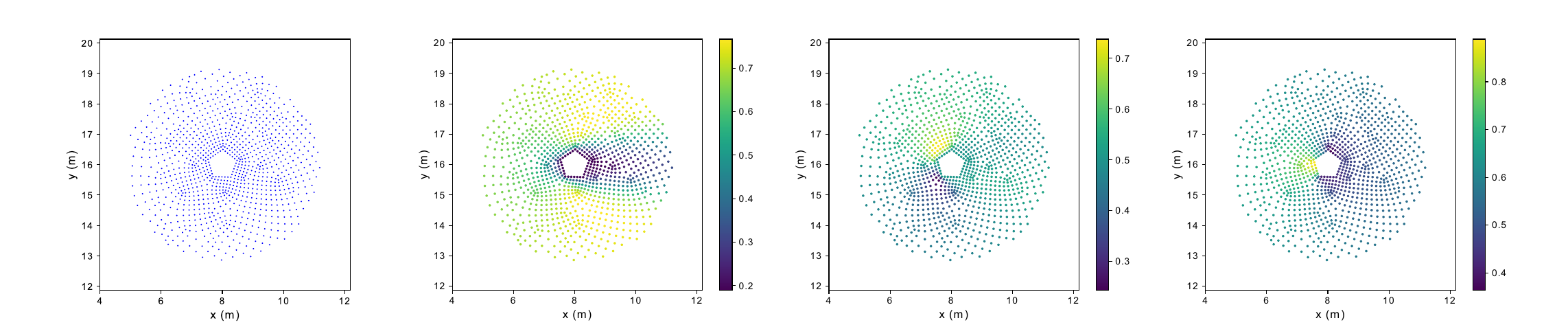}
\includegraphics[width=1.0\linewidth]{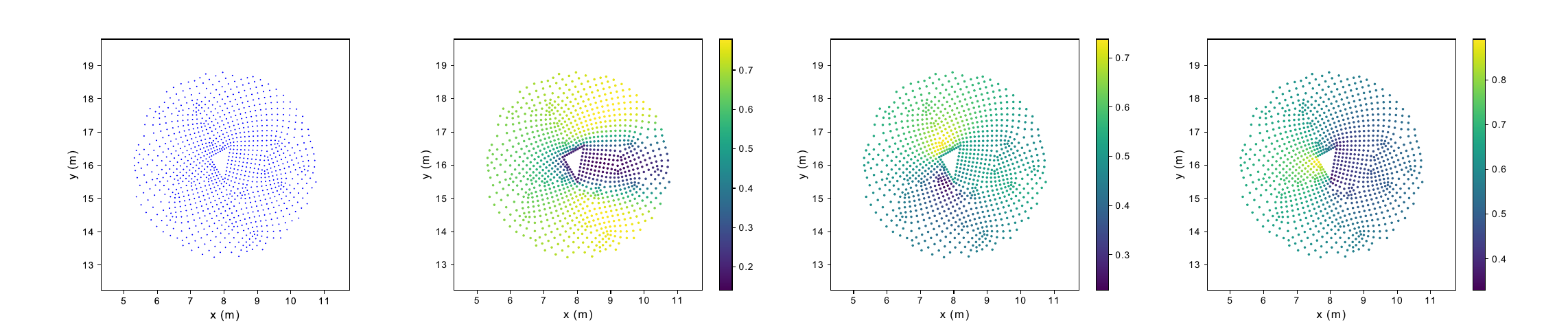}
\includegraphics[width=1.0\linewidth]{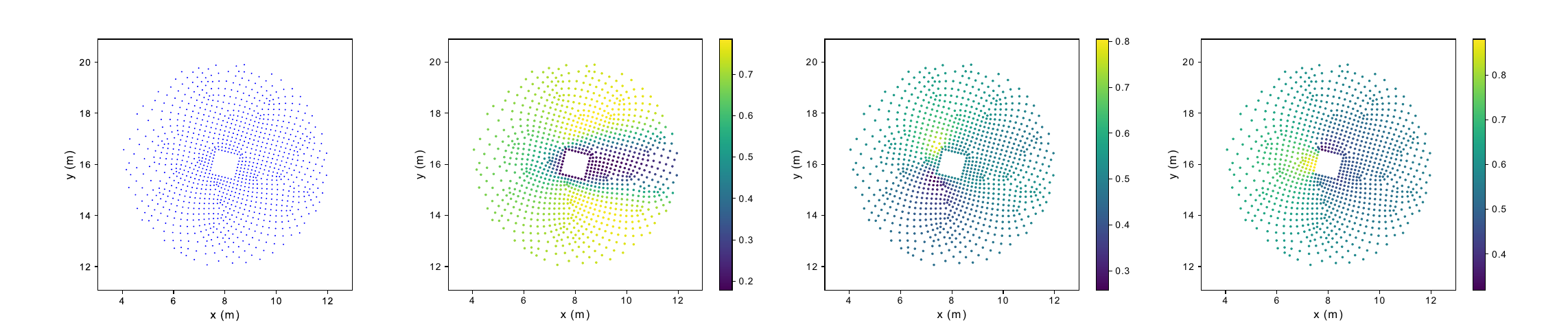}
\includegraphics[width=1.0\linewidth]{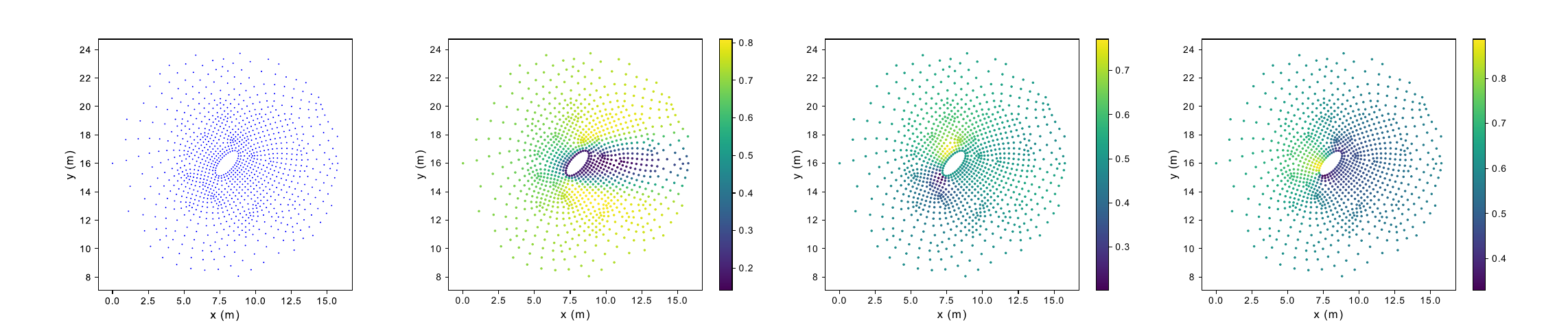}
\caption{Examples of input and output of our data set}
\label{Fig4}
\end{figure*}

We normalize the output data (i.e., $u$, $v$, and $p$) in order to speed up convergence of the network and equalize the contribution of the velocity and pressure variables to the determination of network parameters. We take two separate steps for this matter. First, we make the velocity and pressure variables dimensionless as follows:
\begin{equation}
\label{Eq7}
u^*=\frac{u}{u_\infty},
\end{equation}

\begin{equation}
\label{Eq8}
v^*=\frac{v}{u_\infty},
\end{equation}

\begin{equation}
\label{Eq9}
p^*=\frac{p-p_{0}}{\rho u_\infty^2},
\end{equation}
where $u^*$, $v^*$, and $p^*$ are the corresponding dimensionless variables and $p_{0}$ is the atmospheric pressure. Next, we scale them in a range of $[0, 1]$ by the following formulation:
\begin{equation}
\label{Eq10}
\phi^{'}=\frac{\phi - \min(\phi)}{\max(\phi) - \min(\phi)},
\end{equation}
where $\phi$ is replaced by each set of $u^*$, $v^*$, and $p^*$. Using this simple scheme, the desired normalized output data is obtained as sets of $u^{'}$, $v^{'}$, and $p^{'}$.

Alternatively, one may normalize the input data (i.e., spatial coordinates of grid points); however, our computational experiments showed that such normalization did not lead to a significant improvement in the training practice. Thus, we keep the input data as they are in the physical domain. For the generated data, the density of grid point distribution varies from one point cloud to another such that $x_{min} \in$ [0, 5.46 m], $x_{max} \in$ [10.55 m, 34.36 m], $y_{min} \in$ [0, 13.41 m], and $y_{max} \in$ [18.57 m, 32 m]. Such a variation is not practical by traditional CNNs used in the literature (see e.g., Refs. \onlinecite{sekar2019fast,guo2016convolutional,tompson2017accelerating,thuerey2019deep,bhatnagar2019prediction,jin2018prediction,zhang2018application,han2019novel,hui2020fast,hasegawa2020cnn,miyanawala2017aned,miyanawala2018bluff,mao2018cnn,bukka2020deep,miyanawala2019structure,bukka2020assessment,reddy2019reduced}). Contrarily, providing users with this flexibility is feasible by our neural network introduced in Sect. \ref{22}.

\begin{figure*}
\centering
\includegraphics[width=1.0\linewidth]{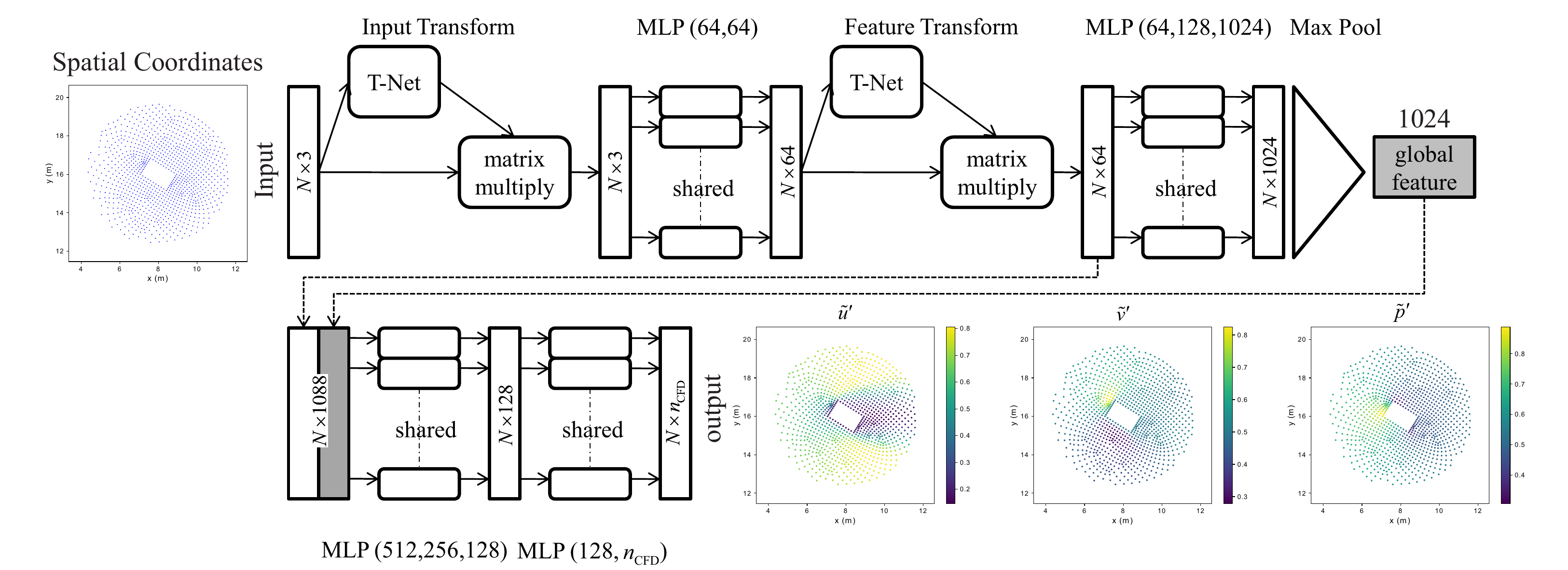}
\caption{Structure of our neural network; Labels in the format of $(A, B)$ demonstrate the size of the first layer, $A$, and the second layer, $B$, of the MLP. Labels in the form of $(A, B, C)$ are similarly classified for three layers. $n_{\text{CFD}}$ indicates the number of CFD variables; in this study, $n_{\text{CFD}}=3$. The figure shows the structure for handling three-dimensional problems; though we consider two-dimensional problems in this study.}
\label{Fig3}
\end{figure*}

\subsection{\label{22}Neural network architecture}

Generally speaking, our goal is to design a neural network that takes unstructured grid vertices inside $V_{NN}$ as point clouds and predicts the CFD quantities of interest. To accomplish this, our neural network is mainly designed according to the PointNet \cite{qi2017pointnet} architecture (see Fig. \ref{Fig3}). PointNet \cite{qi2017pointnet} was first introduced in 2017 as an effective solution to 3D computer vision tasks such as shape classification and segmentation from point clouds. It formed the foundation for numerous followup works, for example in both outdoor \cite{qi2017frustum} and indoor \cite{qi2019deep} object detection, and scene flow prediction from temporal point cloud data \cite{liu2019flownet3d}. Recent work has proposed sophisticated architectures that employ PointNet hierarchically \cite{qi2017pointnet++} or take a different approach of graph \cite{dgcnn} or continuous \cite{kpconv} convolutions on point clouds. However, PointNet \cite{qi2017pointnet} is conceptually and computationally simpler than these approaches, making it easy to implement and efficient during both training and evaluation, while still achieving competitive accuracy on shape classification and segmentation tasks \cite{dgcnn}.

In this section, we denote vectors relevant to machine learning operations with bold letters but not italic to distinguish them from the vectors used for the description of physical spaces. Mathematically, we consider a set of $N$ points of $V_{NN}$ as $\mathcal{X}=\{\mathbf{x}_i \in \mathbb{R}^d\}_{i = 1}^N$, where each point has spatial dimension $d$. We define a regression problem that seeks a function $f$ so that $f(\mathcal{X}) = \mathcal{Y}$, where $\mathcal{Y}=\{\mathbf{y}_i \in \mathbb{R}^{n_{\text{CFD}}}\}_{i = 1}^N$ is the set of desired CFD quantities of dimension $n_{\text{CFD}}$, and each output $\mathbf{y}_i$ corresponds to the input point $\mathbf{x}_i$. We propose to approximate the function $f$ with a neural network that is learned through direct supervision of CFD data. This is a challenging task since $\mathcal{X}$ is an unstructured point cloud, which does not lend itself to typical convolutional architectures that operate on structured grids. In other words, the challenge of learning on point cloud data is the need for ``permutation invariance'': the output of the neural network (e.g., the classification of a 3D object, or a feature descriptor of the shape) should not depend on the ordering of the input points. The key idea behind PointNet \cite{qi2017pointnet} is to achieve this through a symmetric function which aggregates information over points. For example, the maximum, summation, or average of a point set are all permutation invariant functions. This is specifically important in our CFD application, as the network must collect information over the set of vertices to reason about the object's shape and location which determine the flow fields. PointNet  \cite{qi2017pointnet} approximates a function $g$ on a point set by first transforming elements of the set with some function $h$, and then applying a symmetric function $s$. Mathematically, this idea is formulated as 
\begin{equation}
\label{eq:pointnet}
g(\mathcal{X}) \approx s(h(\mathbf{x}_1), \dots, h(\mathbf{x}_N)).
\end{equation}
This formulation results in a permutation invariant function with a single output. In our work, this output is a latent description of the input point cloud referred to as the ``global feature''. However, we desire an output $\mathbf{y}_i$ for every input point $\mathbf{x}_i$; for this we use the PointNet segmentation architecture \cite{qi2017pointnet} as illustrated in Fig. ~\ref{Fig3}.

Before detailing the architecture of our neural network, we briefly review the machine learning concepts leveraged in PointNet \cite{qi2017pointnet}. The point processing function $h$ is implemented with Multilayer Perceptrons (MLP). An MLP consists of a series of Fully-Connected (FC) layers, each parameterized by a weight matrix $\mathbf{W}$ and additive bias vector $\mathbf{b}$. The number of rows in $\mathbf{W}$ is the ``size'' of the layer which refers to its number of nodes. The output of each FC layer $i$ is defined by a recursive relation
\begin{equation}
\label{pointnet1}
\mathbf{a}_i = \sigma(\mathbf{W}_i\mathbf{a}_{i-1} + \mathbf{b}_i),
\end{equation}
where $\sigma$ is a nonlinear activation function that is applied elementwise and $\mathbf{a}_0 = \mathbf{x}$ is an input point. For all layers except the final output layer, we use the Rectified Linear Unit (ReLU) activation function defined as
\begin{equation}
\label{pointnet2}
\sigma(\xi) = \max(0, \xi). 
\end{equation}
In this work, layers in MLPs are separated by ``batch normalization'' which has been shown to greatly improve convergence rate during training \cite{batchnorm}. Note that in Eq.~\ref{eq:pointnet}, the same function $h$ is applied to every point, so we say that the MLP is ``shared'' across points (or that the MLPs ``share weights''). For the symmetric function $s$, we use the ``max pooling'' operation, following the strategy proposed by \citet{qi2017pointnet}. For a set of $m$-dimensional vectors $\{\mathbf{g}_i \in \mathbb{R}^m\}_{i=1}^N$, the max pooling operation is simply $\max(\mathbf{g}_1, \dots, \mathbf{g}_N) \in \mathbb{R}^m$ where the ``$\max$'' function is applied elementwise.

We use the PointNet segmentation \cite{qi2017pointnet} architecture depicted in Fig. ~\ref{Fig3}, which can be broken into the top and bottom parts. The top part implements Eq.~\ref{eq:pointnet} with the elements described previously. In particular, it takes the input point set $\mathcal{X}$ and outputs the global feature of dimension 1024. It operates on each input point independently using two shared MLPs - the first has two FC layers of size (64, 64) and the second with three layers of size (64, 128, 1024) - followed by max pooling to arrive at the global feature. Note that labels in the form of $(A, B)$ indicate the size of the first layer, $A$, and the second layer, $B$, of the associated machine learning component. Similarly, labels in the form of $(A, B, C)$ are defined for three layers. The input points and intermediate features (64 dimensional) undergo affine transformations predicted by a set of Transformation Networks (T-Nets). 
The T-Nets are themselves mini PointNets: they use a small shared three-layer MLP of size (64, 128, 1024) and max pooling, followed by an additional two-layer MLP of size (512, 256) to predict a transformation matrix. For a more detailed description of T-Net implemented here, one may refer to Ref. \onlinecite{qi2017pointnet}.

The global feature aggregates information over all input points, but we need an individual output corresponding to each input point. This is accomplished with the bottom part of the architecture depicted in Fig. \ref{Fig3}. Namely, intermediate features of size 64 are taken from the top part of the neural network, and each is concatenated to a copy of the final global feature. This provides $N$ features of dimension 1088. The concatenated feature goes through a final shared MLP to produce the regressed CFD quantities of size $N \times n_{\text{CFD}}$. As discussed in Sect. \ref{23}, the output data is normalized in the range of $[0, 1]$. Thus, a sigmoid activation function, expressed as
\begin{equation}
\label{pointnet3}
\sigma(\xi) = \frac{1}{1 +e^{-\xi}},
\end{equation}
is used in the output layer to cover the corresponding range. Note that one could employ separate bottom branches to decode the global feature for each CFD quantity, however, it imposes extra computational costs.

As discussed in Sect. \ref{21}, the solution of the Navier-Stokes and continuity equations (see Eqs. \ref{Eq1}--\ref{Eq2}) is a function of the geometry of the cylinder cross section. With this in mind, we devise the deep learning environment such that the point set ``implicitly'' defines the shape of cross sections through its null space. In fact, the cylinder object is contained where there are not points (see e.g., Fig. \ref{Fig4}). We expect the network to learn this function during training. The strategy of implicit shape representation becomes more crucial when a point cloud represents multiple bodies inside $V_{NN}$. This paradigm is discussed further in Sect. \ref{33}.

Note that our neural network is designed for handling three-dimensional $(d=3)$ fluid dynamics problems, however, we restrict our attention to two-dimensional $(d=2)$ applications in this article. Furthermore, we are interested in predicting the velocity ($u$ and $v$) and pressure ($p$) fields and thus we set $n_{\text{CFD}}=3$ in this study, though our formulation is general enough to encompass other CFD variables such as temperature, for instance, when the energy equation is involved.

\begin{figure*}
\centering
\includegraphics[width=1.0\linewidth]{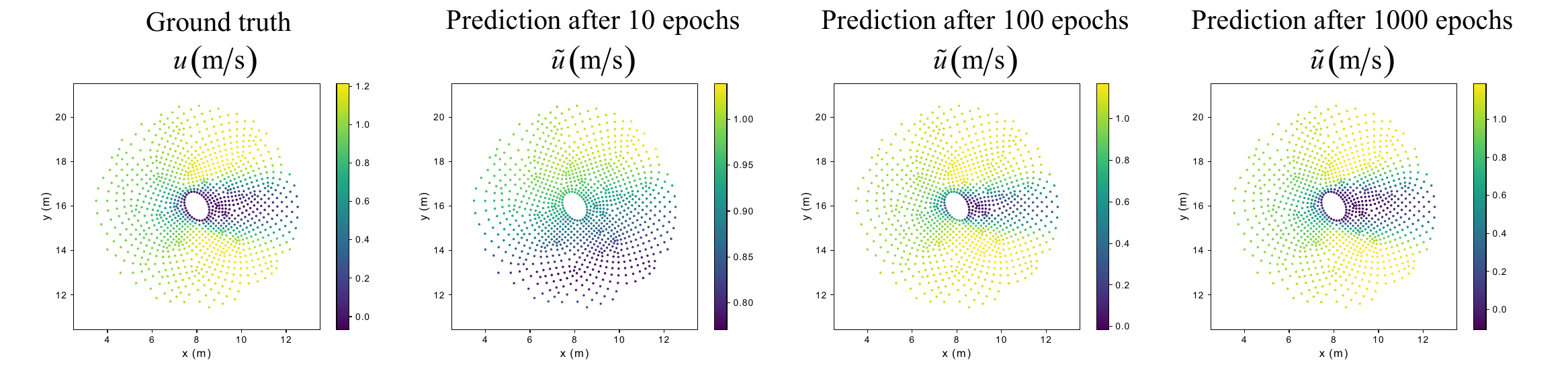}
\includegraphics[width=1.0\linewidth]{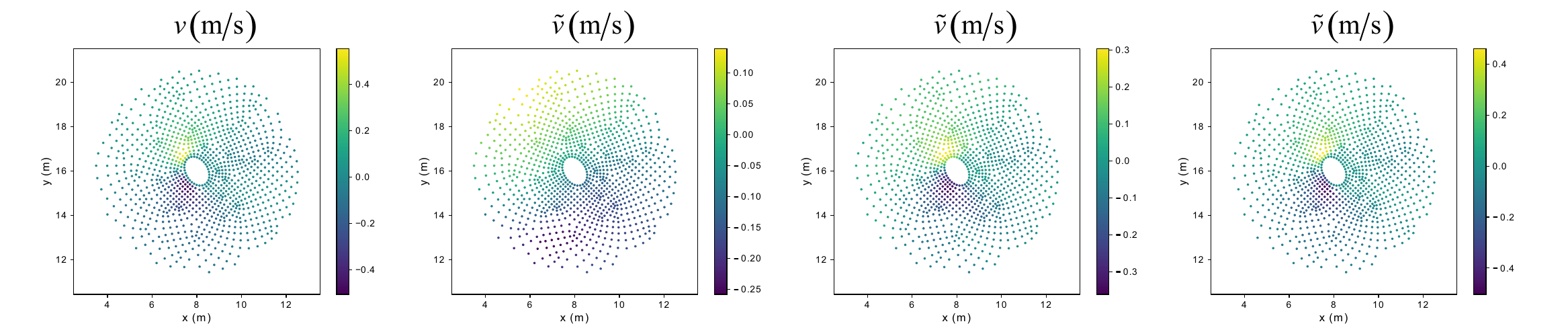}
\includegraphics[width=1.0\linewidth]{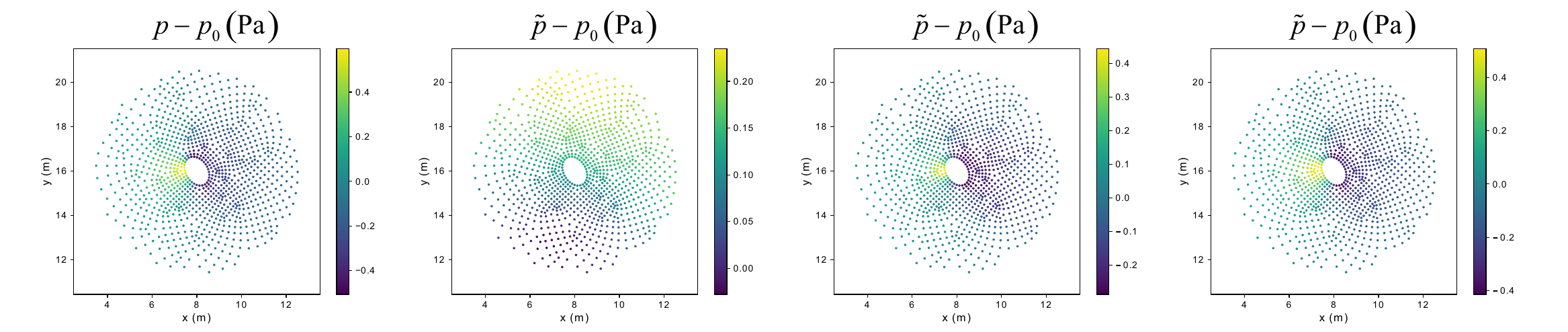}
\caption{A comparison between the ground truth and prediction of the network for the velocity and pressure fields after 10, 100, and 1000 epochs}
\label{Fig6}
\end{figure*}

\subsection{\label{24}Training}

As can be realized from the literature (see e.g., Refs. \onlinecite{sekar2019fast, bhatnagar2019prediction}), the selection of “mean squared error” as the loss function ($\mathcal{L}$) is a reliable norm in deep learning applications of computational mechanics. Hence, we use this function defined as
 
\begin{equation}
\label{Eq11}
\mathcal{L}=\frac{1}{3\times N}\left (\sum_{i=1}^{N} \left [(u_i^{'}-\tilde{u}_i^{'})^2 + (v_i^{'}-\tilde{v}_i^{'})^2 + (p_i^{'}-\tilde{p}_i^{'})^2 \right] \right).
\end{equation}
As mentioned in Sect. \ref{23}, $u_i^{'}$, $v_i^{'}$, and $p_i^{'}$ are respectively the ground truth velocity and pressure data obtained from the CFD solver. $\tilde{u}_i^{'}$, $\tilde{v}_i^{'}$, and $\tilde{p}_i^{'}$ are respectively the velocity and pressure fields predicted by our neural network. We scale the predicted variables back into the physical space and indicate them by $\tilde{u}$, $\tilde{v}$, and $\tilde{p}$. The Adam optimizer \cite{kingma2014adam} is used with the learning rate of $\alpha=5\times10^{-4}$ and the hyper-parameters of $\beta_1=0.9$, $\beta_2=0.999$, and $\hat{\epsilon}=10^{-6}$ (see Ref. \onlinecite{kingma2014adam} for the mathematical definition of $\alpha$, $\beta_1$, $\beta_2$, and $\hat{\epsilon}$). Furthermore, the batch size of 256 is chosen. As explained in Sect. \ref{22}, 2595 data are generated in total. Figure \ref{Fig4} exhibits a few examples of the input and output data. We categorize the generated data into three sets of training (80\%), validation (10\%), and test (10\%) through a random process. Thus, we train the neural network over 2076 data. 260 data are used for the validation and the remaining 259 data (as unseen data) are reserved for evaluation. After each epoch, the network is validated over the validation data set for tracing the convergence rate and avoiding over-fitting. The training procedure is stopped after 4000 epochs. Training 3552588 parameters is performed on an NVIDIA TITAN Xp graphics card with the memory clock rate of 1.582 GHz and 12 Gigabytes of RAM and approximately takes 10 hours.

We listed the values assigned to the hyperparameters of our neural network in the previous paragraph. To determine these hyperparameters, a systematic procedure is carried out. For brevity, we only present one important portion of this process: selection of the batch size and the size of the latent global feature (see Fig. \ref{Fig3}) through a grid search. Mini batch gradient descent techniques allow us to accelerate training by vectorization (in contrast with stochastic gradient descent) and make training feasible by grouping data into small mini batches (in contrast with full batch gradient descent) when a memory limitation exists. However, the chosen size of the mini batch impacts network performance, including its ability to generalize.
On the other hand, the size of the global feature affects the representational capacity of the network as discussed in Sect. \ref{22}. The result of the investigation is tabulated in Table \ref{Tab1prime}. Note that a change in the size of the global feature is followed by an adjustment in the size of the MLP directly following the global feature in the network (see Fig. \ref{Fig3}). This is to ensure that the global feature is the main information bottleneck in the network. As can be realized from Table \ref{Tab1prime}, according to the test loss ($\mathcal{L}$) the optimal choice is the global feature size of 1024 with the batch size of 256. There are several general notes here. First, it is conjectured that the global feature size of 2048 with the batch size of 256 would lead to a higher performance; however, our computational facilities limited our choices. Second, one may investigate the degree of “bias” and “variance” respectively by measuring the distance between the training loss ($\mathcal{L}$) and validation loss ($\mathcal{L}$) and the distance between the validation loss ($\mathcal{L}$) and the test loss ($\mathcal{L}$) for all the cases presented in Table \ref{Tab1prime}. Particularly, we observe low bias and low variance for our choice. Third, we list the training time for each case in Table \ref{Tab1prime} as well, since it could be another restriction for potential users.

\begin{table*}[ht!]

\caption{\label{Tab1prime}Training, validation, and test loss ($\mathcal{L}$) as well as training time for different batch sizes and sizes of the global feature; MLP size shows the size of different layers of the MLP right after the global feature in the network (see Fig. \ref{Fig3}). The cross symbol ($\times$) indicates that the training is not doable due to the lack of sufficient memory.}
\begin{ruledtabular}
\begin{tabular}{lllll}
 global feature size & MLP size & batch size of 64 & batch size of 128 & batch size of 256\\
\hline
\multirow{4}{*}{256} & \multirow{4}{*}{(256,256,128)}  & training time =  24194 s & training time = 24418 s & training time = 24230 s\\
& & training loss = 4.71790E$-$5 & training loss = 4.47410E$-$5 & training loss = 7.91800E$-$5 \\
& & validation loss = 3.60867E$-$4 & validation loss = 3.13515E$-$4 & validation loss = 3.96315E$-$4  \\
& & test loss = 1.62594E$-$3 & test loss = 1.57643E$-$3 & test loss = 1.63278E$-$3 \\
\hline
\multirow{4}{*}{512} & \multirow{4}{*}{(256,256,128)}  & training time = 27666 s & training time = 27866 s & training time = 27699 s\\
& & training loss = 3.96650E$-$5
 & training loss = 4.96650E$-$5 & training loss = 8.11420E$-$5 \\
& & validation loss = 3.40380E$-$4
 & validation loss = 3.16175E$-$4 & validation loss = 3.99710E$-$4 \\
& & test loss = 1.41908E$-$3 & test loss = 1.40274E$-$3 & test loss = 1.63381E$-$3 \\
\hline
\multirow{4}{*}{1024} & \multirow{4}{*}{(512,256,128)}  & training time = 24044 s & training time = 23382 s & training time =  23169 s\\
& & training loss = 3.0541E$-$5 & training loss = 4.08800E$-$5  & training loss =  6.89100E$-$5 \\
& & validation loss = 5.43850E$-$4 & validation loss = 4.81712E$-$4 & validation loss =  3.83702E$-$4 \\
& & test loss = 2.22772E$-$3 & test loss = 1.97172E$-$3 & test loss = 1.37766E$-$3 \\
\hline
\multirow{4}{*}{2048} & \multirow{4}{*}{(512,256,128)}  & training time = 29459 s & training time = 29173 s & training time = $\times$ \\
& & training loss = 2.58860E$-$5 & training loss = 4.30400E$-$5  & training loss = $\times$ \\
& & validation loss = 3.69553E$-$4 & validation loss = 4.97283E$-$4 & validation loss = $\times$\\
& & test loss = 1.50372E$-$3 & test loss = 2.02906E$-$3 & test loss =  $\times$\\
\end{tabular}
\end{ruledtabular}
\end{table*}

Figure \ref{Fig6} shows the velocity and pressure fields obtained by the CFD solver along with those predicted by the network for a geometry belonging to the test set after 10, 100, and 1000 epochs. As can be seen from Fig. \ref{Fig6}, the network prediction is inaccurate after 10 epochs. After 100 epochs, although the general structure of the flow fields predicted by the network is in a right configuration, the homogeneous velocity Dirichlet boundary condition on the object surface is not yet satisfied. Furthermore, the predicted values for the space of velocity and pressure fields are not in a correct range. After 1000 epochs, the latter issue is almost resolved, however, the predicted flow still slips on the body surface (see e.g., $\tilde{v}$ in Fig. \ref{Fig6}). In this sense, the most time consuming component of the training process for the network is to learn the no-slip condition on the edges of bodies. This fact demonstrates the main difference between training a network for the prediction of viscous flows versus inviscid flows (when the flow is permitted to artificially slip on surfaces). Certainly, this is one of the core challenges of designing networks for prediction of realistic flow fields versus artificial flow fields used largely for visual effects (see e.g., Ref. \onlinecite{kim2019deep}).

\begin{figure*}
\centering
\includegraphics[width=1.0\linewidth]{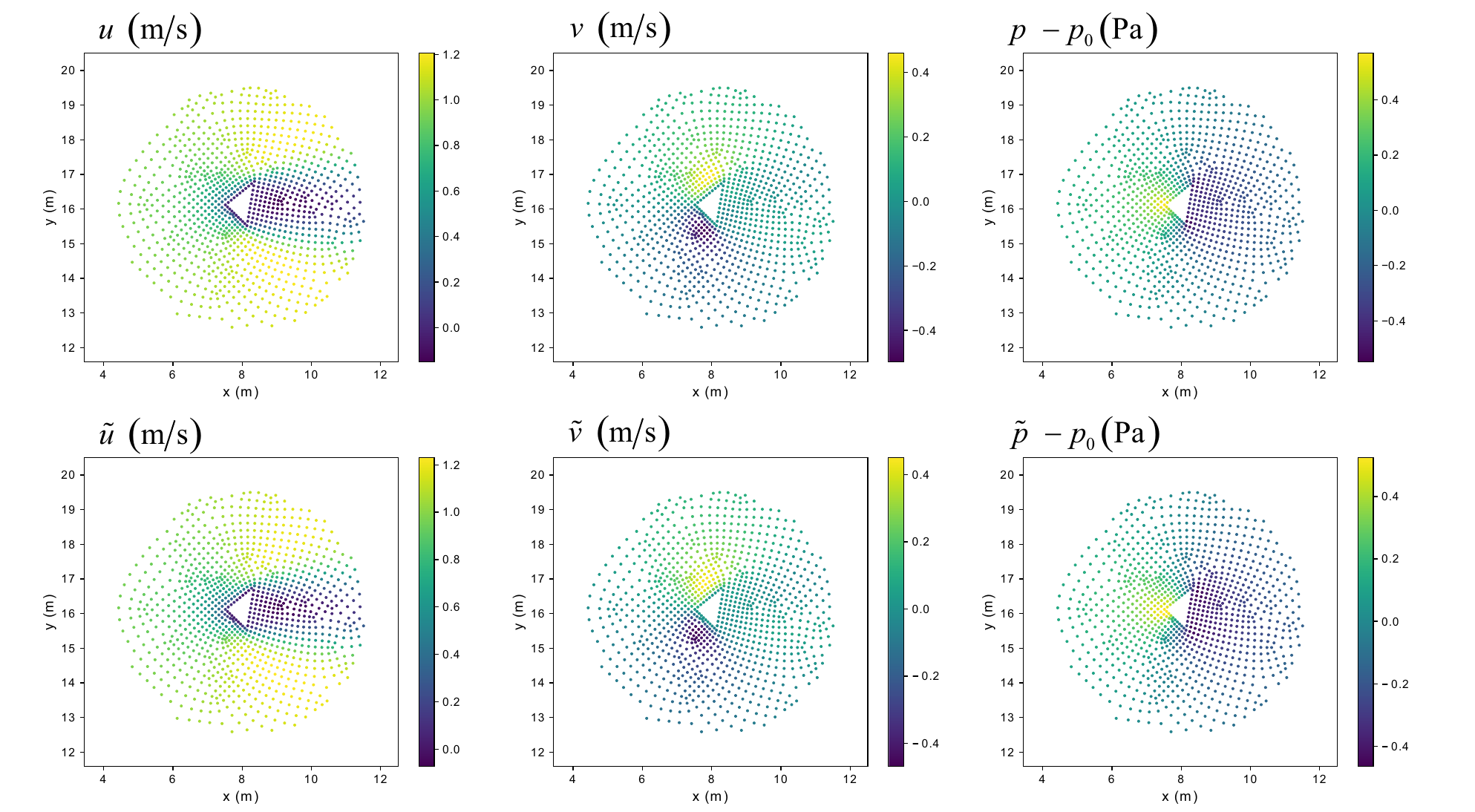}
\includegraphics[width=1.0\linewidth]{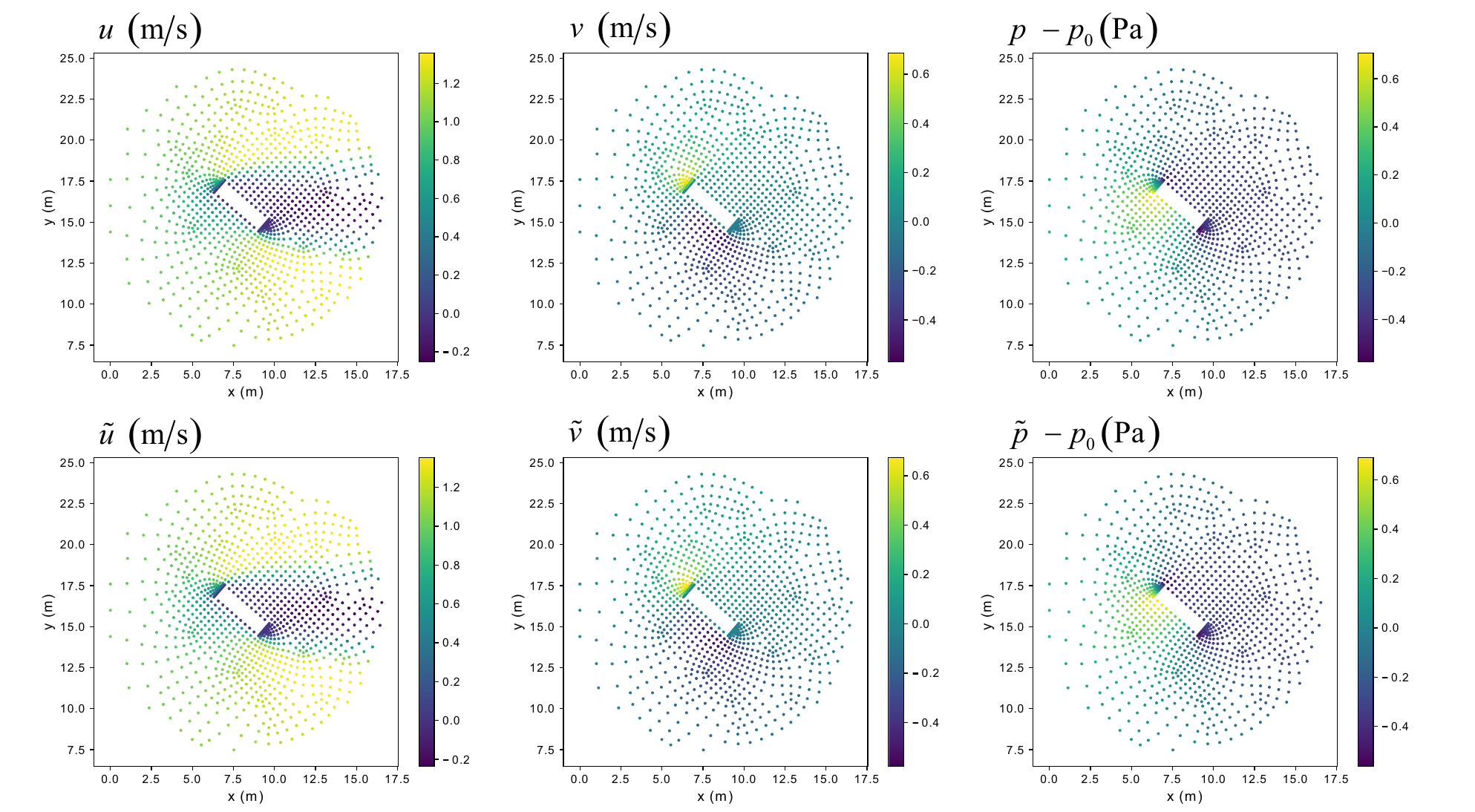}
\caption{The first set of examples comparing the ground truth to our network prediction for the velocity and pressure fields of two different cross sections taken from the test set}
\label{Fig71}
\end{figure*}

\begin{figure*}
\centering
\includegraphics[width=1.0\linewidth]{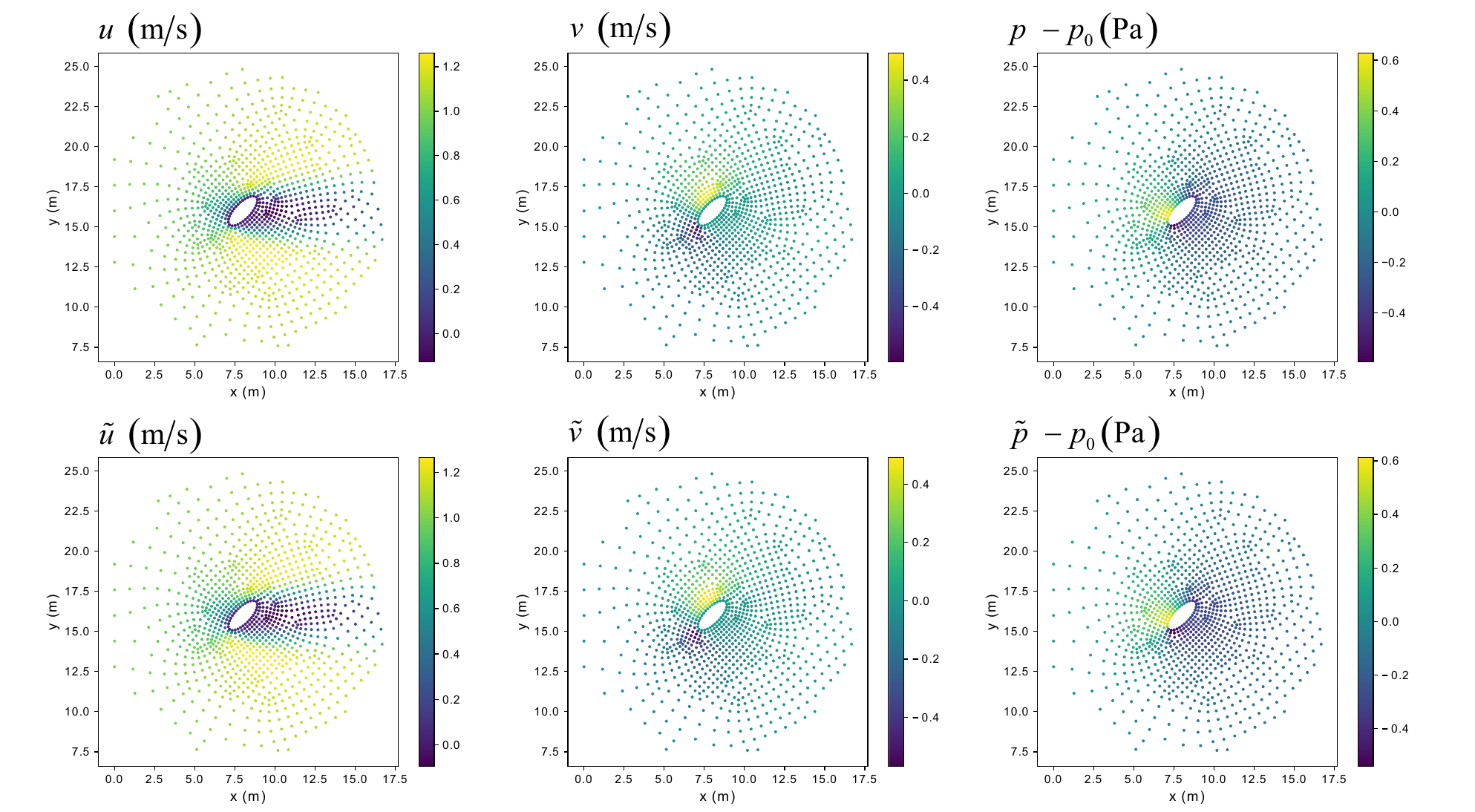}
\includegraphics[width=1.0\linewidth]{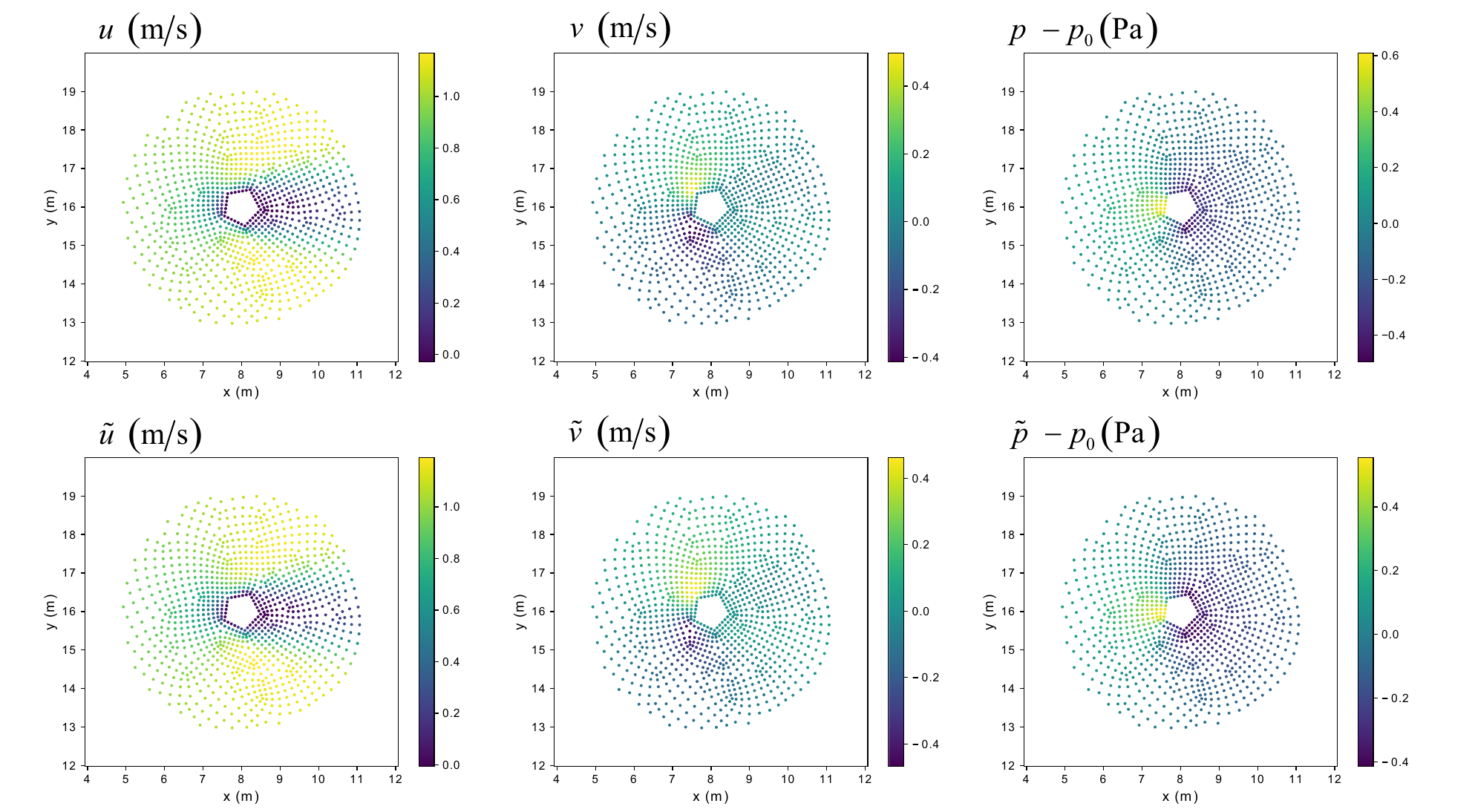}
\caption{The second set of examples comparing the ground truth to our network prediction for the velocity and pressure fields of two different cross sections taken from the test set}
\label{Fig72}
\end{figure*}

\begin{figure*}
\centering
\includegraphics[width=1.0\linewidth]{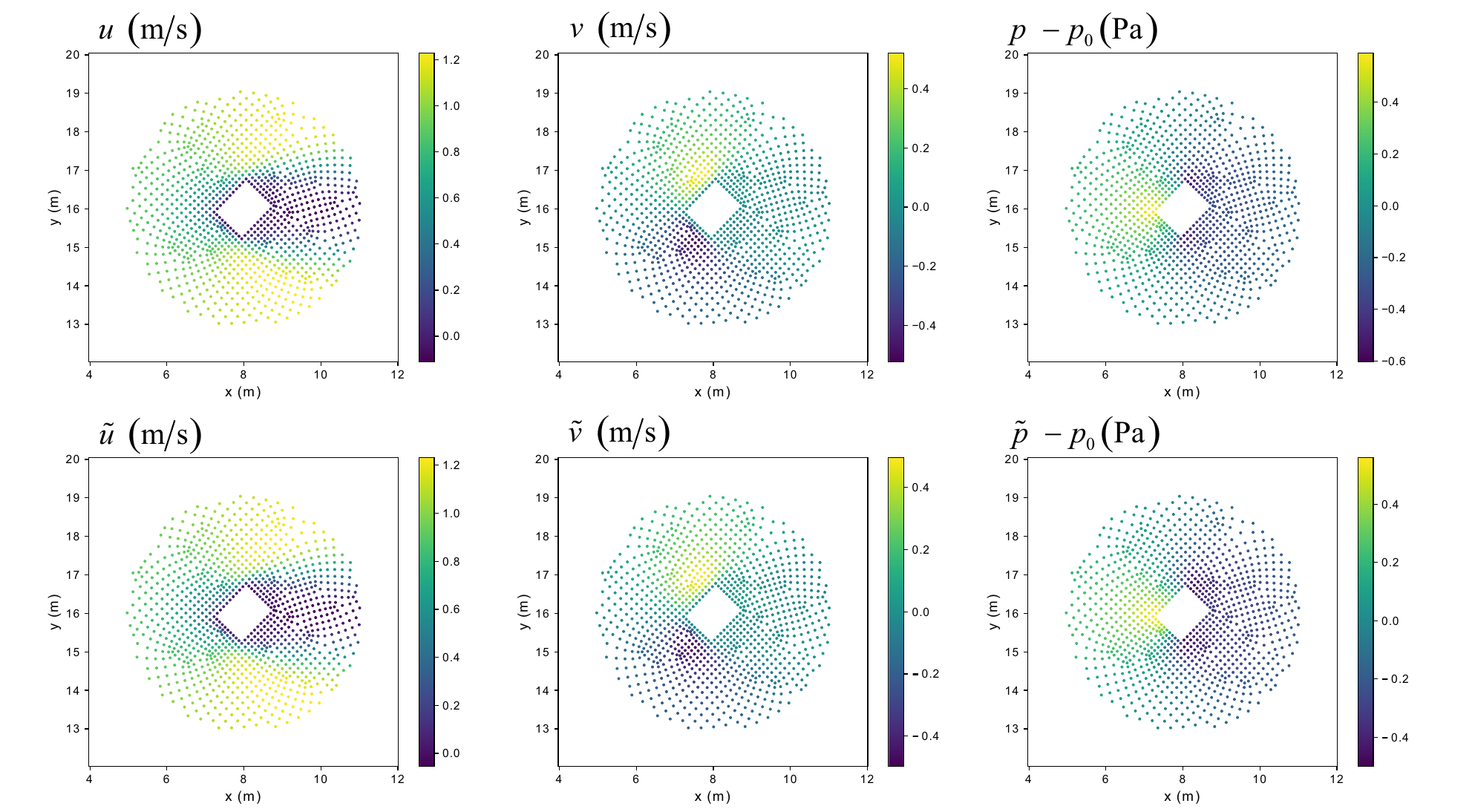}
\includegraphics[width=1.0\linewidth]{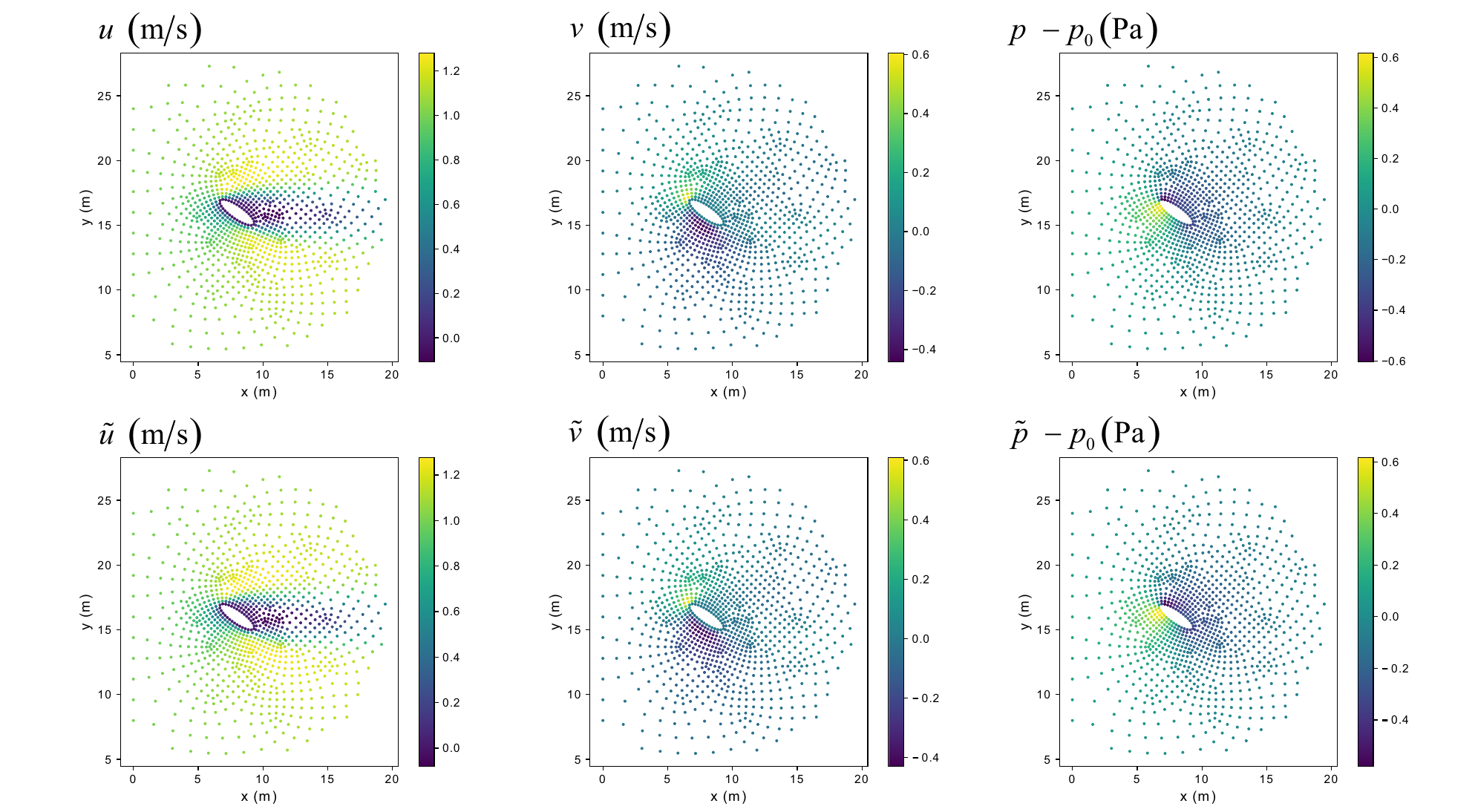}
\caption{The third set of examples comparing the ground truth to our network prediction for the velocity and pressure fields of two different cross sections taken from the test set}
\label{Fig73}
\end{figure*}

\begin{figure*}
\centering
\includegraphics[width=1.0\linewidth]{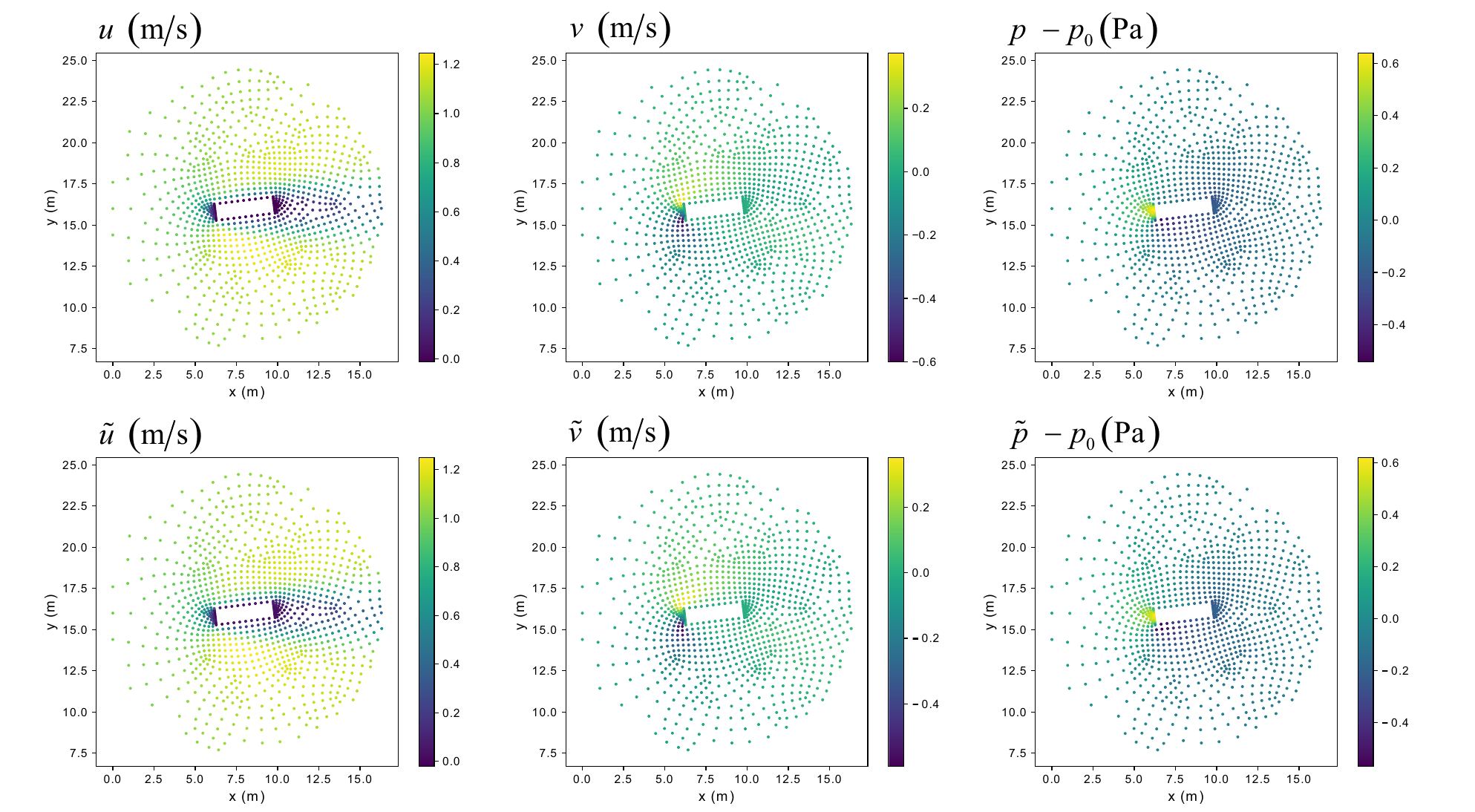}
\includegraphics[width=1.0\linewidth]{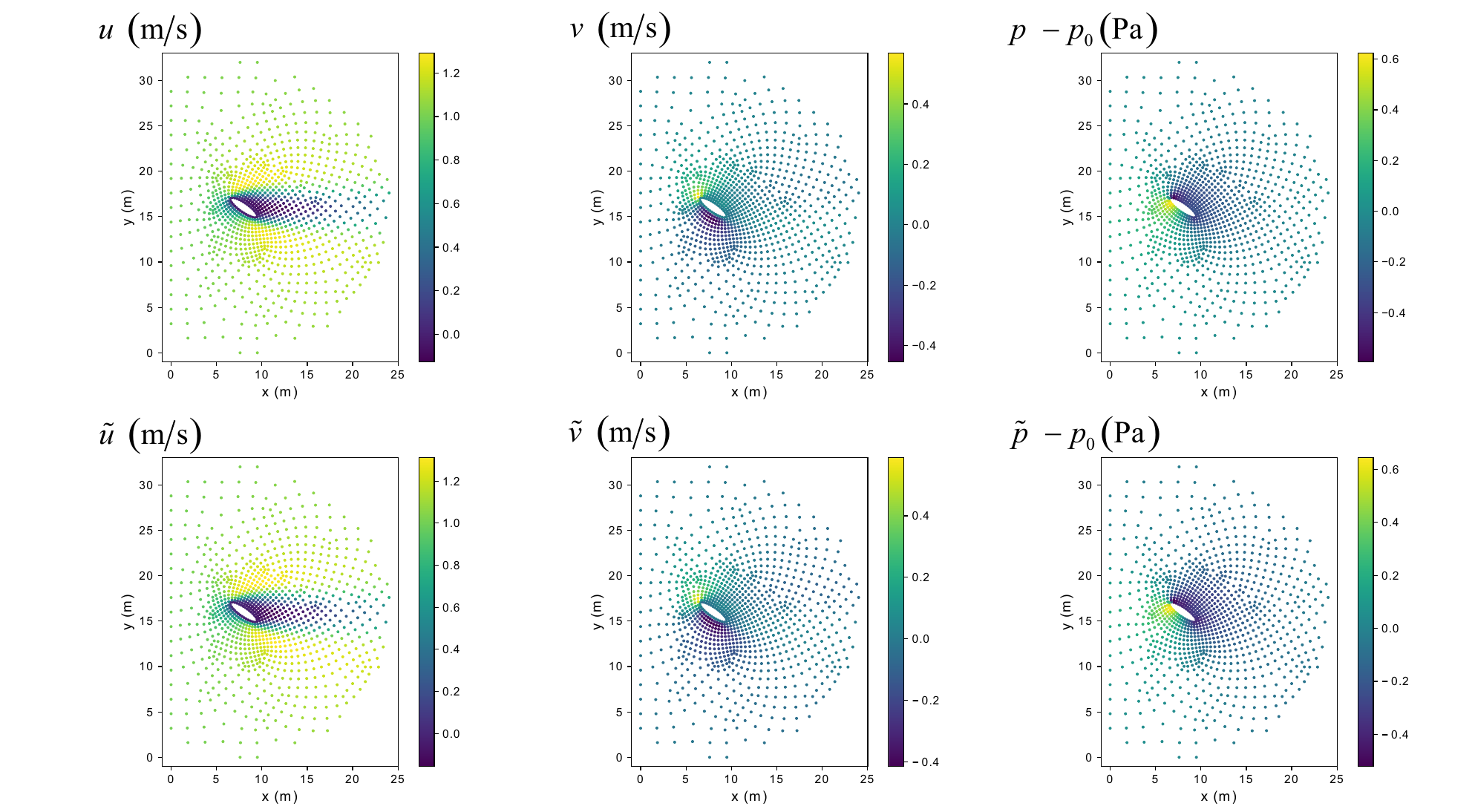}
\caption{The fourth set of examples comparing the ground truth to our network prediction for the velocity and pressure fields of two different cross sections taken from the test set}
\label{Fig74}
\end{figure*}

\section{\label{ResultsAndDiscussion}Results and discussion}

\subsection{\label{31}General analysis}

Figures \ref{Fig71}--\ref{Fig74} illustrate a comparison between the ground truth and our network prediction for the velocity and pressure spaces for eight examples chosen from the test set. These examples display objects from various object classes with different sizes and orientations, and thus different length scales. It is worthwhile to note that our network successfully predicts the flow separation phenomenon (see e.g., the $u$ velocity field for the rectangular cross section in Fig. \ref{Fig71}). Additionally, the input points shown in Figs. \ref{Fig71}--\ref{Fig74} have varying spatial distribution densities, which demonstrates the flexibility and generality of our neural network. The maximum, minimum, and average pointwise error ($L^2$ norm) for the test set are tabulated in Table. \ref{Table2}. Concerning the averaged error (see the first row of Table \ref{Table2}), the largest loss belongs to the $x$-component of velocity ($\tilde{u}$) simply because this is a more complicated field in comparison with the $y$-component of velocity ($\tilde{v}$) and pressure ($\tilde{p}$) fields in the current regime (see e.g., Fig. \ref{Fig2}). On the other hand, the smallest loss occurs in the pressure ($\tilde{p}$) field since no boundary condition is imposed on the pressure space in the domain of interest $V_{NN}$. Recall that $\Gamma_D$ does not overlap $\Gamma_N$. The role of boundary conditions in the learning procedure, and consequently the loss, can be observed by this example. For instance, it is conjectured if an open boundary condition (see e.g., Eq. \ref{Eq5}) had to be satisfied in an assumed $V_{NN}$, the largest level of loss would belong to the pressure field.

\begin{table}
\caption{\label{Table2}Error analysis of the velocity and pressure fields predicted by our neural network for 259 unseen data; $||\ldots||$ indicates the $L^2$ norm.}
\begin{ruledtabular}
\begin{tabular}{llll}
 & $||u-\tilde{u}||$ & $||v-\tilde{v}||$ & $||p-\tilde{p}||$\\
\hline
Average & 4.49666E$-$2 & 3.70540E$-$2 & 2.71661E$-$2 \\
Maximum & 2.49088E$-$1 & 2.34281E$-$1 & 1.16901E$-$1 \\
Minimum & 1.10453E$-$2 & 9.20977E$-$3 & 7.58447E$-$3\\
\end{tabular}
\end{ruledtabular}
\end{table}

\begin{figure*}
\centering
\includegraphics[width=1.0\linewidth]{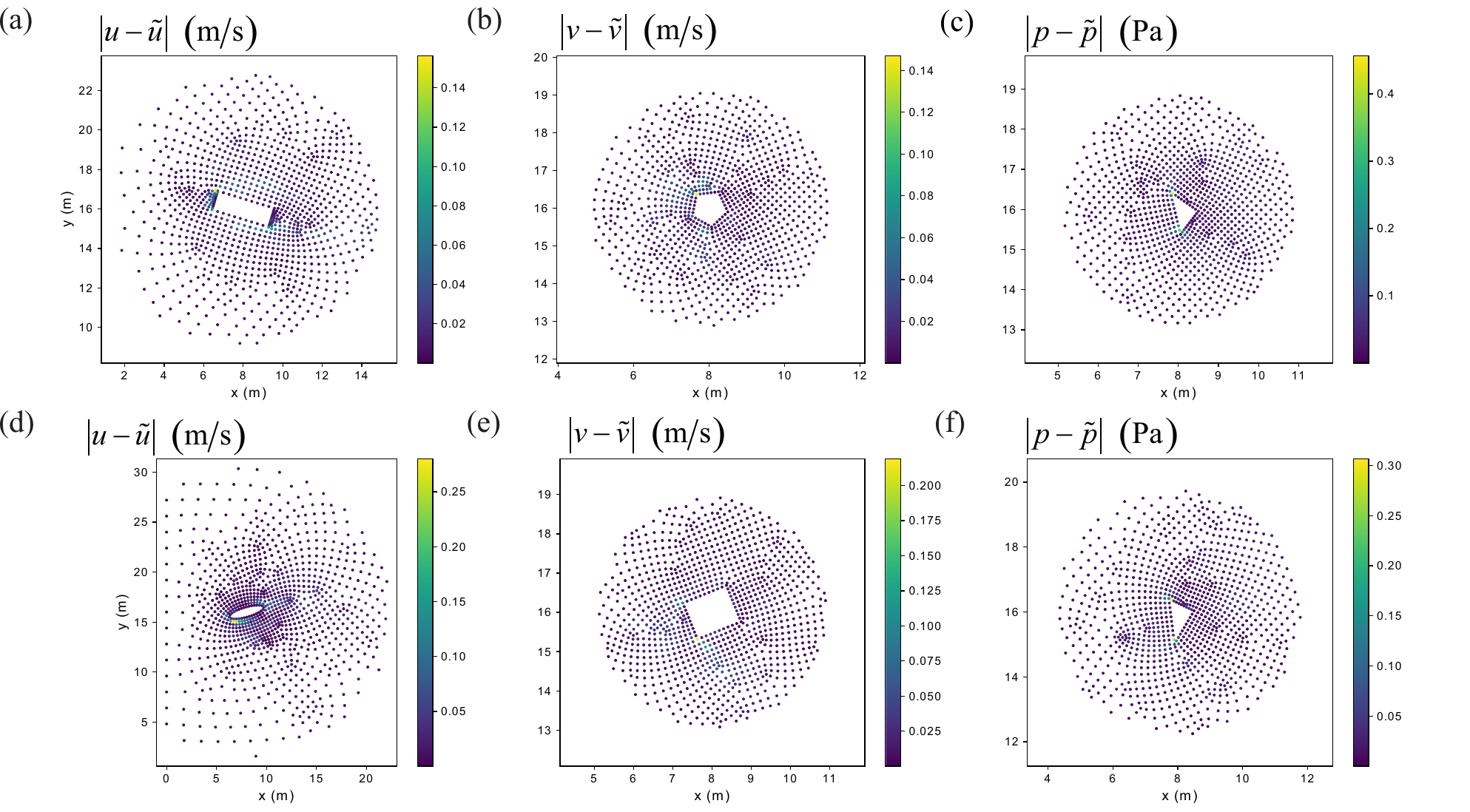}
\caption{Distribution of absolute pointwise error when the mean square error becomes \textbf{(a)} maximum for $\tilde{u}$,  \textbf{(b)} maximum for $\tilde{v}$, \textbf{(c)} maximum for $\tilde{p}$, \textbf{(d)} minimum for $\tilde{u}$, \textbf{(e)} minimum for $\tilde{v}$, and \textbf{(f)} minimum for $\tilde{p}$. }
\label{Fig8}
\end{figure*}

Figure \ref{Fig8} presents the pointwise error distribution along the point cloud for geometries with the maximum or minimum pointwise error ($L^2$ norm) among the test set. By looking at Fig. \ref{Fig8}, it is apparent that the minimum or maximum loss of the predicted velocity and pressure fields does not belong to a specific object, indicating that our network is not overfitted to a particular class of object. Interestingly, the maximum loss does not happen in grids with the highest variation in spatial distribution density.

For all the cases shown in Fig. \ref{Fig8}, the maximum pointwise error ($L^2$ norm) takes place on the edges of objects, where the no slip condition has to be applied. As discussed in Sect. \ref{24}, the most demanding fraction of training is to learn zero velocity Dirichlet Boundary conditions at solid walls, which is due to two main reasons. First, there is a relatively large jump in $u'$ and $v'$ when we move from the wall surfaces toward the inside of the active space. To defend this reasoning, we perform a simple experiment: instead of imposing zero velocity boundary conditions on the object surfaces, we extrapolate the velocity data from inside the domain into the object surfaces, mimicking inviscid flow conditions. In this case, the maximum pointwise error ($L^2$ norm) computed for $\tilde{u}$ reduces by 34\% compared to the viscous flow assumption, indicating a more straightforward process of network training for inviscid flows. Second, the point distribution differs the most across different objects at this boundary.

Concerning the influence of data normalization, scaling the output data in the range of $[-1,1]$ (instead of $[0,1]$) would increase the average pointwise error ($L^2$ norm) roughly by a factor of 2 for the test set based on our numerical experiments.

The last component of our general analysis addresses the speedup factor obtained by means of the neural network introduced in this study. The average wall time consumed for prediction of the 259 unseen data in the test set using our network is approximately 6 seconds. Simulation of the flow fields for these 259 geometries takes approximately 11071 seconds (about 3 hours) using the CFD software on a single Intel(R) Core(TM) processor with a 2.30 GHz clock rate. Thus, the average achieved speedup using our neural network is a factor of approximately 1846 in the framework of our available computational facilities. Note that the speedup factor is strongly a function of the CFD solver efficiency and the types of Central Processing Unit (CPU) and Graphics Processing Unit (GPU) used. Moreover, we expect achieving higher factors of acceleration for turbulent flow regimes.

\subsection{\label{32}Investigation of conservation of mass and momentum}

So far we have examined the efficiency of our neural network in predicting the velocity and pressure fields by the metric of pointwise error ($L^2$ norm). To more precisely analyze the network performance, we assess the capacity of our network to conserve mass and momentum. This assessment has two valuable advantages. First, instead of analyzing the error of each predicted field individually, the examination of conservation of mass and momentum offers the error analysis of all the predicted spaces jointly. Second, since the point cloud distribution is non-uniform, the error of each point is weighted by the volume of its corresponded cell. For this purpose, we define three residuals for steady-state flows according to Eqs. \ref{Eq1}--\ref{Eq2} as follow:
\begin{align}
\label{Eq12}
 & r_{momentum_x}=  \\& \left| \int_{V_{NN}} \left(\rho \left(\tilde{u}\frac{\partial  \tilde{u}}{\partial x} + \tilde{v}\frac{\partial \tilde{u}}{\partial y}\right) + \frac{\partial \tilde{p}}{\partial x} - \mu \left(\frac{\partial^2 \tilde{u}}{\partial x^2} + \frac{\partial^2 \tilde{u}}{\partial y^2}\right)\right)dV \right|, \nonumber
\end{align}

\begin{align}
\label{Eq13}
&r_{momentum_y}= \\ &\left| \int_{V_{NN}} \left(\rho \left(\tilde{u}\frac{\partial \tilde{v}}{\partial x} + \tilde{v}\frac{\partial \tilde{v}}{\partial y}\right) + \frac{\partial \tilde{p}}{\partial y} -\mu \left(\frac{\partial^2 \tilde{v}}{\partial x^2} + \frac{\partial^2 \tilde{v}}{\partial y^2}\right)\right) dV\right|, \nonumber
\end{align}

\begin{equation}
\label{Eq14}
r_{continuity}=\left| \int_{V_{NN}} \left(\frac{\partial \tilde{u}}{\partial x} + \frac{\partial \tilde{v}}{\partial y}\right) dV \right|,  
\end{equation}
where $r_{momentum_x}$ and $r_{momentum_y}$ are respectively the residuals of momentum (Navier-Stokes) equations (see Eq. \ref{Eq1}) in the $x$ and $y$ directions. Similarly, $r_{continuity}$ indicates the residual of the continuity equation (see Eq. \ref{Eq2}) for incompressible flows. The absolute value is shown by $|\dots|$.

To compute the residuals, a first order finite volume method is used to numerically approximate the spatial derivatives in Eqs. (\ref{Eq12})--(\ref{Eq14}). Note that the input of our network is finite volume grids and information such as cell connectivity is available. Although our network does not use this information, we can put the predicted velocity and pressure fields back into the finite volume mesh for a post-processing analysis. The results are shown in Table \ref{Tab3}. Based on the data collected in Table \ref{Tab3}, the neural network conserves the mass and momentum of the flow fields with an excellent to reasonable level of accuracy. The maximum residual occurs for $r_{momentum_x}$. This trend is expected because the maximum loss also occurs for $\tilde{u}$ as tabulated in Table \ref{Table2}. However, the maximum or minimum $r_{momentum_x}$ and $r_{momentum_y}$ does not belong to the point clouds with the maximum or minimum loss in $\tilde{u}$ and $\tilde{v}$, respectively. Furthermore, the maximum or minimum $r_{momentum_x}$, $r_{momentum_y}$, and $r_{continuity}$ happens in different point clouds. In general, $r_{continuity}$ is greater than $r_{momentum_y}$ because mathematically $\tilde{u}$ plays a more significant role in Eq. \ref{Eq14} in comparison with Eq. \ref{Eq13}.

It is worthwhile to note that the error analysis presented in this subsection is closely connected to the idea of semi-supervised learning of the continuity and Navier-Stokes equations. This idea forms one of our future studies. We elaborate this concept further in Sect. \ref{Conclusions}, when we discuss the physics-informed neural networks.

\begin{table}[ht!]
\caption{\label{Tab3}Investigation of conservation of mass and momentum of the flow fields predicted by our neural network for 259 unseen data. All values are reported in the International Unit System.}
\begin{ruledtabular}
\begin{tabular}{llll}
 & $r_{momentum_x}$ & $r_{momentum_y}$ & $r_{continuity}$\\
\hline
Average &  4.14958E$-$3 & 2.46155E$-$3 &  2.99411E$-$3\\
Maximum &  3.38842E$-$1 & 3.59399E$-$2 &  8.74008E$-$2\\
Minimum &  5.69245E$-$6 & 9.03372E$-$7 &  3.58928E$-$6\\
\end{tabular}
\end{ruledtabular}
\end{table}

\subsection{\label{33prime}Physical interpretation}

The concern of this subsection is to explore how different layers of our neural network are connected to the physics of problem. Because we supervise the network output directly with the desired CFD quantities during training, we expect that intermediate layers of the network learn to represent the underlying physics in an explicit or implicit form.

\begin{figure*}
\centering
\subfigure{
\includegraphics[width=.3\textwidth]{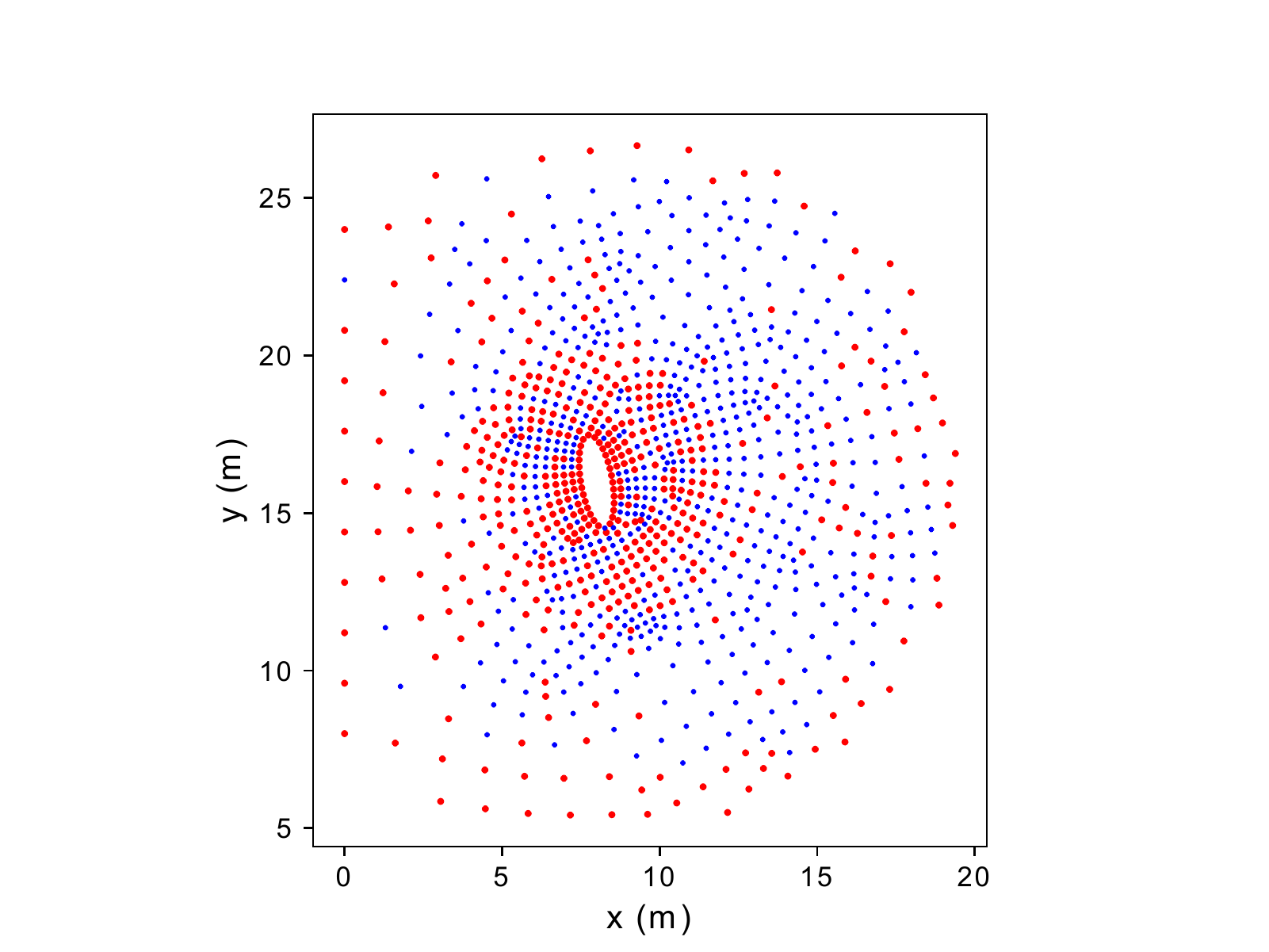}
}
\subfigure{
\includegraphics[width=.3\textwidth]{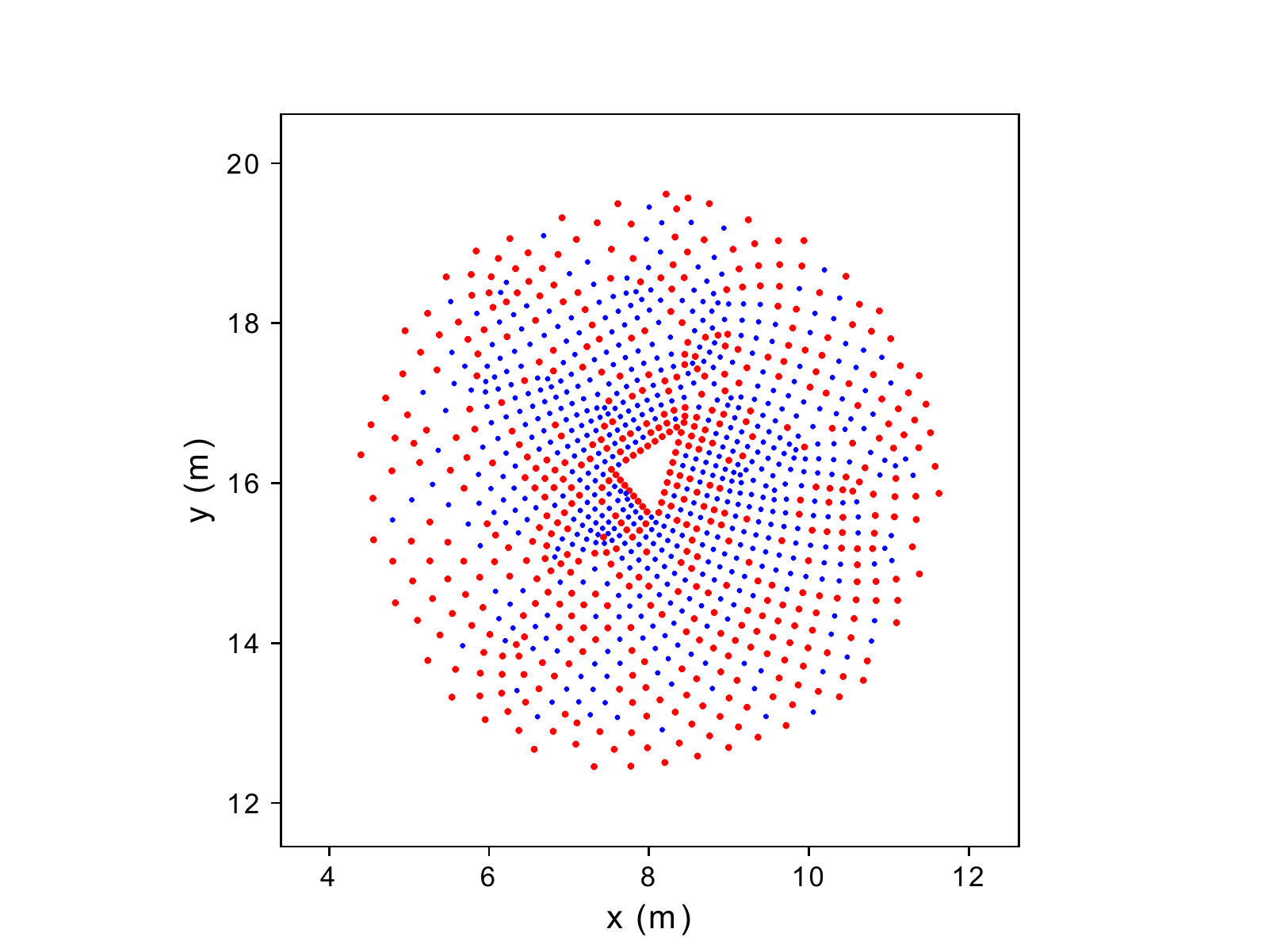}
}
\subfigure{
\includegraphics[width=.3\textwidth]{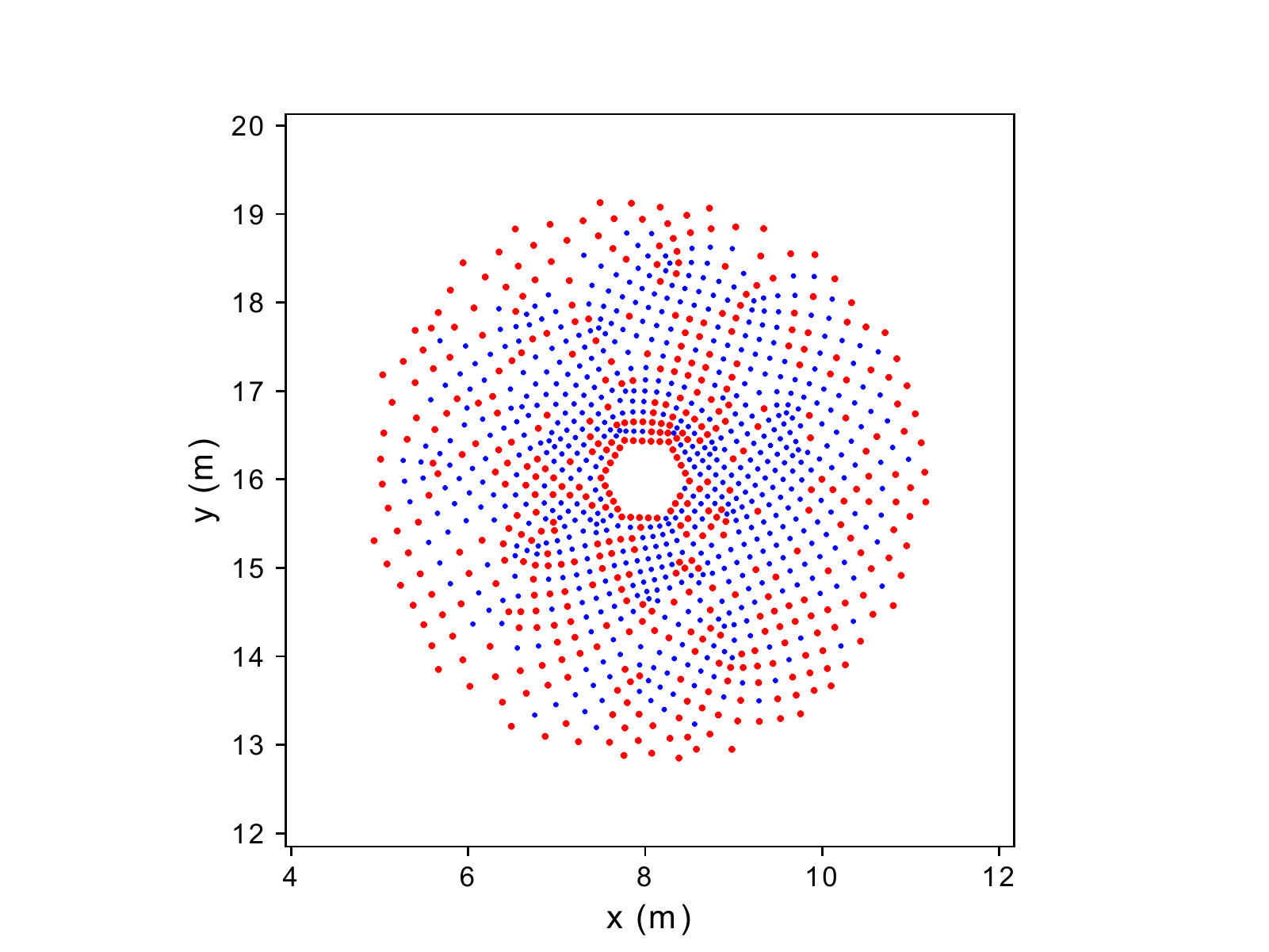}
}

\subfigure{
\includegraphics[width=.3\textwidth]{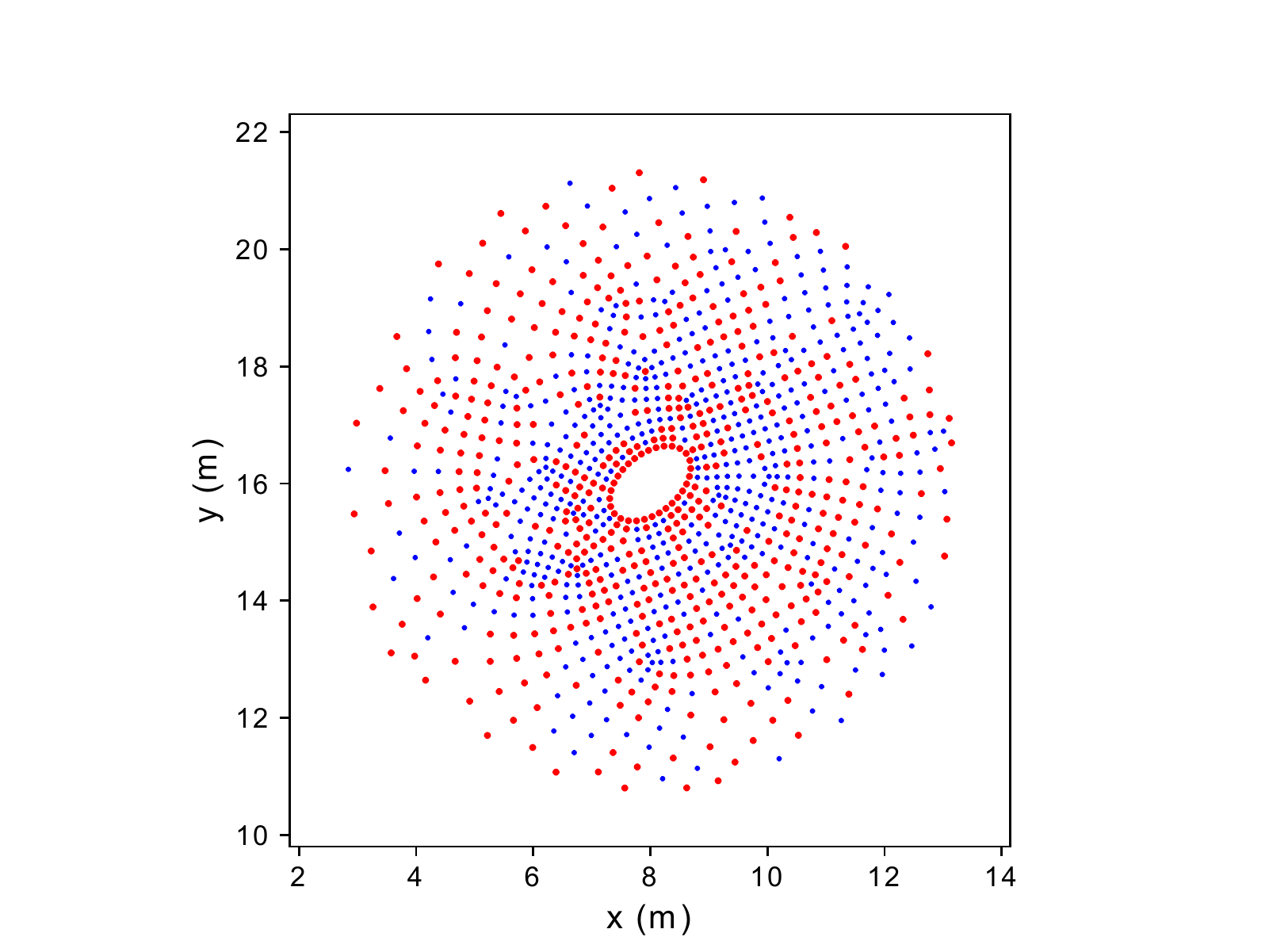}
}
\subfigure{
\includegraphics[width=.3\textwidth]{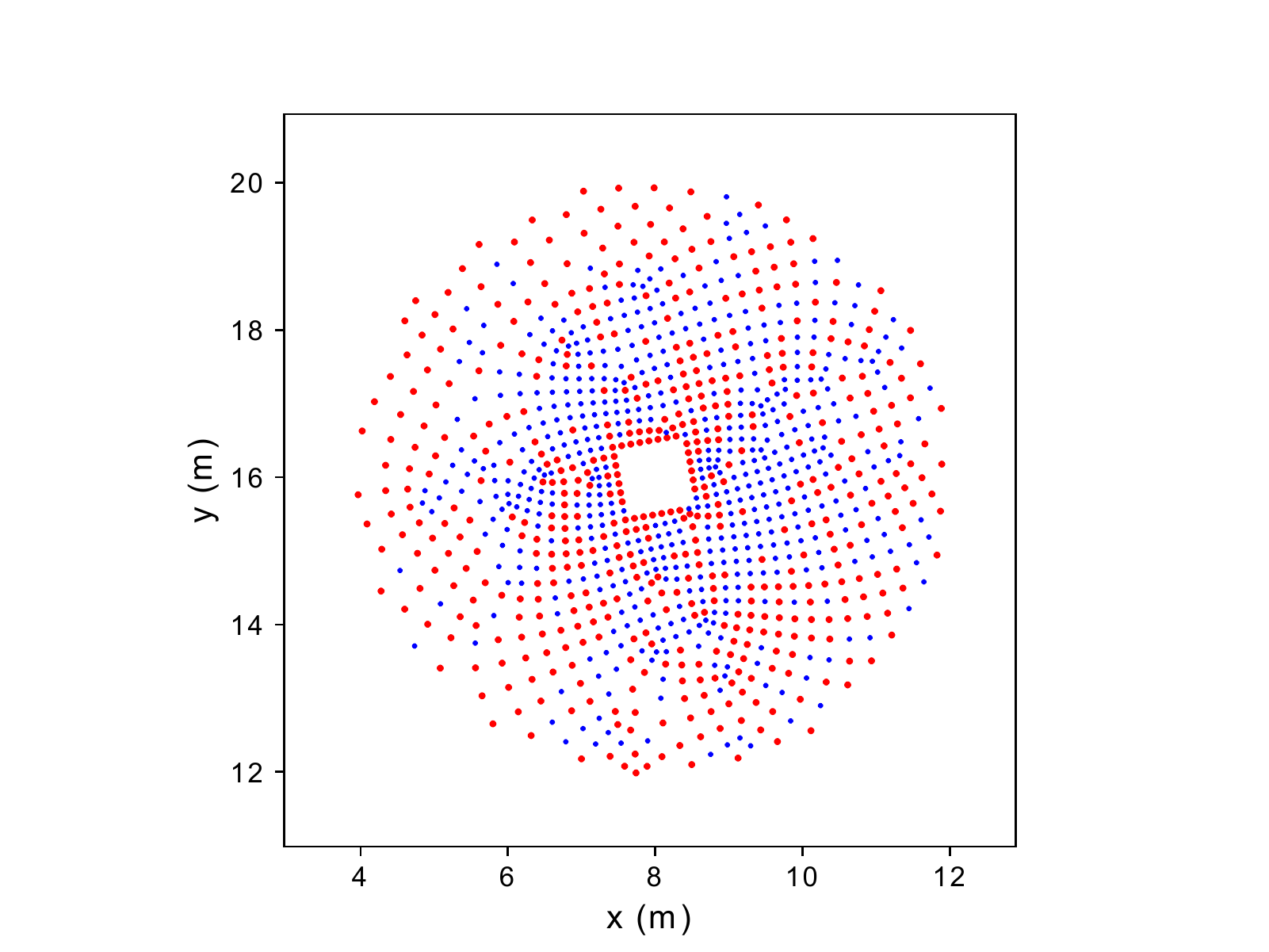}
}
\subfigure{
\includegraphics[width=.3\textwidth]{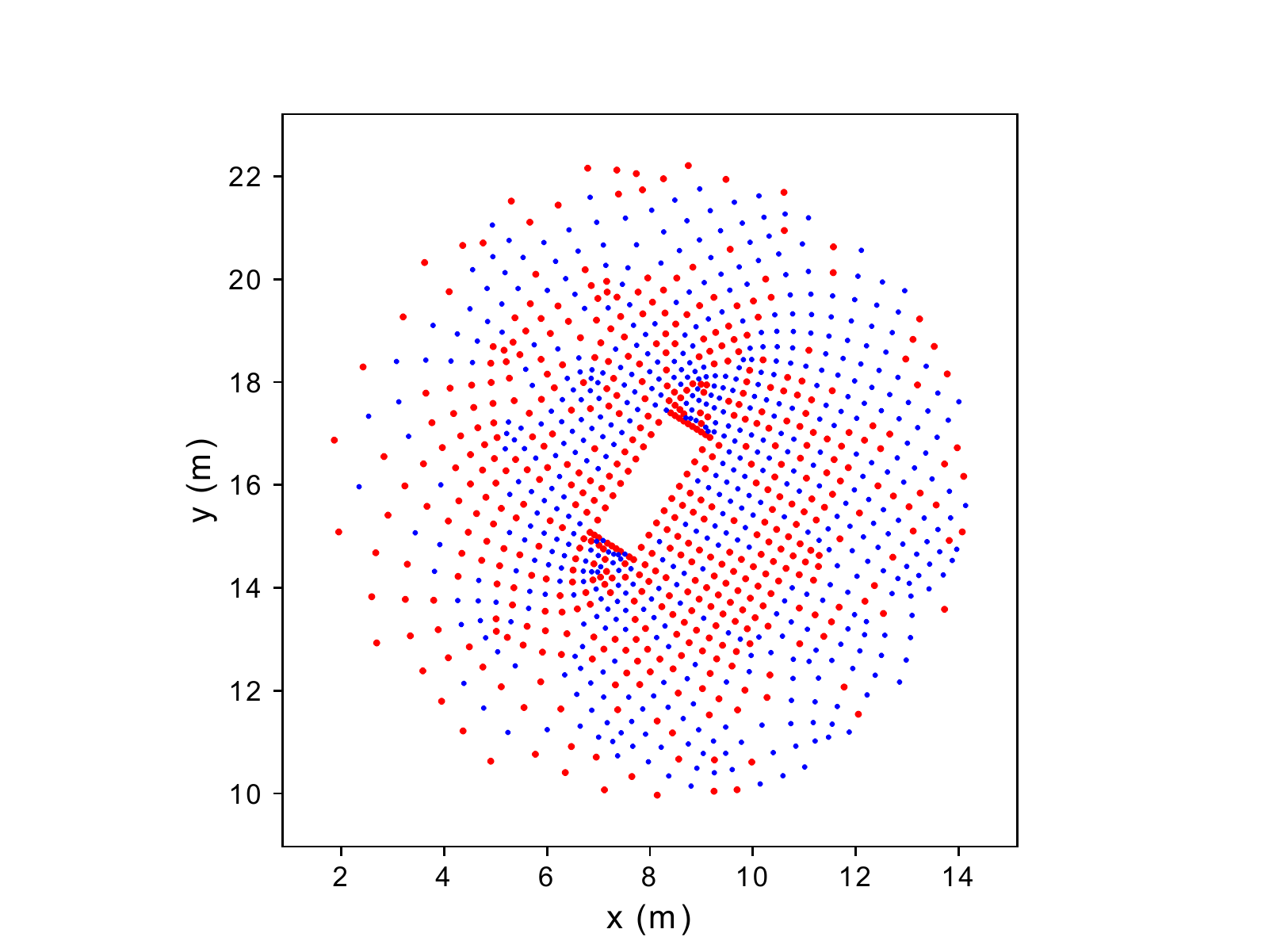}
}
\caption{Critical (red dots) versus non-critical (blue dots) for several point clouds of our test data set}
\label{Critical}
\end{figure*}

One way to gain insight into the features learned by PointNet are to look at the set of ``critical points’’ determined by the network during inference~\cite{qi2017pointnet}. Critical points are defined as the set of input points whose latent features contribute to the global feature extracted by the network. In other words, after the final MLP of the top branch of the network (see Fig.~\ref{Fig3}), a critical point contains the maximum value over all $N$ input points for at least one dimension of its 1024-dimensional latent vector, thereby contributing to the global feature that results from ``max pooling''. Note that ``non-critical'' points can only influence the network output through the 64-dimensional ``local'' feature passed directly to the bottom branch of the architecture. In practice, we identify critical points for each test geometry by computing the ``argmax'' in addition to the ``max pooling'' operation for the global feature; this gives the indices of critical points in the $N\times 3$ input tensor. The critical (red dots) and non-critical (blue dots) points are plotted for several point clouds of our input test data set in Fig. \ref{Critical}. The number of critical points is different for each point cloud and varies between 376 and 526 for the current test data set. Two main insights relevant to the physics are understood by comparing the critical and non-critical points.

The first insight is relevant to the geometry of bluff bodies. As can be seen in Fig. \ref{Critical}, all the points located on the edges of cylinder cross sections are critical. Recall that the flow fields are a function of the geometry of the objects. This demonstrates that the network learns the function by passing all the points placed on the boundary of bluff bodies to the latent global feature.

The second insight is relevant to the momentum (see Eq. \ref{Eq1}) and continuity (see Eq. \ref{Eq2}) of the interior points. We define three residuals as follow:

\begin{align}
\label{EqPhys1}
 & \hat{r}_{momentum_x}=  \\& \left| \frac{1}{M} \sum^{M}_{i=1} \left(\rho \left(\tilde{u}_i\frac{\delta  \tilde{u}_i}{\delta x_i} + \tilde{v}_i\frac{\delta \tilde{u}_i}{\delta y_i}\right) + \frac{\delta \tilde{p}_i}{\delta x_i} - \mu \left(\frac{\delta^2 \tilde{u}_i}{\delta x^2_i} + \frac{\delta^2 \tilde{u}_i}{\delta y^2_i}\right)\right) \right|, \nonumber
\end{align}

\begin{align}
\label{EqPhys2}
&\hat{r}_{momentum_y}= \\ &\left| \frac{1}{M} \sum^{M}_{i=1} \left(\rho \left(\tilde{u}_i\frac{\delta \tilde{v}_i}{\delta x_i} + \tilde{v}_i\frac{\delta \tilde{v}_i}{\delta y_i}\right) + \frac{\delta \tilde{p}_i}{\delta y_i} -\mu \left(\frac{\delta^2 \tilde{v}_i}{\delta x^2_i} + \frac{\delta^2 \tilde{v}_i}{\delta y^2_i}\right)\right) \right|, \nonumber
\end{align}

\begin{equation}
\label{EqPhys3}
 \hat{r}_{continuity}=\left| \frac{1}{M} \sum^{M}_{i=1} \left(\frac{\delta \tilde{u}_i}{\delta x_i} + \frac{\delta \tilde{v}_i}{\delta y_i}\right) \right|,  
\end{equation}
where the $\delta$ operator computes the derivatives of the network-predicted CFD quantities with respect to input points by means of the auto-differentiation capabilities of TensorFlow \cite{tensorflow2015-whitepaper}, in which our network is implemented. In particular, after a forward pass of the network that predicts $(\tilde{u},\tilde{v},\tilde{p})$ at each point, backpropagation computes the derivative of each output with respect to each input dimension $(x,y)$. These terms capture how network predictions will vary with perturbations of the input point locations. $M$ can indicate the number of all critical or non-critical interior points, depending on which set of points we wish to compute the residuals. Note that Eqs. \ref{EqPhys1}—\ref{EqPhys3} are not conceptually equivalent to Eqs. \ref{Eq12}--\ref{Eq14}. As discussed in Sect. \ref{32}, we compute the spatial derivatives of Eqs. \ref{Eq12}--\ref{Eq14} using a finite volume method on a closed volume ($V_{NN}$). In contrast, the spatial derivatives of Eqs. \ref{EqPhys1}--\ref{EqPhys3} are computed by backpropagation on non-neighboring sets of points; this is why we use a different notation for the spatial derivative in Eqs. \ref{EqPhys1}--\ref{EqPhys3}.

Average, maximum, and minimum values of $\hat{r}_{momentum_x}$, $\hat{r}_{momentum_y}$, and $\hat{r}_{continuity}$ for the critical and non-critical points are tabulated in Table \ref{Tab3prime}. As can be realized from Table \ref{Tab3prime}, these residuals are relatively close to zero for both the critical and non-critical points. Note that in contrast with the physics informed neural network techniques (see e.g., Refs. \onlinecite{raissi2019physics,jagtap2020conservative,chen2020physics}), we do not impose Eqs. \ref{EqPhys1}--\ref{EqPhys3} as loss functions ($\mathcal{L}$) to the network. In fact, although the loss function ($\mathcal{L}$) of the neural network is the mean squared error (see Eq. \ref{Eq11}), the values listed in Table \ref{Tab3prime} show that the network learns to represent the underlying physics governed by the Navier-Stokes and continuity equations (see Eqs. \ref{Eq1}--\ref{Eq2}) to a large extent. Notably, the residuals of critical points are higher compared to the non-critical point values. Mathematically, this is because during backpropagation critical point derivatives will involve terms from ``both'' the global and local features, unlike non-critical points which will have no global feature terms due to the ``max'' function. Therefore, we expect derivatives for critical points to have greater magnitude, reflecting their relative influence on network predictions. Physically, this is because of the fact that the neural network represents the physical space of a given point cloud ($V_{NN}$) in an implicit and lower-dimensional space denoted as $U_{NN}$ through the critical points. The space of $U_{NN}$ is represented by the latent global feature (see Fig. \ref{Fig3}). By transition from one point cloud to another, the space of $U_{NN}$ takes a new shape (i.e. the network uses different critical points). During the training process, the weights of the network (see Eq. \ref{pointnet1}) are updated mainly based on the variation in the space of $U_{NN}$, which is constructed by the associated critical points. In other words, the network experiences more difficulty learning the physics on the critical points compared to non-critical points since the critical points create the underlying space ($U_{NN}$); that is why we observe higher residuals at critical points than at non-critical points.

\begin{table*}[ht!]
\caption{\label{Tab3prime}Residuals of momentum and continuity equations computed by the auto-differentiation capabilities of TensorFlow \cite{tensorflow2015-whitepaper} for the interior critical and non-critical points of 259 unseen point clouds. All values are reported in the International Unit System.}
\begin{ruledtabular}
\begin{tabular}{llll}
 & $\hat{r}_{momentum_x}$ & $\hat{r}_{momentum_y}$ & $\hat{r}_{continuity}$\\
\hline
Average for the critical interior points &  4.94592E$-$1 & 3.22539E$-$1 &  5.42833E$-$1\\
Maximum for the critical interior points &  2.26805 & 2.24237 &  3.54595\\
Minimum for the critical interior points &  2.22393E$-$1 & 2.83308E$-$1 &  1.14674E$-$1\\
\hline
Average for the non-critical interior points &  6.32048E$-$2 & 5.77430E$-$2 &  4.89946E$-$2\\
Maximum for the non-critical interior points &  1.45977E$-$1 & 1.35531E$-$1 &  7.97416E$-$2\\
Minimum for the non-critical interior points &  1.71704E$-$2 & 1.38868E$-$2 &  1.91979E$-$2\\
\end{tabular}
\end{ruledtabular}
\end{table*}

\subsection{\label{33}Neural network  generalizability}

In the field of computer graphics, it is common (see e.g., Refs. \onlinecite{rempe2020predicting,jiang2020local}) to test a neural network on a totally new class of shape. For instance, \citet{qi2017pointnet} tested PointNet for the semantic segmentation of unseen categories such as ``face’’, ``house’’, ``rabbit’’, and ``teapot’’, while these objects did not exist in their data set. The outcomes demonstrated that the learned PointNet functions were generalizable, although the results were negatively affected by the training data set. In contrast with the computer graphics community, the idea of fabricating neural networks with an ability to generalize to new domain geometry categories is not yet well-established in the deep learning area of computational mechanics. We hope experiments given in this subsection provides a starting point to this context in the field of study.

To examine the generalizability of our neural network, we consider two sets of tests in the following subsections: prediction of the velocity and pressure fields around multiple cylinders with different cross sections, and an airfoil. None of these domain geometries are in our data set (see Table \ref{Tab1}). Although it is not expected to achieve a high level of accuracy for such cases, this investigation might reveal new aspects of the neural network and provide directions for increasing its performance.

\begin{figure*}
\centering
\includegraphics[width=1.0\linewidth]{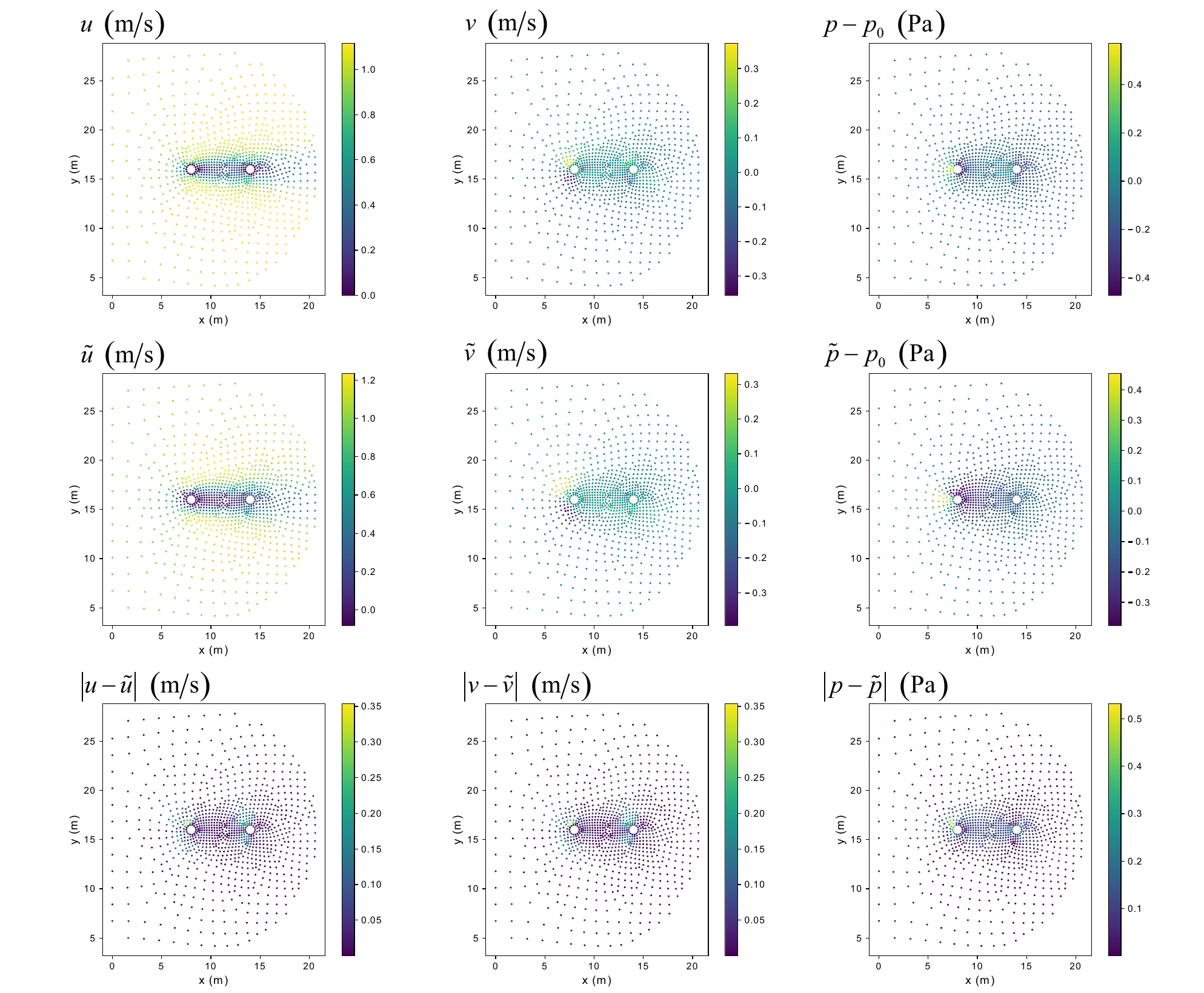}
\caption{A comparison between the ground truth and our network prediction for the velocity and pressure fields of two circular cylinders}
\label{Multi1}
\end{figure*}

\begin{figure*}
\centering
\includegraphics[width=1.0\linewidth]{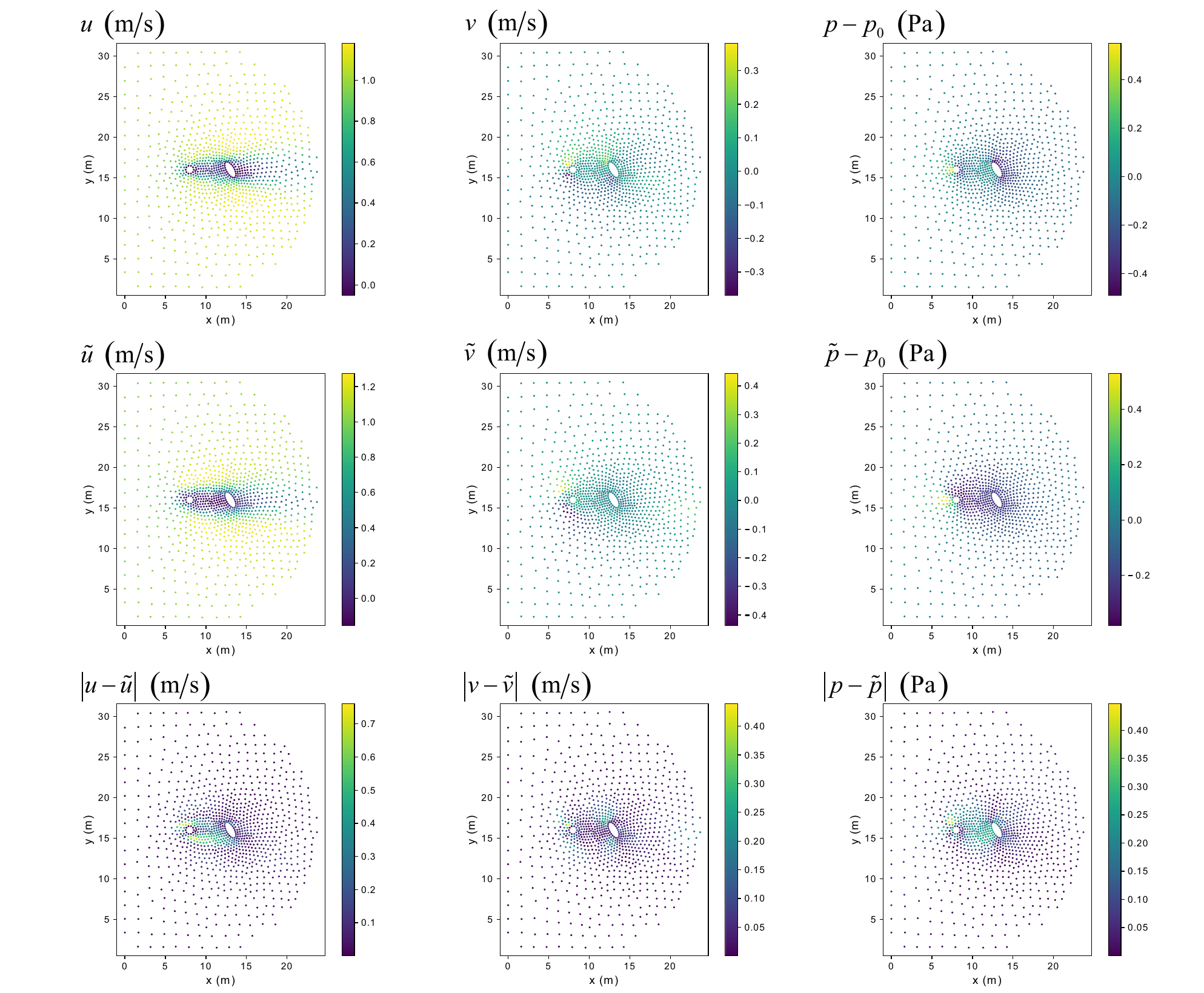}
\caption{A comparison between the ground truth and our network prediction for the velocity and pressure fields of two circular and elliptical cylinders}
\label{Multi2}
\end{figure*}

\begin{figure*}
\centering
\includegraphics[width=1.0\linewidth]{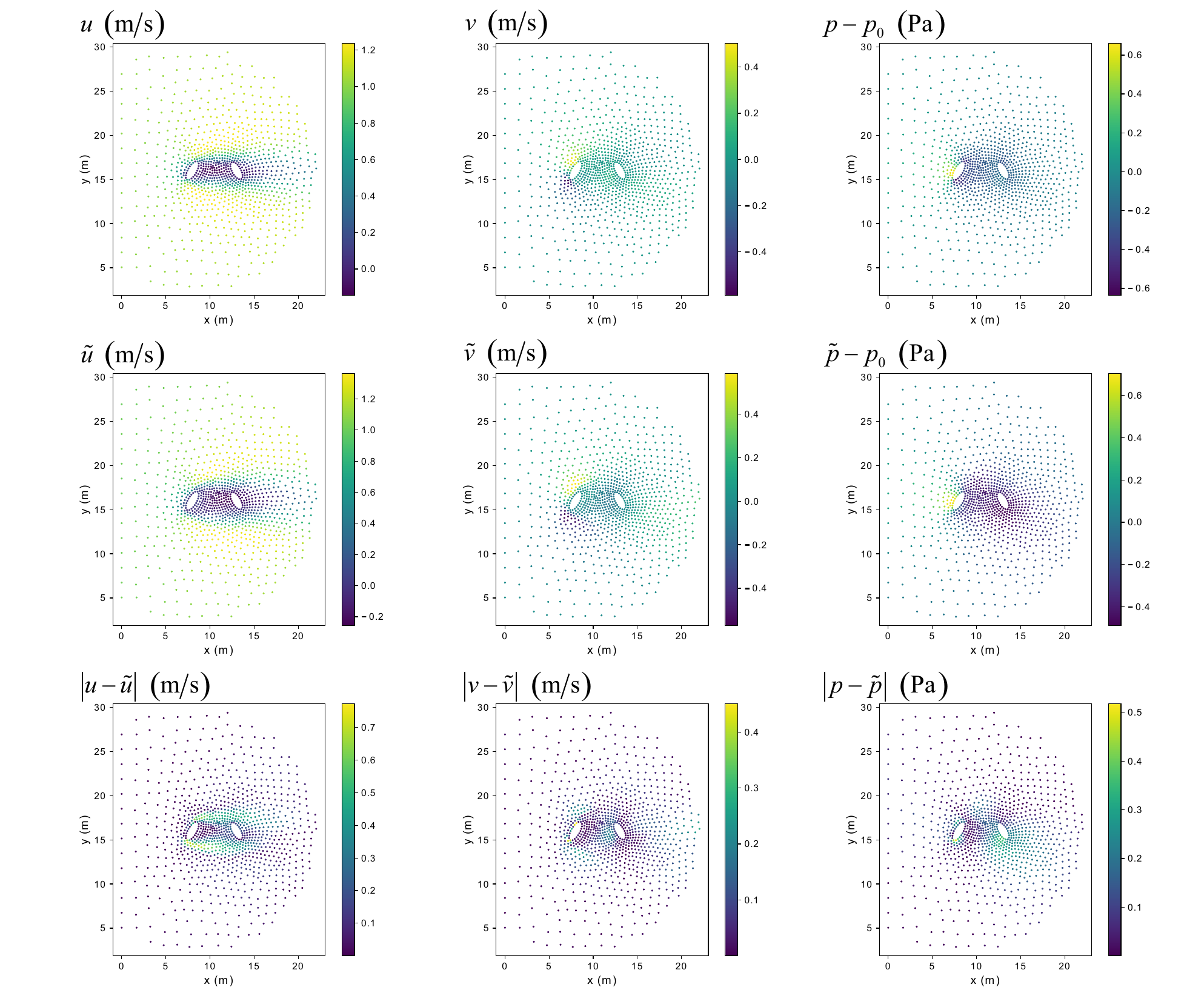}
\caption{A comparison between the ground truth and our network prediction for the velocity and pressure fields of two elliptical cylinders}
\label{Multi3}
\end{figure*}

\begin{figure*}
\centering
\includegraphics[width=1.0\linewidth]{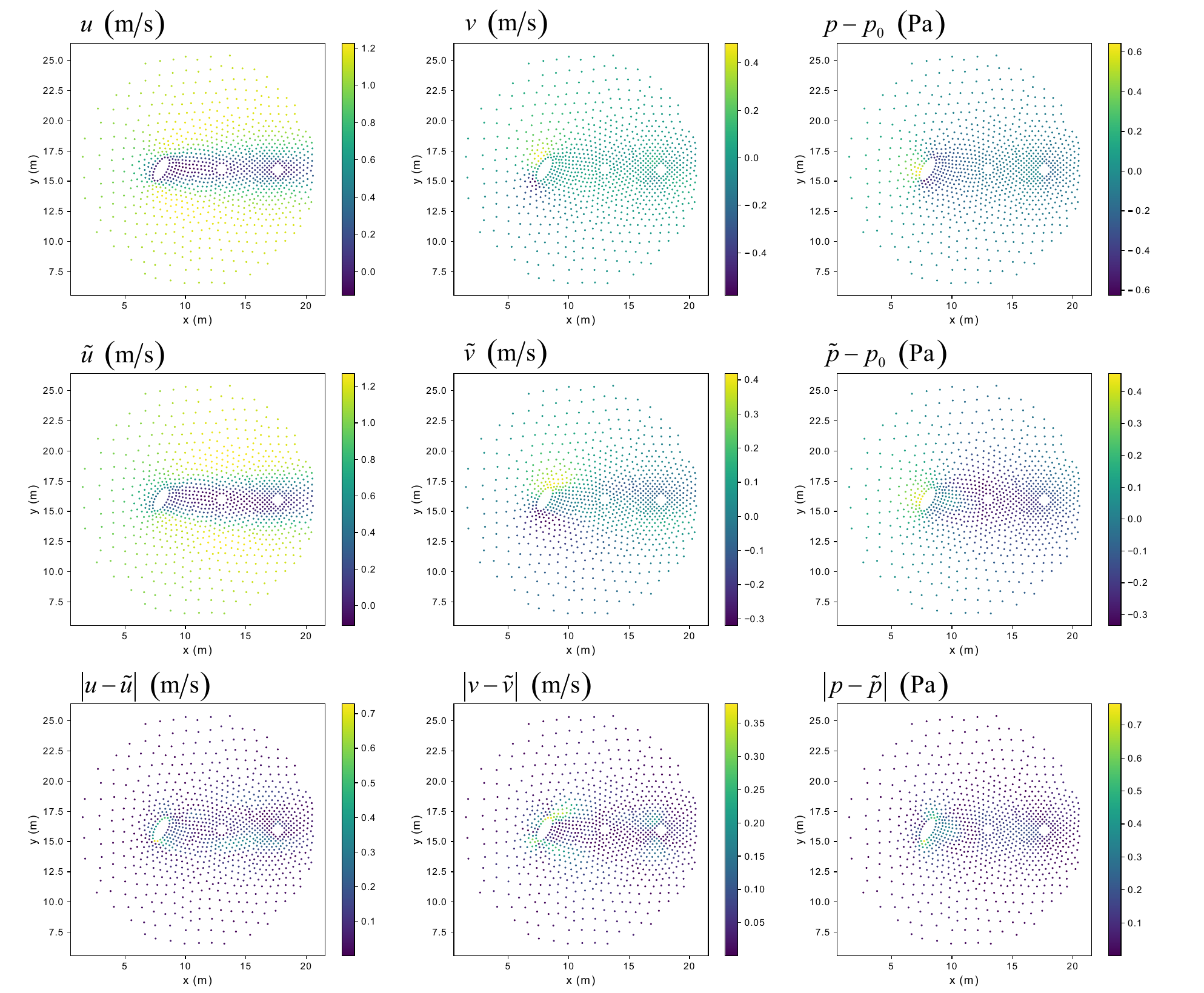}
\caption{A comparison between the ground truth and our network prediction for the velocity and pressure fields of three elliptical, circular, and rectangular cylinders}
\label{Multi4}
\end{figure*}

\subsubsection{\label{331}Prediction of flow around multiple objects}

For the first time, the neural network proposed in this article predicts the velocity and pressure spaces around multiple objects without seeing multi-object data during the training process. In all the test cases, we set the distance between objects such that the flow remains steady and laminar.

\begin{table*}
\caption{\label{Table7}Error analysis of the velocity and pressure fields predicted by our neural network for multiple bodies; $||\ldots||$ indicates the $L^2$ norm.}
\begin{ruledtabular}
\begin{tabular}{llll}
  & $||u-\tilde{u}||$ & $||v-\tilde{v}||$ & $||p-\tilde{p}||$ \\
\hline
Two circular cylinders (Fig. \ref{Multi1}) & 1.51130E$-$1 & 5.80172E$-$2 & 9.22659E$-$2 \\
Circular and elliptical cylinders (Fig. \ref{Multi2})  & 1.63664E$-$1 & 7.50613E$-$2 & 1.09544E$-$1 \\
Two elliptical cylinders (Fig. \ref{Multi3}) & 2.19626E$-$1 & 8.93397E$-$2 & 1.35396E$-$1 \\
Elliptical, circular, and rectangular cylinders (Fig. \ref{Multi4}) & 1.30335E$-$1 & 8.59303E$-$2 & 1.33239E$-$1 \\
\end{tabular}
\end{ruledtabular}
\end{table*}

Figures \ref{Multi1}--\ref{Multi4} compare the solution of the Navier-Stokes and continuity equations obtained by the CFD solver and our neural network for various multi-object test cases. Table \ref{Table7} summarizes the pointwise error ($L^2$ norm) for each test case, showing that the network predicts these flow fields with reasonable levels of accuracy. In all cases, the error of predicting $\tilde{u}$ is greater than the corresponding error of $\tilde{v}$ and $\tilde{p}$, similar to the average error reported in Table \ref{Table2}. Interestingly, the maximum errors of the velocity fields are less than the maximum corresponding errors for prediction of unseen data in the test set reported in Table \ref{Table2} (22\% versus 25\% and 9\% versus 24\% for $\tilde{u}$ and $\tilde{v}$, respectively). For the pressure field, we observe nearly the same extremum (13\% versus 12\% for unseen data from seen categories and unseen data from unseen categories, respectively). We explain details of each test case in the following paragraph.

To more precisely investigate the prediction integrity, let us discuss the outcomes of Fig. \ref{Multi1} where the network predicts the velocity and pressure spaces around two circular cylinders located a horizontal distance from each other. As can be seen in Fig. \ref{Multi1}, the pressure in front of the first cylinder experiences higher values in comparison to the second cylinder. This experiment demonstrates that the network takes into account the presence of both cylinders in the domain. Otherwise, we observed the same pressure values in front of the both cylinders. This scenario is repeated for the prediction of the velocity field as well, as can be monitored in Fig. \ref{Multi1}. A similar conclusion is taken by observing the results exhibited in Figs. \ref{Multi2}--\ref{Multi4}. In comparison with the test case shown in Fig. \ref{Multi1}, Figure \ref{Multi2} depicts a more challenging condition: two cylinders are located in $V_{NN}$, one with circular and another with elliptical cross sections with two different length scales. According to the data collected in Table \ref{Table7}, the pointwise error ($L^2$ norm) of all the velocity and pressure fields slightly increase in comparison with the presence of two circular cylinders. To evaluate a more complicated test case, we consider two elliptical cylinders in the next experiment as shown in Fig. \ref{Multi3}. As expected, the corresponding $L^2$ norm of the velocity and pressure fields moderately increase compared to the two previous test cases as tabulated in Table \ref{Table7}. Lastly, we challenge the network by placing three cylinders with three different cross sections of elliptical, circular, and rectangular shapes inside the flow field as exhibited in Fig. \ref{Multi4}. Note that such an experiment has not been performed in the recent literature \cite{sekar2019fast,guo2016convolutional,tompson2017accelerating,thuerey2019deep,bhatnagar2019prediction,jin2018prediction,zhang2018application,han2019novel,hui2020fast,hasegawa2020cnn} using CNNs due to the limited size of the considered domain, which cannot accommodate several bodies that are a reasonable distance from each other.

We explain this specific characteristic of our network from a computer science perspective at a high level. The network extracts geometrical features of point clouds and represents them as a global feature with a latent code of size 1024 as can be seen in Fig. \ref{Fig3}. Additionally, the bottom branch of the network is a combination of nonlinear functions (see Eqs. \ref{pointnet2}--\ref{pointnet3}) that learned to represent the solution of the Navier-Stokes and continuity equations (see Eqs. \ref{Eq1}--\ref{Eq4}) as a function of the geometrical features. Moreover, for the data generated here we set the Reynolds number to only be a function of the length scale which is a geometrical feature. Recall that the Reynolds number determines the flow regime. Thus, we conclude that our network is trained, in fact, for a specific range of the Reynolds number rather than for a specific number of bodies inside the domain. Therefore, although the network has only seen one object per object class during the training procedure, it predicts the flow field around multiple bodies at least with a reasonable level of accuracy assuming two conditions hold. First, the flow remains in the training regime (e.g., laminar and steady in the current study). Second, the size of the latent code of global feature is large enough to digit the necessary geometrical features of input point clouds. For instance, it is expected that a network with a latent global feature of size 2048 would predict the velocity and pressure fields around three cylinders (see Fig. \ref{Multi4}) more accurately in comparison with the current network. Further elaboration on this feature of our network needs an advanced machine learning discussion and goes beyond the scope of this journal; thus we skip it in this article.

It is essential to mention that our network's ability to predict flow fields around multiple bodies is important mainly because the Navier-Stokes equations (see Eq. \ref{Eq1}) are nonlinear and using superposition is not valid in this case. For instance, if we trained our network for a physical phenomenon governed by a linear equation (e.g., heat conduction governed by the Poisson equation), predicting the temperature field over multiple objects would not be advantageous for our network, since one could simply ask the network to predict the temperature field around each object individually and then superpose them manually.

\begin{figure*}
\centering
\includegraphics[width=1.0\linewidth]{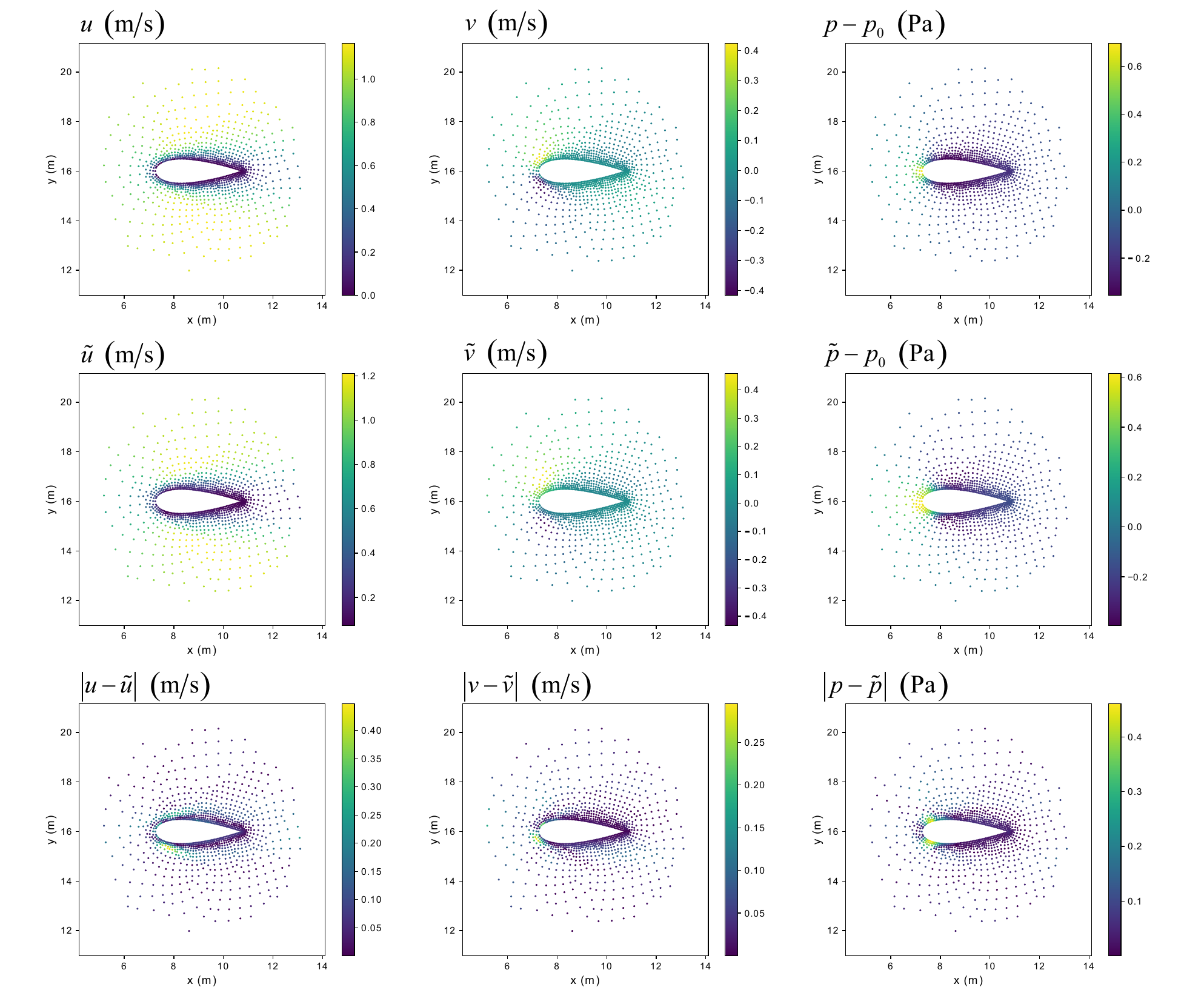}
\caption{A comparison between the ground truth and our network prediction for the velocity and pressure fields of NACA 0028 for the angle of attack of $0^\circ$}
\label{Fig9}
\end{figure*}

\begin{figure*}
\centering
\includegraphics[width=1.0\linewidth]{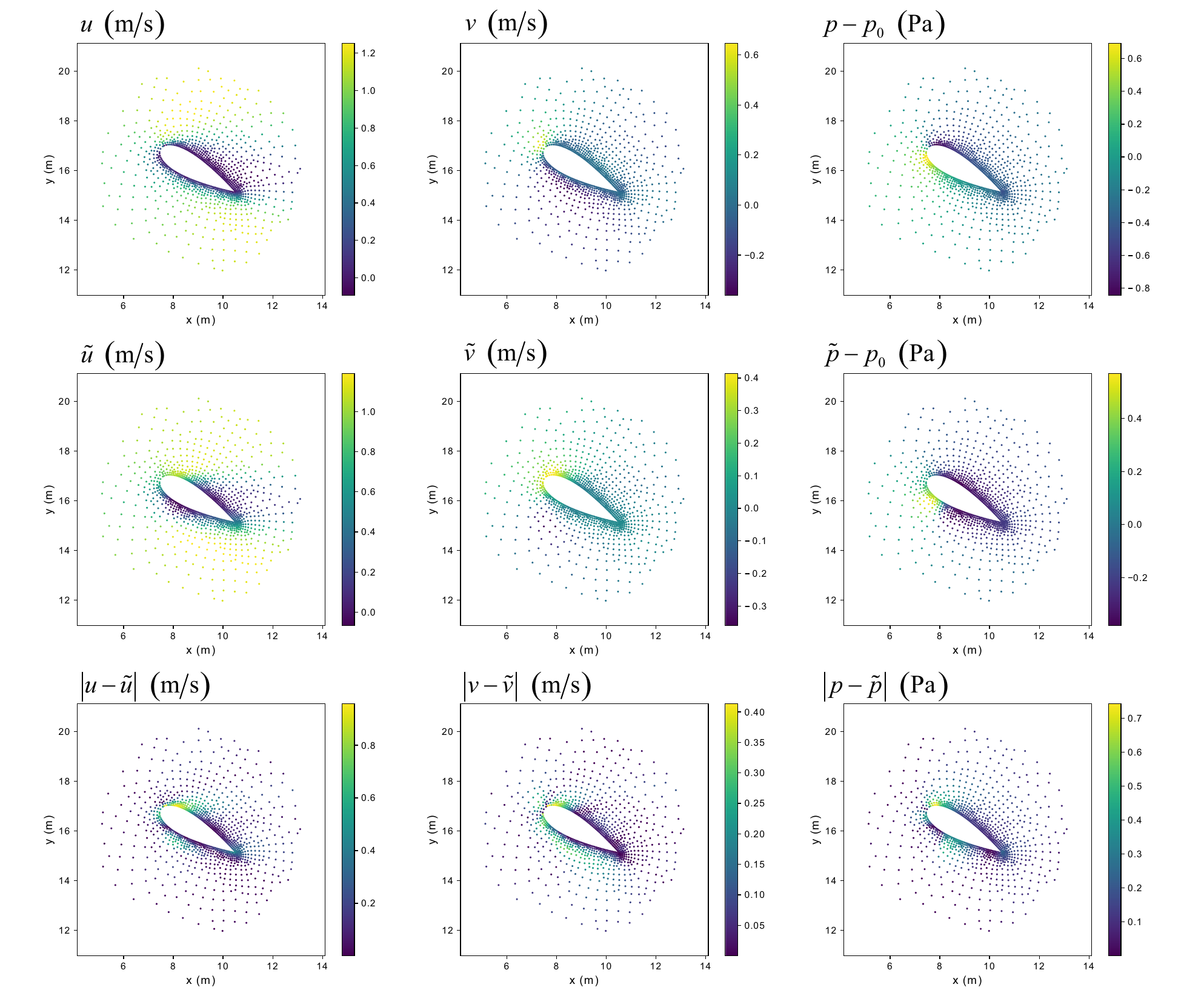}
\caption{A comparison between the ground truth and our network prediction for the velocity and pressure fields of NACA 0028 for the angle of attack of $30^\circ$}
\label{Fig10}
\end{figure*}

\begin{figure*}
\centering
\includegraphics[width=1.0\linewidth]{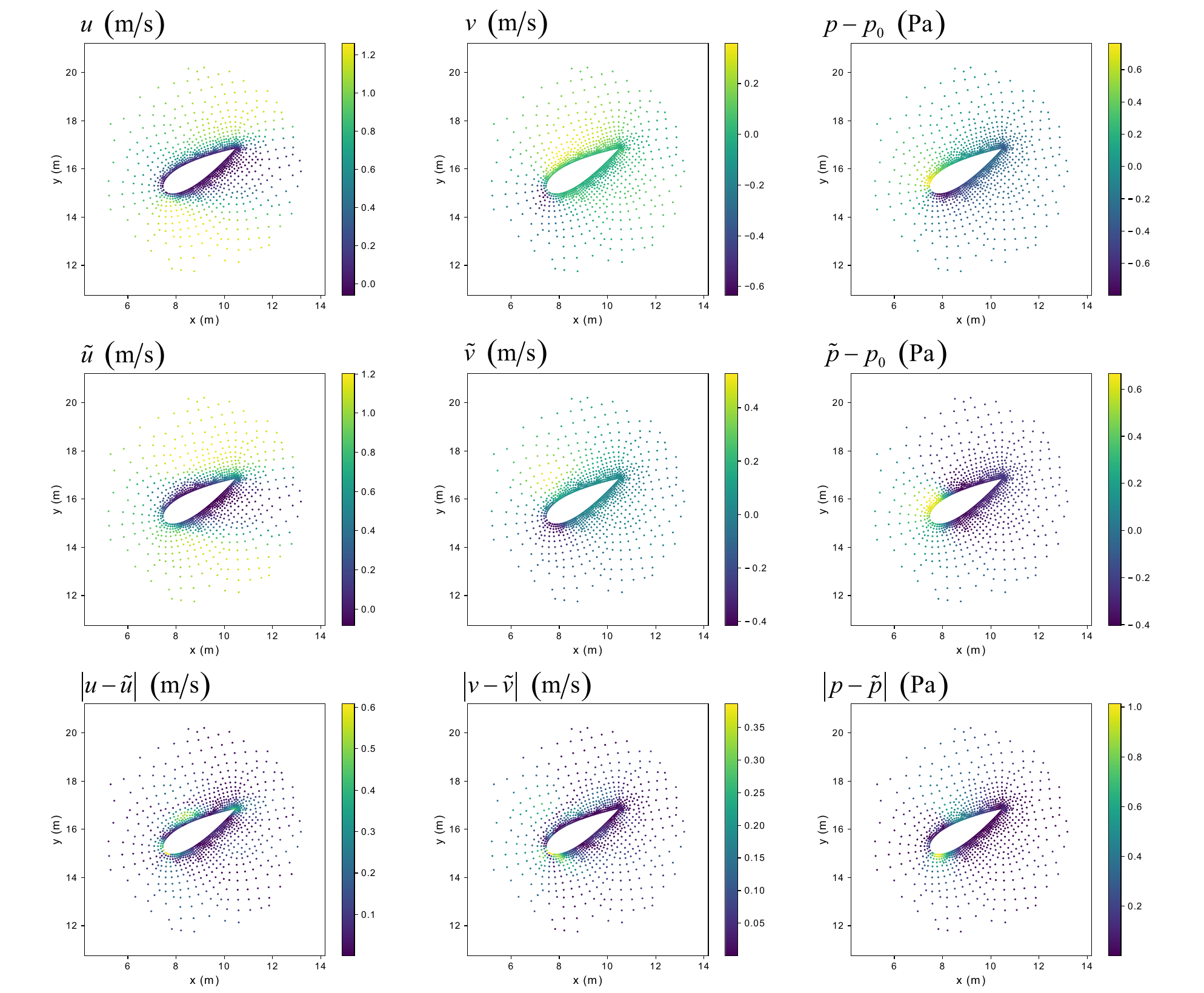}
\caption{A comparison between the ground truth and our network prediction for the velocity and pressure fields of NACA 0028 for the angle of attack of $-30^\circ$}
\label{Fig11}
\end{figure*}

\begin{figure*}
\centering
\subfigure{
\includegraphics[width=.3\textwidth]{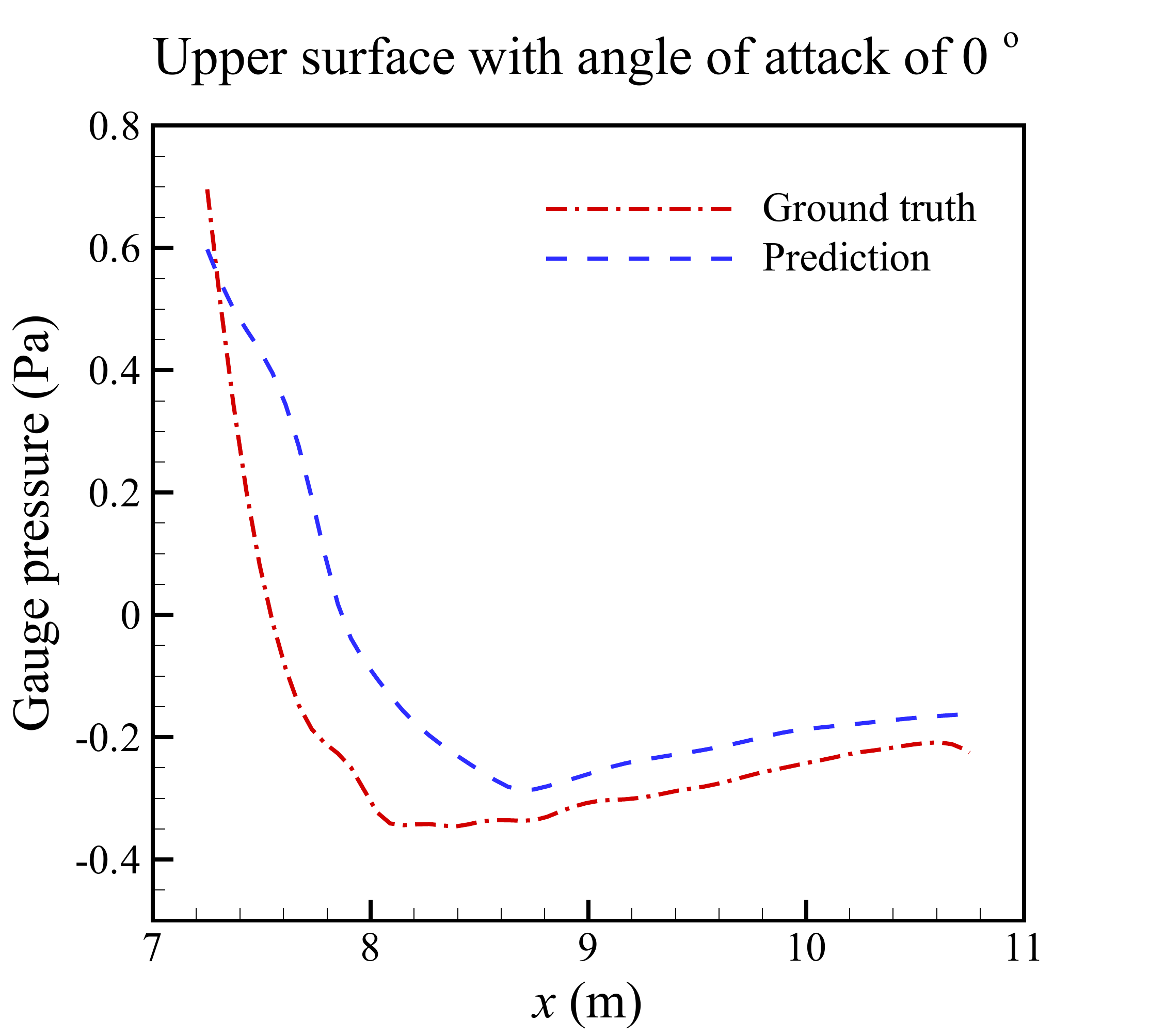}
}
\subfigure{
\includegraphics[width=.3\textwidth]{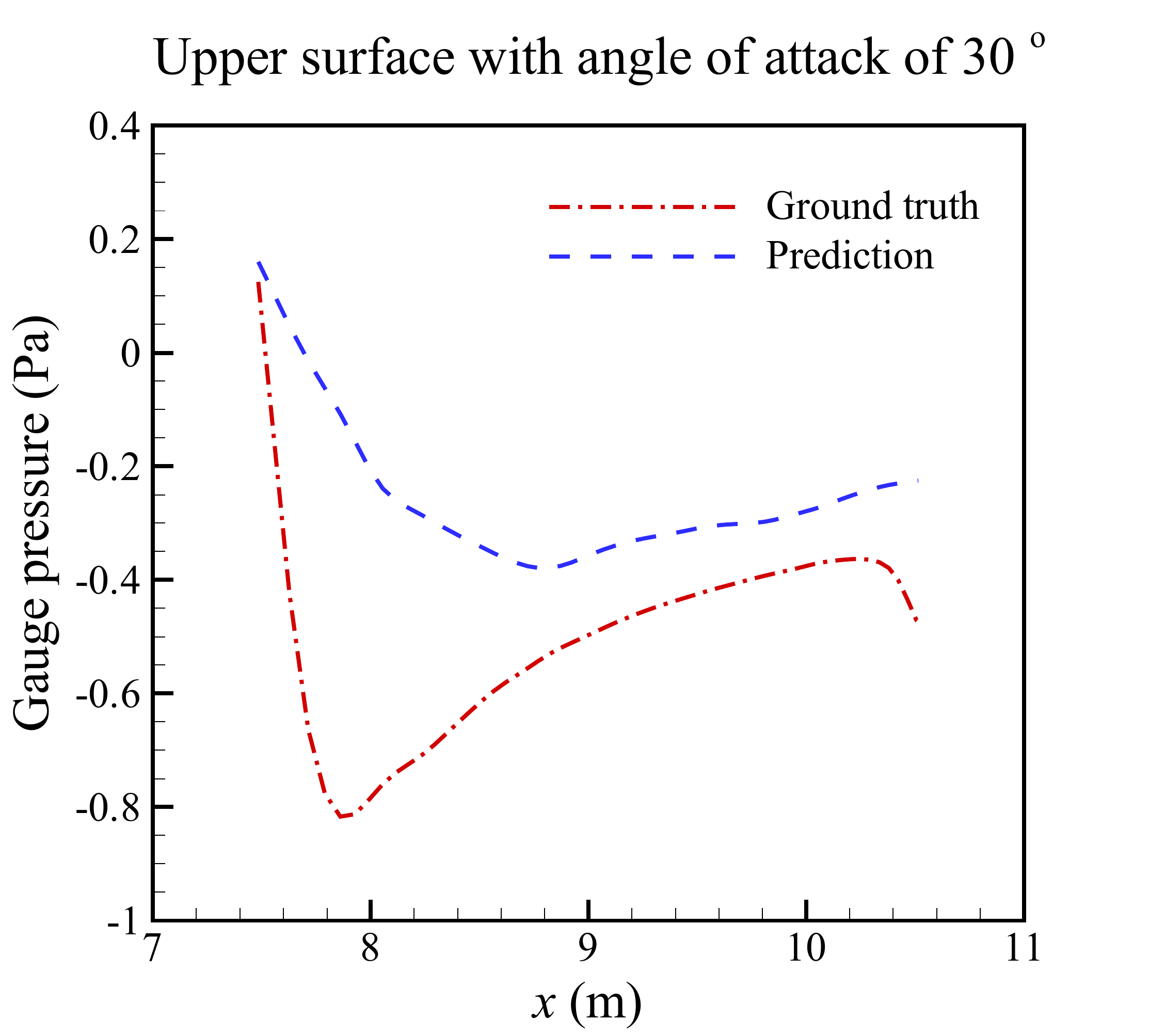}
}
\subfigure{
\includegraphics[width=.3\textwidth]{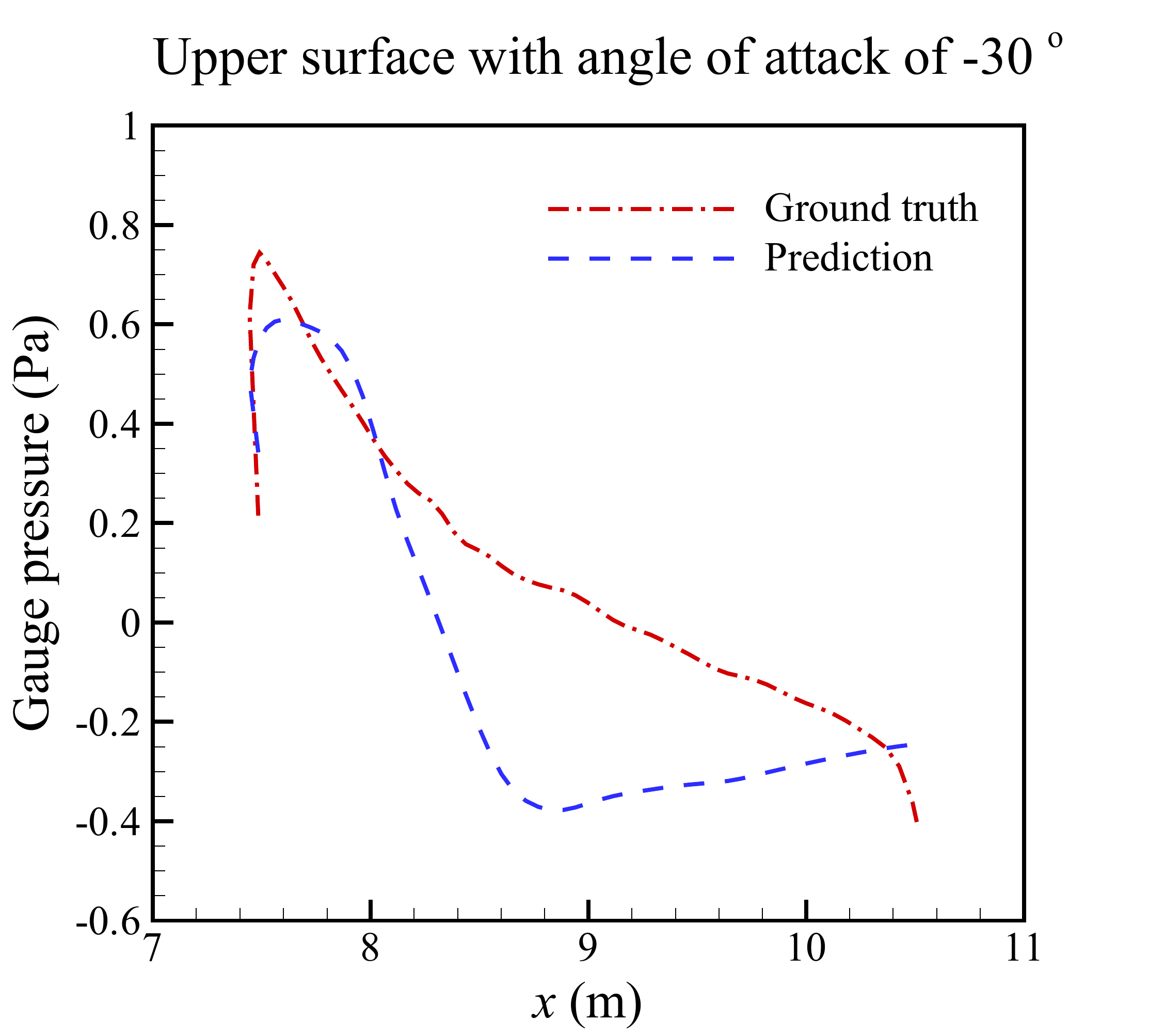}
}

\subfigure{
\includegraphics[width=.3\textwidth]{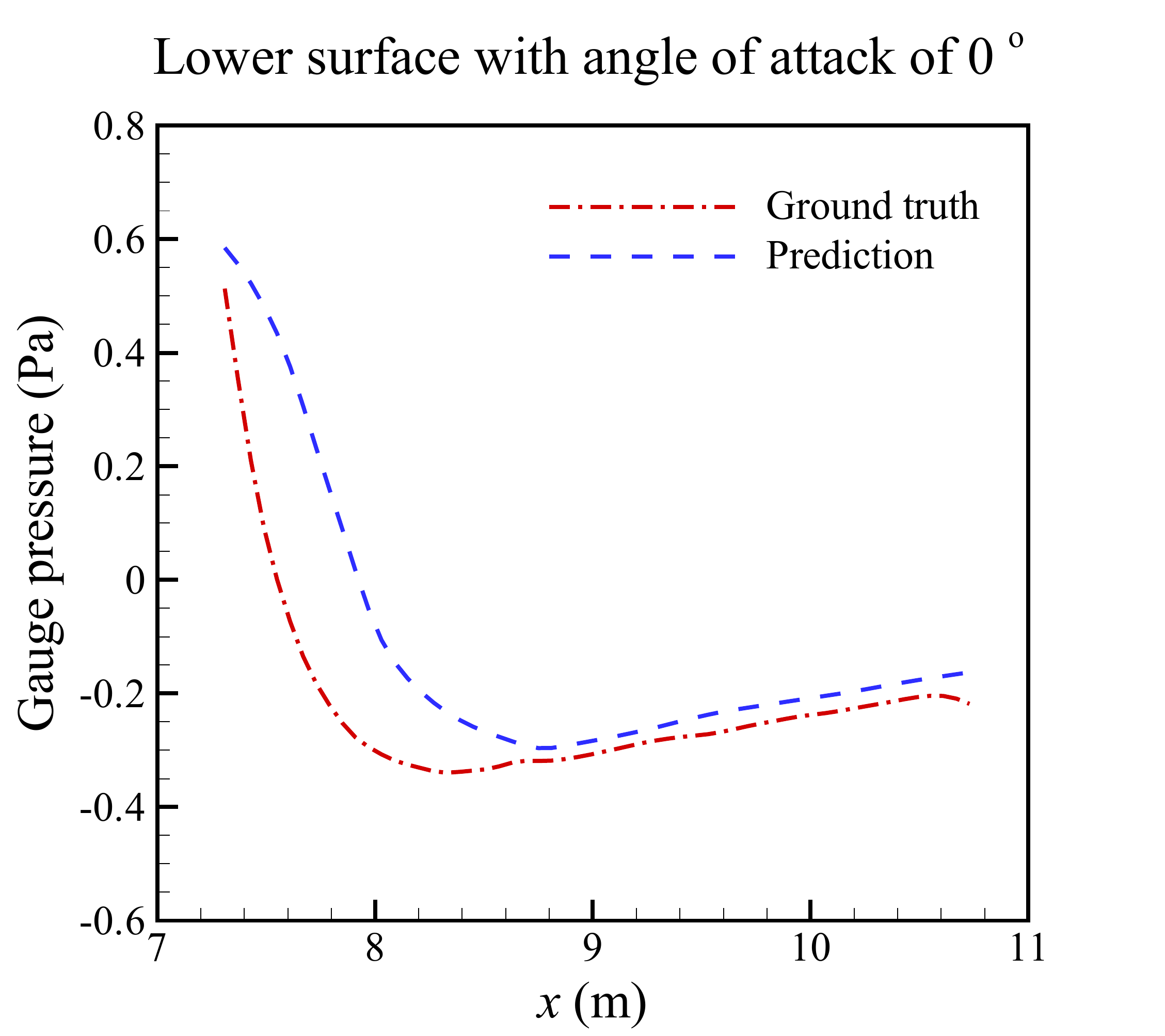}
}
\subfigure{
\includegraphics[width=.3\textwidth]{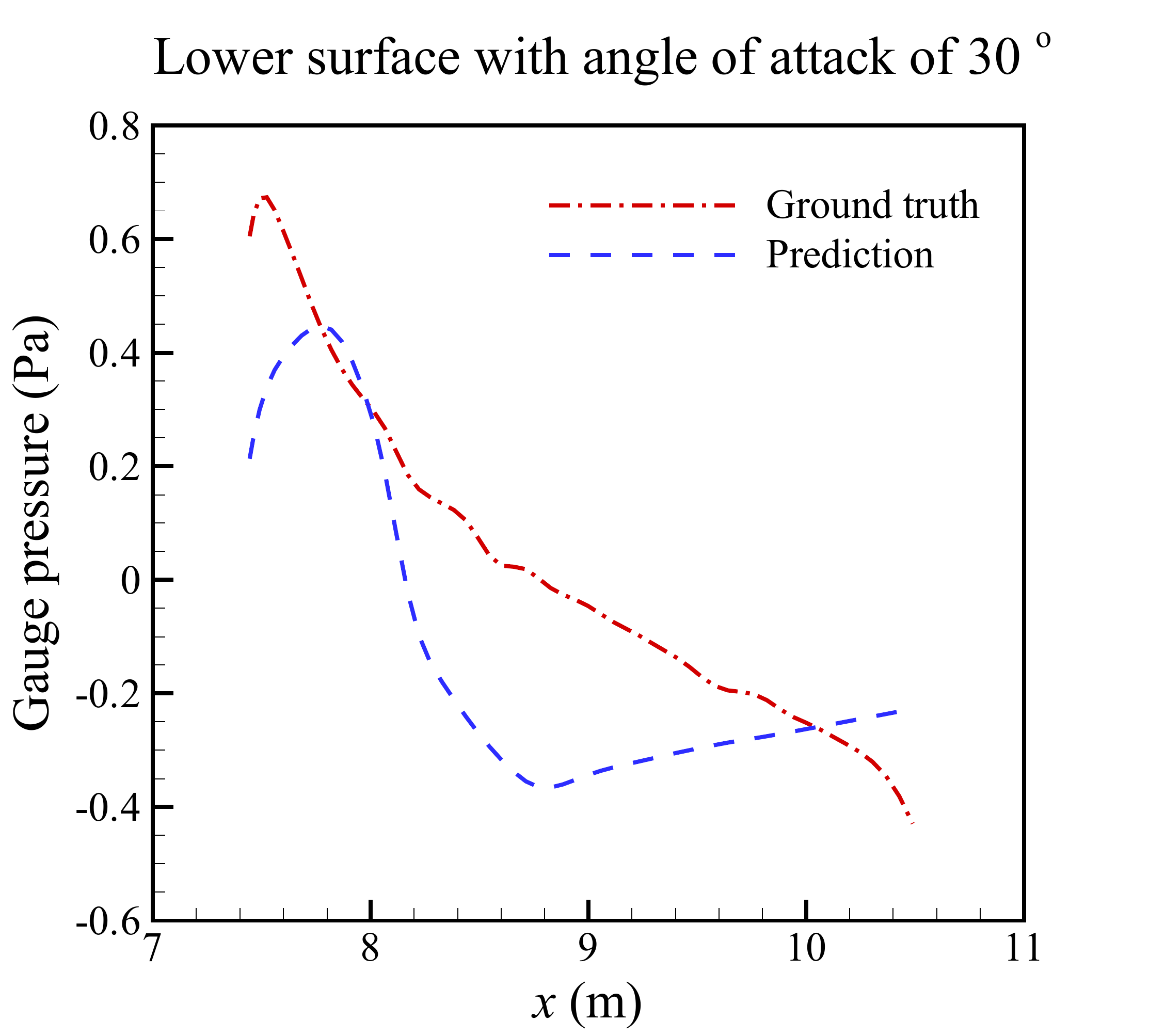}
}
\subfigure{
\includegraphics[width=.3\textwidth]{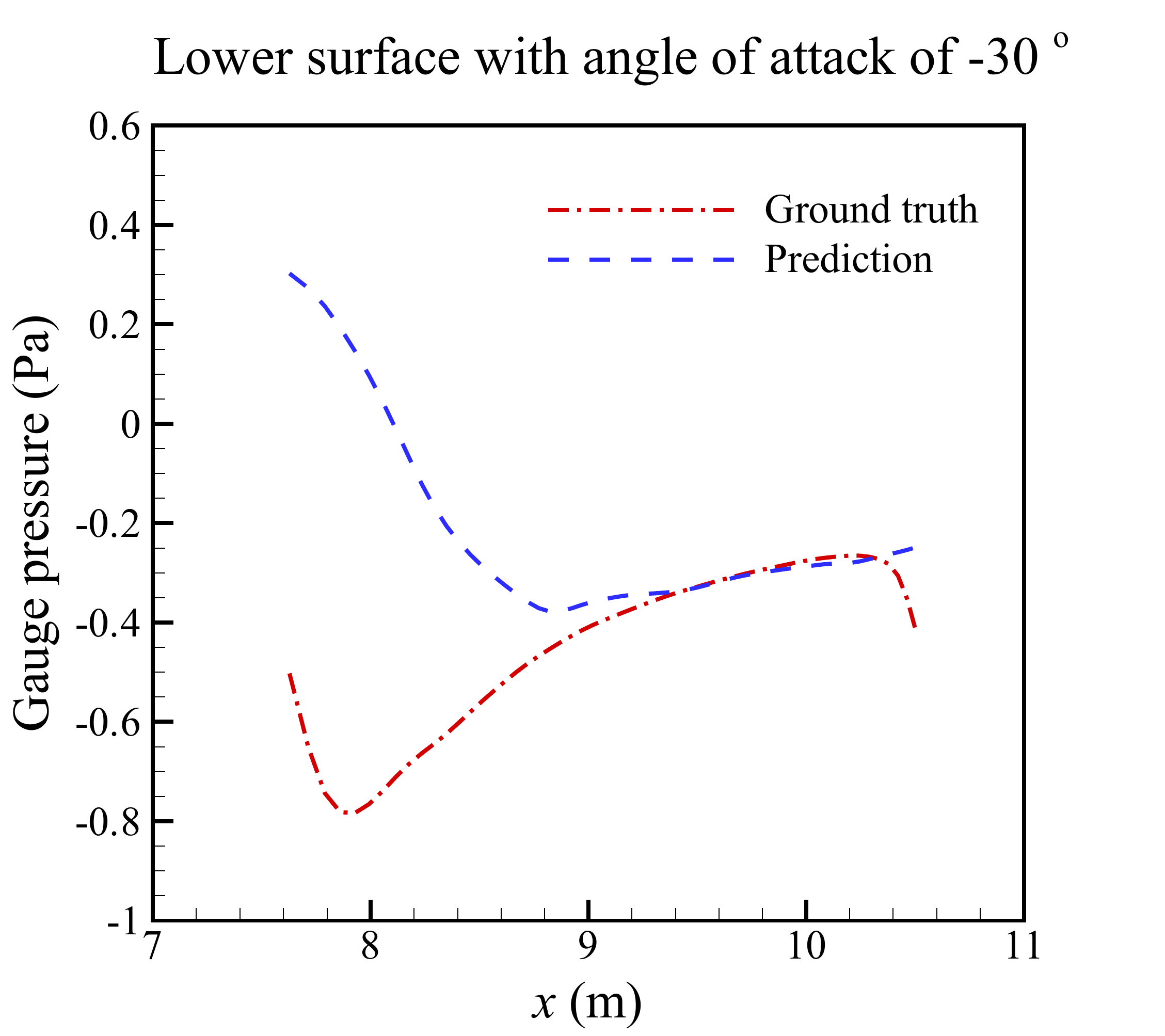}
}
\caption{A comparison between the ground truth and our network prediction for the pressure distribution on the surface of NACA 0028 for the angles of attack of $0^\circ$, $30^\circ$, and $-30^\circ$}
\label{FigAirfoilPressure}
\end{figure*}

\subsubsection{\label{332}Prediction of flow around an airfoil}

Prediction of CFD quantities for airfoils using deep learning techniques has been previously reported in the literature \cite{greenman1999high,zhang2018application,sekar2019fast,thuerey2019deep,bhatnagar2019prediction,han2019novel,hui2020fast}. In all the works cited, the class of airfoil object was among the training data set. For the first time, our neural network tackles prediction of the space of velocity and pressure fields around an airfoil at different angles of attack, while the network has never seen airfoils during the training process. For this purpose, we select NACA 0028, which is an important four-digit symmetrical airfoil and a useful tool for academic and industrial researches (see e.g., Refs. \onlinecite{bertolotti2007rotor,thacher2015solar,hascoet2019two}).

\begin{table}
\caption{\label{Table4}Error analysis of the velocity and pressure fields predicted by our neural network for NACA 0028; $||\ldots||$ indicates the $L^2$ norm.}
\begin{ruledtabular}
\begin{tabular}{llll}
Angle of attack & $||u-\tilde{u}||$ & $||v-\tilde{v}||$ & $||p-\tilde{p}||$ \\
\hline
$0^\circ$ & 1.05631E$-$1 & 5.47357E$-$2 & 1.04842E$-$1 \\
$30^\circ$ & 2.83492E$-$1 & 1.30977E$-$1 & 2.07335E$-$1 \\
$-30^\circ$ & 1.62185E$-$1 & 1.00821E$-$1 & 2.42353E$-$1\\
\end{tabular}
\end{ruledtabular}
\end{table}

A visual comparison between the velocity and pressure spaces predicted by the neural network and obtained by the CFD solver is made in Figs. \ref{Fig9}--\ref{Fig11} respectively for three angles of attack of $0^\circ$, $30^\circ$, and $-30^\circ$; along with the pointwise error distribution on the point clouds. In addition to the graphical comparison, Table \ref{Table4} lists the pointwise error ($L^2$ norm) for the predicted fields with reference to the numerical solution computed by the CFD solver as the ground truth. Since airfoils are not contained in our training data set, only a reasonable level of accuracy is obtained. Based on the data collected in Table \ref{Table4}, the minimum error occurs for the angle of attack of $0^\circ$ in comparison with the other two cases with the angles of attack of $30^\circ$ and $-30^\circ$. In all the cases, the pointwise error ($L^2$ norm) of $\tilde{u}$ is greater than the corresponding error of $\tilde{v}$. Similar trends are observed in Table \ref{Table2}, where we report the error on the test set. Moreover, the maximum error happens in the front of the airfoil for all the angles of attack, due to a more complicated curvature compared to the rear of NACA 0028. Besides the flow fields exhibited in Figs. \ref{Fig9}--\ref{Fig11}, the pressure distribution obtained using the CFD solver and our neural network on the airfoil surface is plotted in Fig. \ref{FigAirfoilPressure} for a more precise evaluation of the network generalizability. We select the pressure distribution because the maximum infinity-norm error on the airfoil surface belongs to this field rather than the velocity fields. Moreover, this maximum error happens on the airfoil surface instead of elsewhere in the domain. As shown in Fig. \ref{FigAirfoilPressure}, there is good agreement between the ground truth and prediction on both the upper and lower surfaces of NACA 0028 for the angle of attack of $0^\circ$. For the angle of attack of $30^\circ$, a relatively good agreement is observable for the front lower surface of the airfoil as well as the rear upper surface. In contrast, the opposite occurs for the angle of attack of $-30^\circ$. In this case, although the predicted pressure on the front upper and rear lower surfaces of NACA 0028 are in good agreement with the ground truth, relatively high errors happen at other locations of the airfoil surface. As mentioned earlier in this subsection, this generalizability experiment provides insight that helps us increase the network performance. Here is the main insight from the investigation of the predicted pressure on the airfoil surface: by rotating the airfoil, the error of the pressure prediction by the network increases in the front or rear of the airfoil, but not in both parts for a given angle of attack. This is mainly because the network has never seen the curvature associated with the upper or lower surface. Therefore, at prediction time, sometimes the network only considers the curvature pattern of the airfoil front and treats the airfoil similar to a circle or an ellipse with a small length scale. In these cases, the maximum error happens in rear of the airfoil (either on the upper or lower surface). On the other hand, sometimes the network only views the curvature pattern of the airfoil rear and treats the airfoil similar to an ellipse, triangle, or rectangle with a large length scale. In these cases, the maximum error occurs in front of the airfoil (either on the upper or lower surface). To resolve this issue, we need to adjust our neural network such that it views the curvature of both the front and rear of the airfoil simultaneously. PointNet\texttt{++} (Ref. \onlinecite{qi2017pointnet++}) makes this feasible by means of two fundamental features of multi-scale grouping (MSG) and multi-resolution grouping (MRG) which allows the network to learn better local geometric features in addition to global ones. As is apparent from the name, PointNet\texttt{++} \cite{qi2017pointnet++} is an advanced version of PointNet \cite{qi2017pointnet}, which our current neural network is based on. One may refer to the original PointNet\texttt{++} paper \cite{qi2017pointnet++} for the further details.

To close this subsection, we discuss a few points. First, as can be seen in Fig. \ref{Fig9}, the point cloud accurately illustrates the geometry of NACA 0028 through the null space. We have also explained this characteristic of our network in Fig. \ref{Fig0} and its caption. Second, the high density of points chosen near the airfoil surface had not been experienced by the network during training, demonstrating another key feature of our neural network. As pointed out earlier, this organization cannot be captured using traditional two or three-dimensional CNNs. Third, even though the network had never seen airfoils during the training operation, the NACA 0028 length scale, and consequently the associated Reynolds number, was in the range of training data set.

\subsection{\label{34}Potentials for turbulent flow predictions}

In this article, our main focus is the prediction of laminar steady flows. However, the scientific field of deep learning of turbulent flows is advancing (see Ref. \onlinecite{duraisamy2019turbulence} for a review of recent articles in this scope). To assess the capability of our neural network for turbulent flow predictions, a thorough set of machine-learning experiments is mandatory. However, several critical considerations must be addressed before starting such experiments. First, to obtain a resolved turbulent flow solution for a given geometry, a finer mesh is usually required in comparison with the corresponding laminar flows (see e.g., Ref. \onlinecite{kravchenko1999b}). Thus, more input points (increased $N$) are essential to capture all the grid vertices of the area of interest. Moreover, due to the greater complexity of physics in turbulent flows compared to laminar flows, a larger global feature size is necessary to capture all geometric and physical features. Consequently, a deeper neural network is required. The architecture of our neural network (see Fig. \ref{Fig3}) has this flexibility. For instance, one may make the network deeper simply by adding two MLP components with three layers: one right before the ``max pooling'' operator, and one just after the ``global feature'' (see Fig. \ref{Fig3}). Such a change preserves the key features of our network and the integrity of the associated underlying mathematical theories presented in the original PointNet paper \cite{qi2017pointnet}. Second, efficient unstructured grids are the key tool for simulating turbulent flows in boundary layers and complex geometries (see e.g., Ref. \onlinecite{mahesh2004numerical}). As shown in Sect. \ref{31} and Sect. \ref{33} (specifically, see Figs. \ref{Fig9}--\ref{Fig11}), our deep learning approach is designed to connect unstructured data to our neural network for the aim of training and prediction. Third, recent research studies have highlighted the contribution of deep learning techniques to derive different terms in large eddy simulation equations (see e.g., Refs. \onlinecite{maulik2017neural,maulik2019subgrid,beck2019deep}). However, the proposed deep learning networks in these studies \cite{maulik2017neural,maulik2019subgrid,beck2019deep} are limited to Cartesian grids with uniform grid distribution. Considering the features of our neural network, it can leverage these deep learning approaches for the large eddy simulation of turbulent flows on irregular geometries.

\section{\label{Conclusions}Conclusions and future directions}

Deep learning algorithms have become popular among the CFD community, particularly for the study of geometrical parameters for the purpose of design optimization. Realistic CFD problems usually involve complicated domains, where Cartesian grids are not suitable. However, recent strategies used in the community have been to project scattered CFD data into Cartesian grids through pixelation and then utilize CNNs as deep learning tools.

In this article, we proposed a novel deep learning framework for the prediction of velocity and pressure fields (or any other field of interest) in complicated domains. The construction of this framework was mainly accorded to the segmentation architecture of PointNet \cite{qi2017pointnet}. Using our framework, the order of accuracy of the CFD data was preserved. Moreover, the framework was able to capture the effect of small changes in the geometry of domains, something that is essential for an efficient design optimization. Additionally, no artificial effect such as spurious roughness was introduced to the geometry of the domain, specifically, the smoothness of curved surfaces was preserved. By means of this framework, there was freedom to set a non-uniform distribution of points for inquiries over the CFD domain. Thus, users could select a fine-scaled point distribution in important zones (from a CFD point of view) of the domain and a coarse scale in other areas. None of these key features were achievable by the extant pixelation techniques. 

In order to inspect the performance of our neural network, two-dimensional incompressible laminar steady flow past a cylinder with different shapes for its cross section was considered. While the density and viscosity of the fluid were fixed, the length scale of cross section varied, mainly due to the rotation of the cross section, and changes in its shape and size. This setting produced our data set. The $L^2$ norm of the predicted velocity and pressure fields over unseen data demonstrated an excellent performance of the presented machine learning framework. The network predictions were hundreds of times faster compared to our traditional CFD solver. By computing the residuals of the continuity and Navier-Stokes equations, it was demonstrated that the mass and momentum were conserved with an excellent to reasonable level of accuracy. We discussed the physical interpretation of different layers of our neural network through the concept of critical and non-critical points. Furthermore, the generalizability of our neural network was discussed by two examples. First, while the network had seen point clouds representing only one object during training, it reasonably predicted the flow fields past two and three blunt bodies existed in a domain. Second, even though the neural network had never seen airfoils during the training procedure, it could reasonably predict the flow fields for NACA 0028 for three angles of attack of $0^{\circ}$, $30^{\circ}$, and $-30^{\circ}$. In this study, we replaced the whole CFD solver by our deep learning framework; however, it is possible to only replace a specific component of a CFD solver by our neural network. For example, one may replace a Poisson solver by the network embedded in the projection schemes to predict pressure or intermediate pressure fields (as the network output) from the scalar field of the divergence of intermediate velocity (as the network input). As a final note, although we explored the application of our deep learning framework in CFD, it is applicable to other areas of computational mechanics (see e.g., Ref. \onlinecite{nie2020stress}).

There are several potential directions for our future studies. One of these is to perform unsteady fluid dynamics prediction using deep learning techniques. Generally speaking, fluid flow becomes unsteady due to physical features and/or geometrical changes. For instance, when the Reynolds number ($Re$) approximately exceeds 40.0 in the case of a two-dimensional flow past a circular cylinder, a transition from steady to unsteady behavior is observed \cite{ding2004simulation}. Although this is an unsteady test case, the geometry, and consequently the generated mesh, is fixed in time. However, we are particularly interested in transient problems wherein an object or interface moves inside the CFD domain, and consequently unstructured grids need to be regenerated during simulations \cite{tremel2007parallel}. Examples of classical benchmark problems are shock-box interaction and objects falling into free stream \cite{lohner1989adaptive}. Therefore, the target is to design a deep learning framework to handle these types of unsteady problems. This can be accomplished by integrating our current neural network with a recurrent autoencoder (see e.g., Ref. \onlinecite{chaitanya2017interactive}) or using spatio-temporal point cloud representation techniques (see e.g., Ref. \onlinecite{niemeyer2019occupancy}).

Another topic for future investigation is the prediction of flow fields in three-dimensional spaces when the influence of variations in the shape of a three-dimensional object (e.g., an airplane wing) on the velocity and pressure fields in an area of interest matters. For this subject, our methodology is slightly different from what was proposed in the current work. In fact, we only consider the grid points located on the surface of the three-dimensional object (e.g., the wing) and extract its features using the encoder component of our current neural network. Afterwards, the latent code is concatenated with the spatial coordinates of the area of interest. The new code is used as the input of a decoder, similar to the DeepSDF network (see e.g., Ref. \onlinecite{DBLP:conf/cvpr/ParkFSNL19}), to predict the velocity and pressure fields of the area of interest. We hope that this design could manipulate a wide range of scientific and engineering applications.

Finally, we are interested in merging our network introduced here with the physics-informed neural network (PINN) \cite{raissi2019physics,jagtap2020conservative,chen2020physics}. The application of current versions of PINN is limited to fixed geometries (see e.g., Ref. \onlinecite{mao2020physics}). This combination assembles a semi-supervised deep learning configuration for prediction of the solution of partial differential equations in realistic and complicated domains as well as dynamic geometries.

\begin{acknowledgments}
The authors would like to thank the Vannevar Bush Faculty Fellowship (VBFF) from the Department of Defense (DoD) and gifts from Amazon AWS and Google for providing the funding support for this study. Moreover, the authors wish to thank the reviewers for their insightful comments.
\end{acknowledgments}

\section*{Data availability}
The data that support the findings of this study are available from the corresponding author upon reasonable request.

\section*{References}
\bibliography{aipsamp}

\end{document}